%% file: main.tex
\documentclass[11pt]{article}

\PassOptionsToPackage{numbers, compress}{natbib}
\usepackage[utf8]{inputenc}
\usepackage{amsfonts}
\usepackage{mathrsfs}
\usepackage{amssymb}
\usepackage{amsthm}

\usepackage{libertine}
\usepackage{wrapfig}
\usepackage{graphicx}
\usepackage{multirow}
\usepackage{epstopdf}
\usepackage{multicol}
\usepackage{subfigure}
\usepackage{booktabs}

\usepackage{enumerate}
\usepackage{algorithmic}
\usepackage{algorithm}
\usepackage{fontawesome}
\usepackage[algo2e,linesnumbered]{algorithm2e}
\usepackage{stackengine}
\usepackage{graphicx}
\usepackage{marvosym}
\usepackage{multirow}
\usepackage{epstopdf}
\usepackage{multicol}
\usepackage{caption}
\usepackage{subfigure}
\usepackage{enumitem}
\usepackage{amsmath}
\usepackage{pifont}
\usepackage{natbib}
\usepackage{bbm}
\usepackage{makecell}
\usepackage{mathtools}
\usepackage[table]{xcolor}
\definecolor{myblue}{HTML}{4594c1}
\definecolor{mylinkcolor}{HTML}{E74D3B}
\definecolor{myurlcolor}{HTML}{980000}
\definecolor{mydarkred}{rgb}{0.6,0,0}
\definecolor{mydarkgreen}{rgb}{0,0.6,0}
\usepackage[colorlinks,
linkcolor=mydarkred,
urlcolor=mylinkcolor,
citecolor=mydarkgreen]{hyperref}
\usepackage{url}
\usepackage{bm}
\usepackage{pifont}
\usepackage{stfloats}
\usepackage[T1]{fontenc}
\usepackage{arydshln}
\usepackage{colortbl}
\usepackage{balance}

\newcolumntype{L}[1]{>{\raggedright\let\newline\\\arraybackslash\hspace{0pt}}m{#1}}
\newcolumntype{Y}{>{\centering\arraybackslash}X}
\newcolumntype{s}{>{\hsize=.3\hsize}Y}
\newcolumntype{t}{>{\hsize=1.5\hsize}X}
\newcolumntype{u}{>{\hsize=0.8\hsize}Y}
\usepackage{makecell}
\usepackage{mathtools}
\usepackage[utf8]{inputenc}
\usepackage{stfloats}
\usepackage{amsmath}

\usepackage{ragged2e}
\justifying\let\raggedright\justifying

\newcommand\nnfootnote[1]{%
  \begin{NoHyper}
  \renewcommand\thefootnote{}\footnote{#1}%
  \addtocounter{footnote}{-1}%
  \end{NoHyper}
}

\title{\textbf{A Comprehensive Survey of Data Augmentation \\in Visual Reinforcement Learning}}

\author{
  \textbf{Guozheng Ma$^{1}$,}  
  \textbf{Zhen Wang$^{2}$,}   
  \textbf{Zhecheng Yuan$^{3}$,}\\[0.5ex]
  \textbf{Xueqian Wang$^{3}$,}
  \textbf{Bo Yuan$^{4,}$\textsuperscript{\Letter},}
  \textbf{Dacheng Tao$^{1,}$\textsuperscript{\Letter}}\\[1ex]
  \small{$^1$Nanyang Technological University,}
  \small{$^2$The University of Sydney,} \\[0.5ex]
  \small{$^3$Tsinghua University,}
  \small{$^4$The University of Queensland}\\[1ex]
  \small{GUOZHENG001@e.ntu.edu.sg}\\
  \small{boyuan@ieee.org \quad dacheng.tao@ntu.edu.sg}
\date{\today}
}

\begin{document}
\maketitle
\begin{abstract}
Visual reinforcement learning (RL), which makes decisions directly from high-dimensional visual inputs, has demonstrated significant potential in various domains.
However, deploying visual RL techniques in the real world remains challenging due to their low sample efficiency and large generalization gaps.
To tackle these obstacles, data augmentation (DA) has become a widely used technique in visual RL for acquiring sample-efficient and generalizable policies by diversifying the training data.
This survey aims to provide a timely and essential review of DA techniques in visual RL in recognition of the thriving development in this field. 
In particular, we propose a unified framework for analyzing visual RL and understanding the role of DA in it.
We then present a principled taxonomy of the existing augmentation techniques used in visual RL and conduct an in-depth discussion on how to better leverage augmented data in different scenarios.
Moreover, we report a systematic empirical evaluation of DA-based techniques in visual RL and conclude by highlighting the directions for future research.
As the first comprehensive survey of DA in visual RL, this work is expected to offer valuable guidance to this emerging field.
A well-classified paper list that will be continuously updated can be found at this \href{https://github.com/Guozheng-Ma/DA-in-visualRL}{\faGithub ~GitHub site}.\nnfootnote{\Letter: Corresponding authors.}
\end{abstract}

\newpage
\tableofcontents

\newpage
\input{textfile/1_Introduction}
\input{textfile/2_Preliminaries}
\input{textfile/3_How_to_Augment_Data_in_Visual_RL}

\input{textfile/4_How_to_Leverage_Augmented_Data_in_Visual_RL}
\input{textfile/5_Experimental_Evaluation}
\input{textfile/6_Discussion_and_future_works}
\input{textfile/7_Conclusion}


\newpage

{
\small
\bibliographystyle{unsrtnat}
\bibliography{ref}
}

\end{document}

%% file: textfile/1_Introduction.tex
\section{Introduction}
Reinforcement learning (RL) addresses sequential decision-making problems in which an agent seeks to discover the optimal policy via trial-and-error interactions with the environment~\cite{introduction2drl, dqn2013, ppo, ddpg}. 
Visual RL, a variant that learns directly from visual observations such as images, has gained widespread application across various domains due to its intuitive and cost-effective approach to environmental perception~\cite{vai, sac_ae}. 
This paradigm has been successfully employed in video games~\cite{AlphaZero}, autonomous driving~\cite{2021autonomous_driving}, robot control~\cite{qt-opt}, and other areas.
However, learning directly from high-dimensional visual observations remains largely hindered by the challenges of low sample efficiency and large generalization gaps~\cite{dmc,SLAC,sac_ae,rad,drq,drq_v2,svea}.

\vspace{0.5\baselineskip}
To learn sample-efficient and generalizable visual RL agents, a considerable amount of effort has been devoted to developing diverse approaches, including 
$(1)$ applying \textbf{explicit regularization} techniques such as entropy regularization~\cite{study_on_overfitting, sac} to constrain the model's weights~\cite{CoinRun, regularization_matter, Consistent_Dropout};
$(2)$ performing joint learning with RL loss and \textbf{auxiliary tasks} to provide additional representation supervision~\cite{sac_ae, mlr, PI-SAC, spr, playvirtual, vcd, curl, CCLF, ADAT, dribo, st_dim, ccfdm, M-CURL, DRIML, cody};
$(3)$ building a \textbf{world model} of the RL environment that allows learning behaviors from imagined outcomes~\cite{planet,dreamer,dreamerv2,SLAC}; and
$(4)$ \textbf{pretraining an encoder} that can project high-dimensional observations into compact state representation~\cite{atc, proto, apt, cic, pieg, VRL3, PVR, mvp, RRL, rakelly2021mutual}.

\vspace{0.5\baselineskip}
Although these approaches have achieved remarkable success, they remain challenged by limited interaction data and poor sample diversity~\cite{rad, drq, drq_v2}. 
To address these limitations by increasing the quantity and diversity of training data, data augmentation (DA) has garnered increasing attention from the visual RL community in recent years~\cite{drq, drq_v2, svea, spr, secant}.
As a data-driven method, DA is orthogonal to the aforementioned approaches and can be combined with them to further enhance performance~\cite{spr, mixreg}. 
For instance, DA plays a crucial role in contrastive-based auxiliary tasks, injecting prior knowledge of task invariance~\cite{curl, atc, huang2021towards}.
In addition, DA is essential for pre-training a cross-task representation~\cite{atc, proto}. 
Furthermore, various DA techniques, such as random cropping, have been incorporated into almost all visual RL algorithms as a form of data preprocessing~\cite{dreamerv2, ccfdm, vai}.

\begin{figure}[ht] 
\centering 
\includegraphics[width=\textwidth]{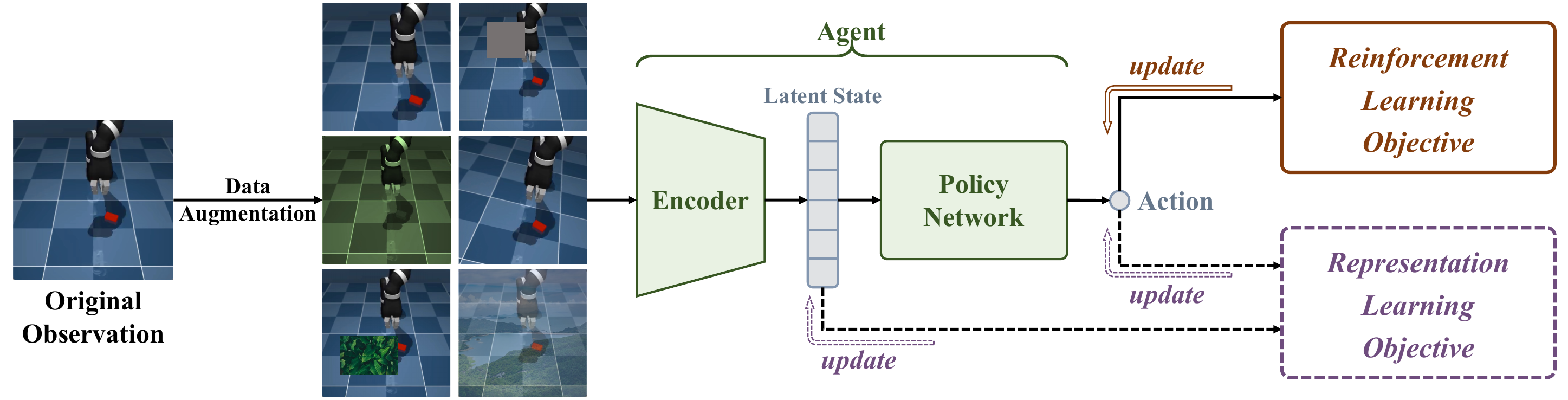}
\caption{The generic workflow diagram for leveraging DA in visual RL.} 
\label{generic workflow} 
\end{figure}

In general, DA refers to strategies for generating synthetic training data from existing data without additional collection or interaction efforts~\cite{image_aug_survey, NLP1}. Figure~\ref{generic workflow} illustrates the generic workflow for leveraging DA in visual RL: diverse augmented data are generated by manipulating the original interaction data and then exploited to optimize the RL objective~\cite{rad, drq, drq_v2}. Moreover, DA can further enhance the representation learning in visual RL by incorporating auxiliary objectives~\cite{soda, drac, curl, spr}.
Despite the surge of related studies on leveraging DA in visual RL scenarios, this fast-evolving and expanding field still lacks clarity and coherence. Therefore, this comprehensive survey aims to provide a bird's-eye view of DA-based methods in visual RL with the following main contributions:

\begin{enumerate}
    \item[1.] Based on previous works~\cite{Generalisation_survey, observational_overfitting}, we present High-Dimensional Contextual Markov Decision Process (HCMDP) as a general framework to formalize visual RL. This framework provides deep insights into the challenges of \textbf{low sample efficiency} and \textbf{large generalization gaps} in visual RL, which serve as the primary motivations for introducing DA.
    \item[2.] We identify two key assumptions of DA with different motivations: the \textbf{optimality invariance} assumption for improving sample efficiency and the \textbf{prior-based diversity} assumption for narrowing the generalization gap.
    \item[3.] We categorize related studies from two principled perspectives: \textbf{how to augment data} and \textbf{how to leverage augmented data} for improved clarity and coherence. This classification provides a structured framework for systematically reviewing existing work, offering a clear and logical organization of the field's current state.
    \item[4.] We conduct \textbf{a unified empirical evaluation} of extensive DA-based methods on representative benchmarks to evaluate their sample efficiency and generalization abilities.
    \item[5.] We present \textbf{an in-depth discussion} of the current landscape of DA in visual RL, forming the core of our survey. This section delves into the unique mechanisms of DA in this context, explores its significant challenges and opportunities, and provides crucial insights into both the present state and promising future directions of DA in visual RL.
\end{enumerate}

\begin{figure}[ht] 
\centering 
\includegraphics[width=\textwidth]{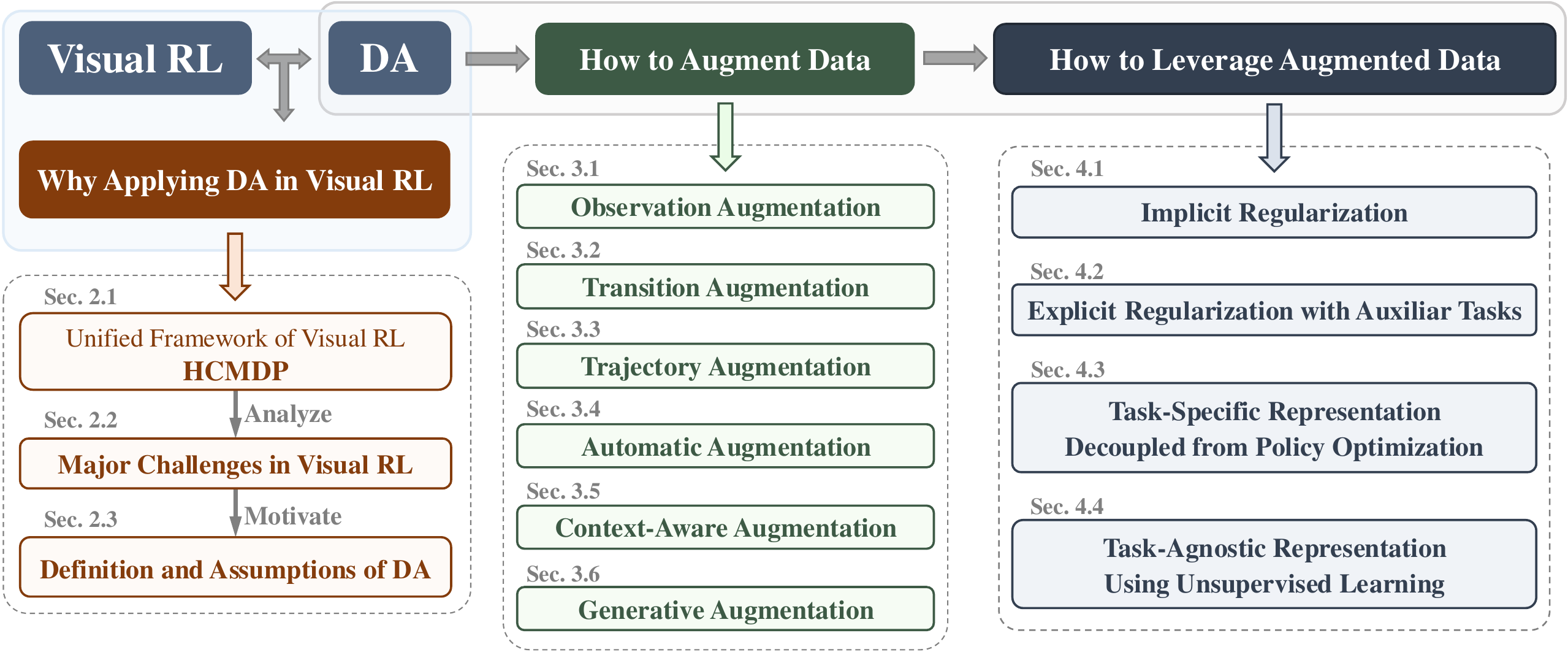}
\caption{The schematic structure of this survey.} 
\label{Outline} 
\end{figure}

The body of this survey is organized as Figure~\ref{Outline}.
In Section~\ref{Preliminaries}, we propose a unified high-dimensional contextual Markov decision process (HCMDP) framework (Section~\ref{High Dimensional Contextual MDPs}) to formalize the visual RL scenario and highlight its major challenges (Section~\ref{Major Challenges in Visual RL}), as well as present the motivations and definitions of DA in visual RL (Section~\ref{Data Augmentation in Visual RL}).
We then conduct a systematic review of the previous work from two perspectives: how to obtain and how to leverage augmented data in visual RL (Section~\ref{How to augment the data in Visual RL?} and Section~\ref{How to leverage}).
In Section~\ref{How to augment the data in Visual RL?}, we categorize DA approaches in visual RL into \textit{observation augmentation}, \textit{transition augmentation}, and \textit{trajectory augmentation}, based on the type of data each technique aims to modify.
Moreover, we introduce three advanced extensions: \textit{automatic augmentation}, \textit{context-aware augmentation}, and \textit{generative augmentation}.
In Section~\ref{How to leverage}, we present the different mechanisms used to leverage DA in visual RL, including implicit and explicit regularization, task-specific representation learning decoupled from policy optimization, and task-agnostic representation learning using unsupervised learning.
To reveal the practical effect of DA, we introduce the typical benchmarks and summarize the empirical performance of recent DA-based methods in Section~\ref{Experimental Evaluation}.
In Section~\ref{Discussion of future works}, we put forward a critical discussion concerning future research directions, including the opportunities, challenges, limitations, and underlying mechanisms of DA.
Finally, this survey is concluded in Section~\ref{Conclusion} with a list of key insights.

\paragraph{Scope.}
Given the multitude of topics and research areas related to DA and visual RL, we constrain the scope of this survey in several ways to ensure its feasibility.
\textcolor{myblue}{$\bullet$}~\textit{Firstly}, this survey does not cover the related topic of domain randomization (DR)~\cite{dr1,dr2}, which aims to solve the sim-to-real problem in robot control by tuning the physical simulator's parameter distribution to align as closely as possible with reality~\cite{posterior_domain_randomization,horvath2022object, huber2024domain}. 
In contrast, DA can only manipulate observations post-rendering, without access to the simulator's internal parameters, which affords it greater flexibility~\cite{Generalisation_survey}.
\textcolor{myblue}{$\bullet$}~\textit{Secondly}, this survey focuses on scenarios that involve learning directly from visual inputs (visual RL) rather than handcrafted state inputs (state-based RL). Consequently, several works that introduce DA in state-based RL, will not be prominently featured in this survey~\cite{counterfactual_da, lin2020invariant}.
\textcolor{myblue}{$\bullet$}\textit{Thirdly}, while DA is a powerful tool, it is not the sole approach for improving sample efficiency and generalization in visual RL.
To maintain coherence and focus, this survey does not provide detailed coverage of works that use DA as a foundational technique but whose primary research focus lies elsewhere\cite{BBF,yuan2024learning}.
We strongly recommend readers interested in generalization issues in RL to refer to another comprehensive survey~\cite{Generalisation_survey} that focuses on generalization in deep RL.
\textcolor{myblue}{$\bullet$}\textit{Finally}, to ensure this survey aligns with the latest developments in AI field, we provide a detailed introduction to recent research on DA using advanced generative models in Section 3.6. Additionally, in Section 6.5, we critically examine the role and relevance of visual RL and DA in the context of the rapidly evolving landscape of foundation models.

%% file: textfile/2_Preliminaries.tex
\newpage
\section{Preliminaries}
\label{Preliminaries}
Visual RL addresses high-dimensional image observations instead of well-designed states and has encountered a series of new challenges~\cite{sac_ae, drq}.
This section analyzes visual RL in depth and introduces the formalism of DA used for visual RL.
In Section~\ref{High Dimensional Contextual MDPs}, we present a novel framework, HCMDP, to formalize the paradigm of visual RL.
Based on this framework, we analyze the major challenges faced by visual RL in Section~\ref{Major Challenges in Visual RL}.
Finally, Section~\ref{Data Augmentation in Visual RL} introduces the formalism of DA in visual RL, including its motivation, definition and two key assumptions.

\subsection{High-Dimensional Contextual MDP (HCMDP)}
\label{High Dimensional Contextual MDPs}

The standard RL task is often defined as a {\textbf{Markov Decision Process (MDP)}}~\cite{dqn2013}, which is specified by a tuple $\mathcal{M}=(\mathcal{S},\mathcal{A}, r, \mathcal{P},p,\mathcal{\gamma})$ where $\mathcal{S}$ is the state space; $\mathcal{A}$ is the action space; $r: \mathcal{S} \times \mathcal{A} \times \mathcal{S} \mapsto \mathbb{R}$ is the scalar reward function; $\mathcal{P}(s^{\prime}|s,a)$ is the transition function; $p(\cdot)$ is the initial state distribution; and $\gamma \in (0,1] $ is the discount factor.
The goal of RL is to learn an optimal policy $\pi^{*}(a \mid s)$ that maximizes the expected cumulative discounted return $\mathcal{R}(\pi,\mathcal{M})$, which is defined as:
\begin{equation}
\mathcal{R}(\pi,\mathcal{M})=\mathbb{E}_{{s_o \sim p(\cdot)},{a_{t} \sim \pi\left(\cdot \mid s_{t}\right)},{s_{t+1} \sim \mathcal{P}(\cdot \mid s_{t}, a_t)}}\left[\sum_{t=0}^{\infty} \gamma^{t} r\left(s_{t}, a_{t}, s_{t+1}\right)\right]
\end{equation}

\begin{figure}[ht] 
\centering 
\includegraphics[width=\textwidth]{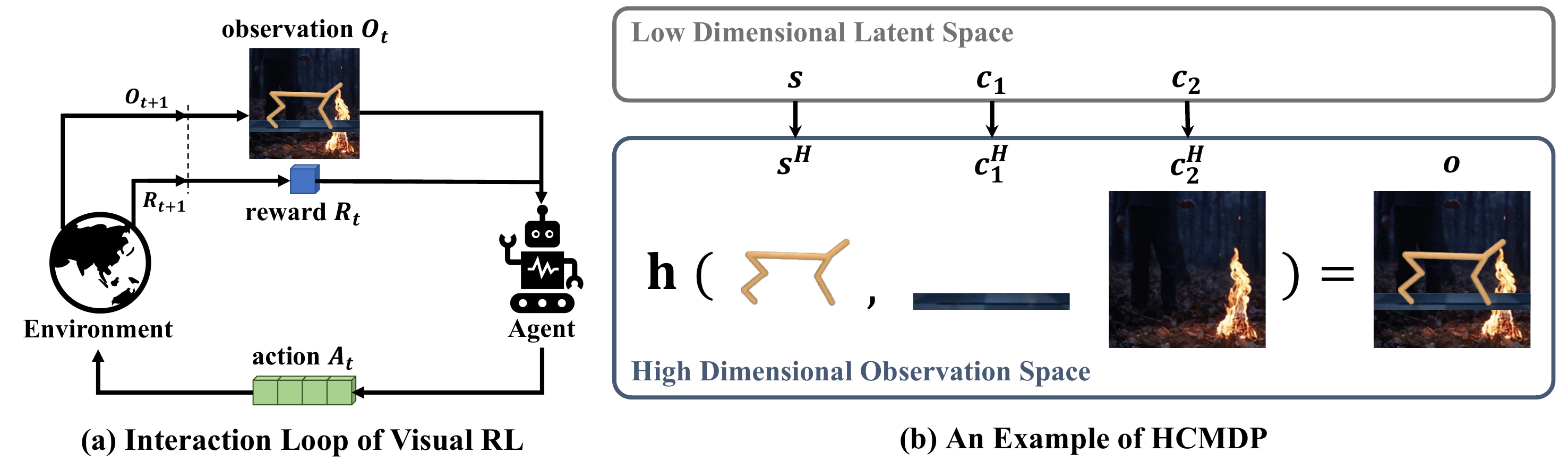} 
\caption{The agent-environment interaction loop of visual RL and an example of HCMDP.}
\label{Visual RL and HCMDP}
\end{figure}

Although the MDP is the standard paradigm of RL, it ignores a crucial factor of visual RL: agents only have direct access to high-dimensional observations instead of the actual state information.
To properly formulate the visual RL scenarios, as shown in Figure~\ref{Visual RL and HCMDP},
many variants of MDPs~\cite{block_mdp,gbmdp,epomdp,Hidden_parameter_mdp} have been introduced by using the high-dimensional observation space $\mathcal{O}$ to represent the image inputs.
Depending on the specific assumptions, an emission function $\phi:\mathcal{S} \mapsto \mathcal{O}$ can be designed to simulate the mapping from the state space $\mathcal{S}$ to the observation space $\mathcal{O}$.
For example, the $(f, g)$-scheme~\cite{observational_overfitting} constructs an emission function as the combination of generalizable and non-generalizable features while the contextual MDP (CMDP)~\cite{cmdp15,Hidden_parameter_mdp,what&why,Generalisation_survey} introduces context $c$ to distinguish contextual information from the underlying state information.
However, these MDP variants mainly focus on how to explain the generalization effect in visual RL, and ignore the issue of constructing a compact representation from high-dimensional observations.

\vspace{0.5\baselineskip}
To better understand visual RL scenarios and provide a unified view of its specific challenges,
we propose \textbf{High-Dimensional Contextual MDP (HCMDP)} as a general modeling framework of visual RL.
Following the previous formalism~\cite{Generalisation_survey,observational_overfitting}, the HCMDP $\mathcal{M}|_{C}$ can be defined as a family of environments:
\begin{equation}
\mathcal{M}|_{C}=\{\mathcal{M}|_{c}=(\mathcal{M},\mathcal{O}_c,{\phi}_c) \mid c\sim p(c),c \in C \}
\end{equation}
where $\mathcal{M}=(\mathcal{S},\mathcal{A}, r, \mathcal{P},p,\mathcal{\gamma})$ specifies the dynamics of the underlying system.
With the fixed base MDP $\mathcal{M}$, the observation space $\mathcal{O}_c$ and emission function ${\phi}_c$ depend on the context $c$, which refers to the peripheral parameters that are not essential for agents to make decisions.
$p(c)$ is the \textit{context distribution}, and $C$ represents the entire \textit{context set}.
For example, the colors and styles of backgrounds in robot scenarios are extraneous to control tasks, and are thus being referred to as task-irrelevant features.

\vspace{2\baselineskip}
To be more specific, context $c$ can be denoted as a set of parameters $\{c_1,c_2,\dots,c_n\}$, where $n$ is the number of task-irrelevant properties in this system.
Each $c_i$ corresponds to a task-irrelevant property, all of which are distributed over a fixed range: $\left\{c_{1} \in C_{1}, c_{2} \in C_{2}, \ldots, c_{n} \in C_{n}\right\}$.
Consider an autonomous driving example such as CARLA~\cite{carla}: an agent learns to control the car directly from pixels in changing environments. 
Therefore, the agent must distinguish between task-relevant and task-irrelevant components in the image observations.
For instance, we can denote the style of the background buildings as $c_1$, the color of the driving car as $c_2$ and the number of people walking on the sides of the road as $c_3$.

\vspace{0.5\baselineskip}
The state $s$ and context $c$ constitute the complete information (parameters) used by the system to render the final observed images~\cite{observational_overfitting}.
However, they both exist in the low-dimensional latent space, which cannot be directly observed.
In fact, $\mathcal{O}$ is the only observable high-dimensional space where agents perceive task information.
Following the assumptions~\cite{what&why,observational_overfitting} that observations are high-dimensional projections of the state $s$ and task-irrelevant contexts $c$, the emission function $\phi_c$ mapping from state $s \in \mathcal{S}$ to observation $o \in \mathcal{O}_c$ can be defined as:
\begin{equation}
o = \phi_c(s) := \mathbf{h}(s^H,c_1^H,c_2^H,\dots,c_n^H)
\end{equation}
where $s^H$ is the high-dimensional representation mapped from the underlying state $s$, and each $c_i^H$ is the high-dimensional representation uniquely determined by the latent context $c_i$.
Similar to the formalism in~\cite{observational_overfitting}, $\mathbf{h}$ is a "combination" function that combines the task-relevant state representation $s^H$ and task-irrelevant context representations $(c_1^H,c_2^H,\dots,c_n^H)$ to render the final observation.
Based on the HCMDP framework, Figure~\ref{Visual RL and HCMDP} shows an illustration of a robot control environment from the DeepMind control suite~\cite{dmc}.
In this scenario, contexts $c_1$ and $c_2$ separately denote the floor color and background style, respectively, which are both irrelevant to the control task.
Correspondingly, $c_1^H$ and $c_2^H$ are the high-dimensional representations mapped from $c_1$ and $c_2$.
The final observation $o$ is the combination of the state representation $s^H$ and the task-irrelevant representations $c_1^H$ and $c_2^H$.

\vspace{0.5\baselineskip}
An HCMDP $\mathcal{M}|_{C}$ consists of a family of specific environments, where $c$ follows the \textit{context distribution} $p(c)$ over the entire \textit{context set} $C$. In a given system, $\mathcal{M}$ and the rendering rules from $s$ and $c_i$ to the high-dimensional representations $s^H$ and $c_i^H$ are established.
Hence, different combinations of the \textit{context distribution} $p(c)$ and \textit{context set} $C$ produce different HCMDPs.
For any HCMDP $\mathcal{M}|_{C}$, the expected return of a policy is defined as:
\begin{equation}
\mathbf{R}(\pi, \mathcal{M}|_{C}):=\mathbb{E}_{c \sim p(c),c \in C }\left[\mathcal{R}\left(\pi,\mathcal{M}|_{c}\right)\right]
\end{equation}
where $\mathcal{R}$ is the expected return of policy $\pi$ in a specific MDP.
In practice, we assume that the context distribution is uniform over the entire context set~\cite{Generalisation_survey} so that different HCMDPs can be specified by their context sets $C=(C_1,C_2,\dots,C_n)$.
By choosing a training context set $C_{\rm train}$ and a test context set $C_{\rm test}$, we can separately define the training context set HCMDP $\mathcal{M}|_{C_{\rm train}}$ and the test context set HCMDP $\mathcal{M}|_{C_{\rm test}}$.
Agents are only allowed to be trained in $\mathcal{M}|_{C_{\rm train}}$ and evaluated in the same HCMDP $\mathcal{M}|_{C_{\rm train}}$ or HCMDP $\mathcal{M}|_{C_{\rm test}}$, whose context exhibits a distribution shift from the training context set.

\begin{figure}[h] 
\centering 
\includegraphics[width=\textwidth]{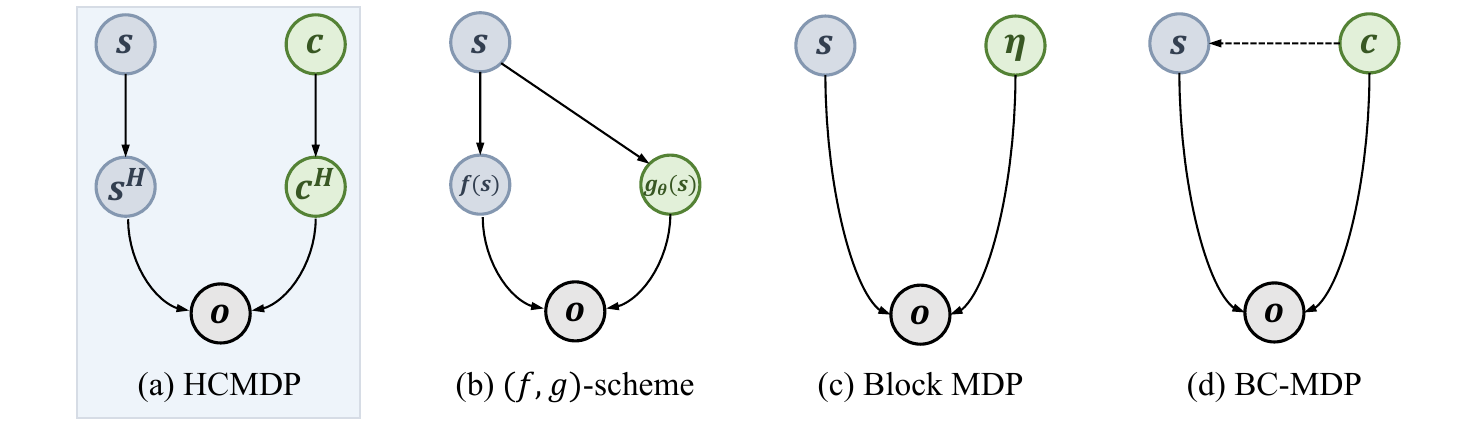}
\caption{A graphical model of the emission function of a HCMDP (a) compared with three other representative MDP variants: (b) $(f, g)$-scheme~\citep{observational_overfitting}, (c) Block MDP~\citep{causal_for_block_mdps} and (d) BC-MDP~\citep{BC-MDP}.}
\label{HCMDP Compare}
\end{figure}

\paragraph{\textbf{Remarks.}}
The main difference between HCMDP and other MDP variants lies in the emission function $\phi:\mathcal{S} \mapsto \mathcal{O}$, as shown in Figure~\ref{HCMDP Compare}.

First, the HCMDP highlights the high dimensionality of the observation space $\mathcal{O}$ by explicitly specifying the mapping between the latent variables $s,c$ and their high-dimensional representations $s^H, c^H$.
Second, it presents a unified perspective to understand the challenges of generalizing learned policy to unseen visual environments based on existing assumptions~\cite{observational_overfitting, causal_for_block_mdps, BC-MDP}.
Specifically, the HCMDP assumes that the task-relevant features of state $s$ and the task-irrelevant features of context $c$ are combined in the final observation without further assumptions about their relationship.
In the training process, agents tend to overfit the irrelevant context features and cannot effectively generate to unseen environments.
As a general framework, the HCMDP can be transformed to other feasible MDP variants by making additional assumptions.
For example, 
$(f, g)$-scheme~\cite{observational_overfitting} assumes that the unimportant features that do not contribute to extra generalizable information in observations are projected from the latent state with function $g_\theta(\cdot)$ dependent on the sampled parameter $\theta$;
Block MDP~\cite{causal_for_block_mdps} assumes that the emission function is the concatenation of the noise and state variables as $s \oplus f(\eta)$, where $\eta$ denotes spurious noise;
and BC-MDP~\cite{BC-MDP} assumes that the agent only has access to a partial state space $\mathcal{S}^c$, determined by the context $c$.
By contrast, the HCMDP ignores the specific relationship between task-relevant and task-irrelevant features, focusing instead on the compound of these components.

\vspace{3\baselineskip}
Note that the HCMDP framework does not take into account the partially observable features of the underlying states in a partially observable MDP (POMDP)~\cite{pomdp}. 
Following~\cite{dqn2013,proto,drq}, we assume that the complete state information can be reasonably constructed by stacking three consecutive previous image observations into a trajectory snippet~\cite{drq_v2}.
In summary, the motivation of HCMDP is to emphasize the fact that the underlying state $s$ is projected to the high-dimensional observation space along with the task-irrelevant information of context $c$.
With this unified framework, the unique challenges of visual RL scenarios compared with standard RL can be clearly analyzed.

\subsection{Major Challenges in Visual RL}
\label{Major Challenges in Visual RL}
Despite the success of visual RL in complex control tasks with visual observations, sample efficiency and generalization remain two major challenges that may lead to ineffective agents~\cite{spr, svea, secant, ijcai2022-543,CRESP}.
In this subsection, we present the formal definitions of sample efficiency and the generalization gap based on the HCMDP framework and discuss their mechanisms.

\subsubsection{Sample Efficiency}
This term measures how well the interaction data are leveraged to train a model~\cite{Sampl_Efficient}.
In practice, we consider an agent sample-efficient if it can achieve satisfactory performance within limited environment interactions~\cite{drq, spr}. 
In other words, the goal of sample-efficient RL is to maximize the policy's expected return during the training of HCMDP $\mathcal{M}|_{C_{\rm train}}$ based on as few interactions as possible.
The expected return of policy $\pi$ in $\mathcal{M}|_{C_{\rm train}}$ can be defined as:
\begin{equation}
J(\pi):=\mathbf{R}(\pi,\mathcal{M}|_{C_{\rm train}})
\end{equation}

Instead of making decisions based on predefined features, agents in visual RL need to learn an appropriate representation that maps a high-dimensional observation $\mathbf{h}(s^H,c^H)$ to the latent space $\mathbf{h}(s,c)$ to obtain decision-critical information~\cite{spr, drq, rad}.
Since standard RL algorithms already require large amounts of interaction data~\cite{sac}, learning directly from high-dimensional observations suffers from prohibitive sample complexity~\cite{sac_ae}.

\vspace{0.5\baselineskip}
One solution to the sample inefficiency problem in visual RL is by training with auxiliary losses, such as pixel or latent reconstruction~\cite{sac_ae, mlr}, future prediction~\cite{PI-SAC, spr, playvirtual, vcd} and contrastive learning for instance discrimination~\cite{curl, CCLF, ADAT, dribo} or temporal discrimination~\cite{cpc, st_dim, ccfdm, M-CURL, DRIML}.
Meanwhile, several model-based methods explicitly build a world model of the RL environment in pixel or latent spaces to conduct planning~\cite{planet,dreamer,dreamerv2,SLAC}.
Recently, pretrained encoders have demonstrated great potential in downstream tasks where the visual RL environment is explored in an unsupervised manner to obtain a task-agnostic pretrained encoder that can quickly adapt to diverse downstream tasks~\cite{atc, proto, apt, cic}.
In addition, applying the pretrained encoders from other domains such as ImageNet~\cite{imagenet} to visual RL also has shown its efficiency in downstream tasks~\cite{VRL3, PVR, mvp, RRL}.
The aforementioned methods have significantly improved the sample efficiency of visual RL, but the lack of training data remains a fundamental issue, which can be effectively solved by DA.
Moreover, abundant auxiliary tasks and world models are designed and trained based on the augmented data~\cite{curl, spr, playvirtual, SLAC}.
Hence, DA plays a vital role in improving the sample efficiency of visual RL algorithms.

\subsubsection{Generalization}
An agent's generalization ability can be measured by the generalization gap when transferred to unseen environments, which has been extensively investigated~\cite{observational_overfitting, block_mdp, cmdp} and reviewed~\cite{Generalisation_survey}.
For an HCMDP with varying context sets ${C_{\rm train}}$ and ${C_{\rm test}}$, the generalization gap of policy $\pi$ can be defined as:
\begin{equation}
\mathrm{GenGap}(\pi):=\mathbf{R}(\pi,\mathcal{M}|_{C_{\rm train}})-\mathbf{R}(\pi,\mathcal{M}|_{C_{\rm test}})
\end{equation}

As mentioned in Section~\ref{High Dimensional Contextual MDPs}, the task-relevant information of state $s$ is often conflated with the task-irrelevant information of context $c$, which may cause agents to overfit the task-irrelevant components~\cite{observational_overfitting}.
How to train generalizable agents across different environments remains challenging in visual RL, and distinguishing between the task-relevant and task-irrelevant components of the observed images is essential for narrowing the generalization gap.

\vspace{0.5\baselineskip}
A naive approach to enhancing generalization is to apply regularization techniques originally developed for supervised learning~\cite{CoinRun, regularization_matter}, including $\ell_2$ regularization~\cite{bn_rl}, entropy regularization~\cite{study_on_overfitting, sac}, dropout~\cite{Consistent_Dropout} and batch normalization~\cite{SNI_IB}.
However, these traditional regularization techniques show limited improvement in generalization and may even negatively impact sample efficiency~\cite{CoinRun, drq, SNI_IB}.
As a result, recent studies focus on learning robust representations to improve the agent's generalization ability by introducing bisimulation metrics~\citep{dbc, robust_bisimulation}, multi-view information bottleneck (MIB)~\citep{dribo}, pretrained image encoder~\citep{pieg} etc.
From an orthogonal perspective, DA has been effective in enhancing generalization by generating diverse synthetic data~\citep{rad, drq}. Moreover, DA can implicitly provide prior knowledge to the agent as a type of inductive bias or regularization~\citep{Generalisation_survey, instead_of_explicit_regularization}.
A detailed elaboration of the generalization issue in RL is provided in~\citep{Generalisation_survey}, which systematically reviews the related studies.

\subsection{DA in Visual RL}
\label{Data Augmentation in Visual RL}

As discussed in Section~\ref{Major Challenges in Visual RL}, the quantity and diversity of training data are crucial for achieving sample-efficient and generalizable visual RL algorithms.
DA, as a data-driven approach, has demonstrated significant potential for visual RL in terms of both sample efficiency and generalization ability~\citep{rad,drq,drq_v2,svea,secant,tlda,atc,drac}.
The advantages of DA for visual RL can be viewed from two aspects:
$(1)$ it can significantly expand the volume of the original interaction data, thus improving the sample efficiency~\citep{rad}; $(2)$ it introduces additional diversity into the original training data, making agents more robust to variations and enhancing their generalization capabilities~\citep{Generalisation_survey, svea}.
Furthermore, theoretical foundations have also been developed for DA, such as invariance learning~\cite{Invariance_Learning_220222, 0308_Regularising_for_invariance} and feature manipulation~\cite{Feature_Manipulation}.
Hence, DA has been well recognized as a viable solution for the challenges in visual RL~\cite{2022image_da_survey,Generalisation_survey,svea}.
Following the conventions in~\cite{drq,drq_v2,svea}, we define a general augmentation $\tau: \mathcal{O} \times \mathcal{V} \mapsto \mathcal{O}^{aug}$ as a mapping from the original observation space $\mathcal{O}$ to the augmented observation space $\mathcal{O}^{aug}$:
\begin{equation}
o^{aug} \triangleq \tau (o; \nu) \qquad \forall o \in \mathcal{O} , \nu \in \mathcal{V}
\end{equation}
where $\nu \in \mathcal{V}$ is a set of random parameters and $\tau(\cdot)$ is the transformation function acting on the observation $o$.
To gain an intuitive understanding of the effect of DA, we identify two assumptions of $\tau(\cdot)$ corresponding to the challenges that DA seeks to address:
the assumption of {\textbf{optimality invariance}} for improving the sample efficiency and the assumption of {\textbf{prior-based diversity}} for narrowing the generalization gap.

\subsubsection{Optimality Invariance}
In supervised learning (SL), DA methods usually assume that the model's output is invariant after transformations; therefore, they can be directly applied to labeled samples to produce supplementary data~\cite{image_aug_survey, cutout}.
Considering the property of RL, DrQ~\cite{drq} defines the \textit{optimality invariance} assumption as adding a constraint to the transformation $\tau$, which induces an equivalence relation between state $s$ and its augmented counterpart $s^{aug}$ constructed from observations $o$ and $o^{aug}$, respectively~\cite{svea}.
Hence, an optimality-invariant state transformation $\tau: \mathcal{O} \times \mathcal{V} \mapsto \mathcal{O}$ can be defined as a mapping that preserves the Q-values~\cite{svea}, V-values and policy $\pi$~\cite{drac} :
\begin{equation}
Q(o, a)=Q(\tau(o; \nu), a), V(o)=V(\tau(o; \nu)) \text{  and  } \pi(o)=\pi(\tau(o; \nu)) \quad \forall o \in \mathcal{O}, a \in \mathcal{A}, \nu \in \mathcal{V}
\label{Optimality Invariance}
\end{equation}
where $\nu $ is the set of parameters of $\tau(\cdot)$, drawn from the set of all possible parameters $ \mathcal{V}$.
Note that optimality invariance relies on strict restrictions on $\tau(\cdot)$ and the size of $\mathcal{V}$ to ensure that the same $s$ can be constructed from the original and augmented observations. 
In the HCMDP framework, optimality invariance means that augmentation transformations only change the selected contexts in the high-dimensional observation space while preserving the entire (conceptual) state information in the latent space.

\begin{wrapfigure}[9]{r}{0.51\textwidth}
\centering
\includegraphics[width=0.508\textwidth]{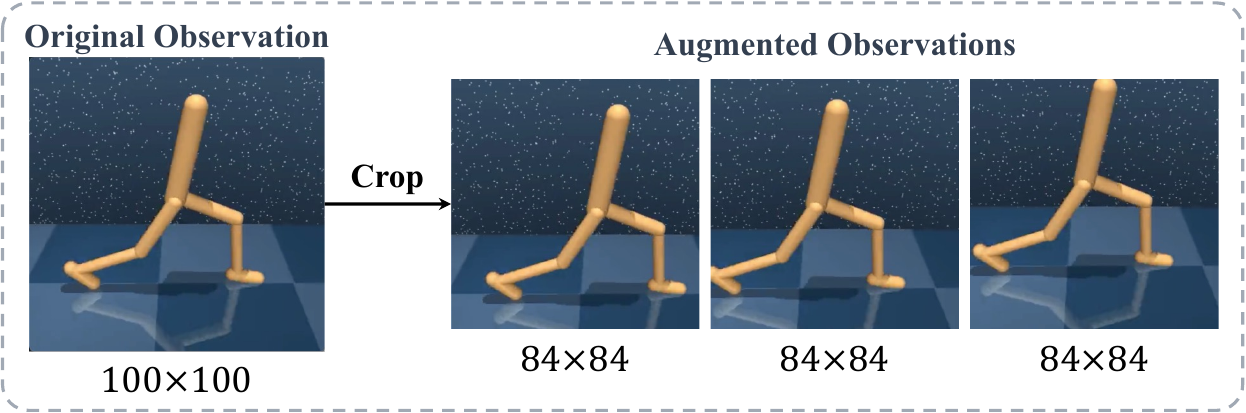}
\caption{Optimality-Invariant augmentation.}
\label{Example of Optimality Invariant Image Transformation}
\end{wrapfigure}
\vspace{0.5\baselineskip}
For instance, random cropping~\cite{rad,drq} satisfies the optimality invariance assumption in most robot control environments such as the DeepMind control suite~\cite{dmc}.
In Figure~\ref{Example of Optimality Invariant Image Transformation}, cropping generates augmented observations by randomly extracting central patches from the original image.
Since the robot is centrally placed in the images, cropping only eliminates irrelevant information such as the background color while preserving the task-relevant information such as the robot's posture~\cite{what&why}.

\vspace{0.5\baselineskip}
With the optimality-invariant augmentation of the original observations, we can obtain sufficient training data based on limited interactions with the environment so that the sample efficiency can be significantly improved~\cite{drq,drac}.
However, due to the constraint of Eq.~\ref{Optimality Invariance}, optimality-invariant augmentations cannot provide sufficient diversity to enhance the agent's generalization ability~\cite{drq, svea}.
Consequently, it is necessary to break the limitation of optimality invariance to capture the variation between the training and test environments~\cite{svea,Generalisation_survey}.

\subsubsection{Prior-Based Diversity}
\label{Prior-based Variant Image Transformation}
\begin{figure}[ht] 
\centering 
\includegraphics[width=\textwidth]{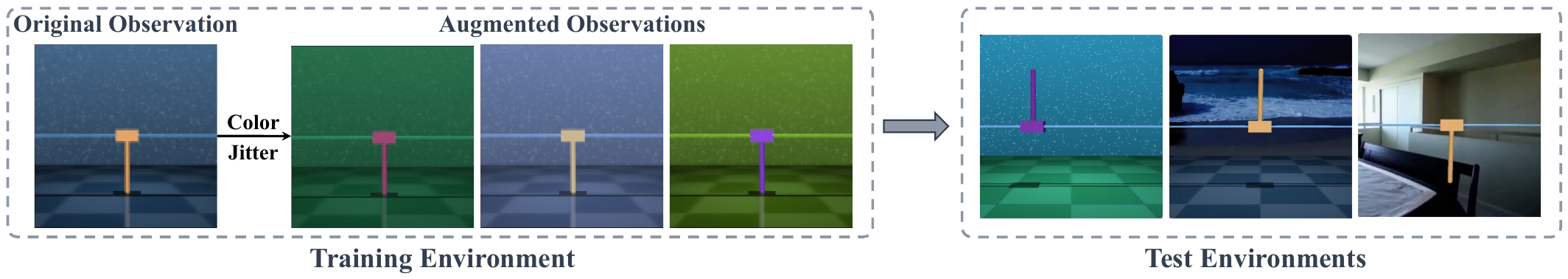}
\caption{Examples of applying DA under the assumption of prior-based diversity.}
\label{Example of Prior-based Variant Image Transformation}
\end{figure}

Based on the prior knowledge of the task-irrelevant contexts that vary between the training and test environments, targeted augmentations can be applied to effectively capture these variations~\cite{Generalisation_survey}.
Consequently, \textit{prior-based diversity} can be introduced by modifying the corresponding features in the observed images.
Note that DA can only manipulate the observed images and cannot directly change the distribution of the latent context.
Figure~\ref{Example of Prior-based Variant Image Transformation} shows a typical scenario of DMControl-GB~\cite{soda}.
With the knowledge that the background color and style may vary when transferring the agent from training environments to test environments, we can purposefully employ augmentation techniques such as color jitter to diversify the color of the training observations~\cite{svea}.
By developing an invariant policy or a latent representation from the prior-based strong augmentation (under the prior-based diversity assumption), the agents can successfully learn to identify these task-irrelevant features~\cite{Generalisation_survey}.

Strong augmentation under the prior-based diversity assumption breaks the limitation of the optimality invariance assumption and therefore has tremendous potential for improving the agent's generalization ability.
However, this approach inevitably increases the estimation variance of the Q-values and thus may harm the stability of the RL optimization process~\cite{svea, secant}.

%% file: textfile/3_How_to_Augment_Data_in_Visual_RL.tex
\newpage
\section{How to Augment Data in Visual RL?} \label{How to augment the data in Visual RL?}
The aim of DA is to increase the amount and diversity of the original training data so that agents can learn more efficient and robust policies~\cite{svea}.
Thus, a primary focus of previous research was to design effective augmentation approaches~\cite{mixreg, playvirtual}.
In this section, we introduce the mainstream augmentation techniques and discuss the pros and cons of these methods.

\begin{figure}[ht] 
\centering
\includegraphics[width=\textwidth]{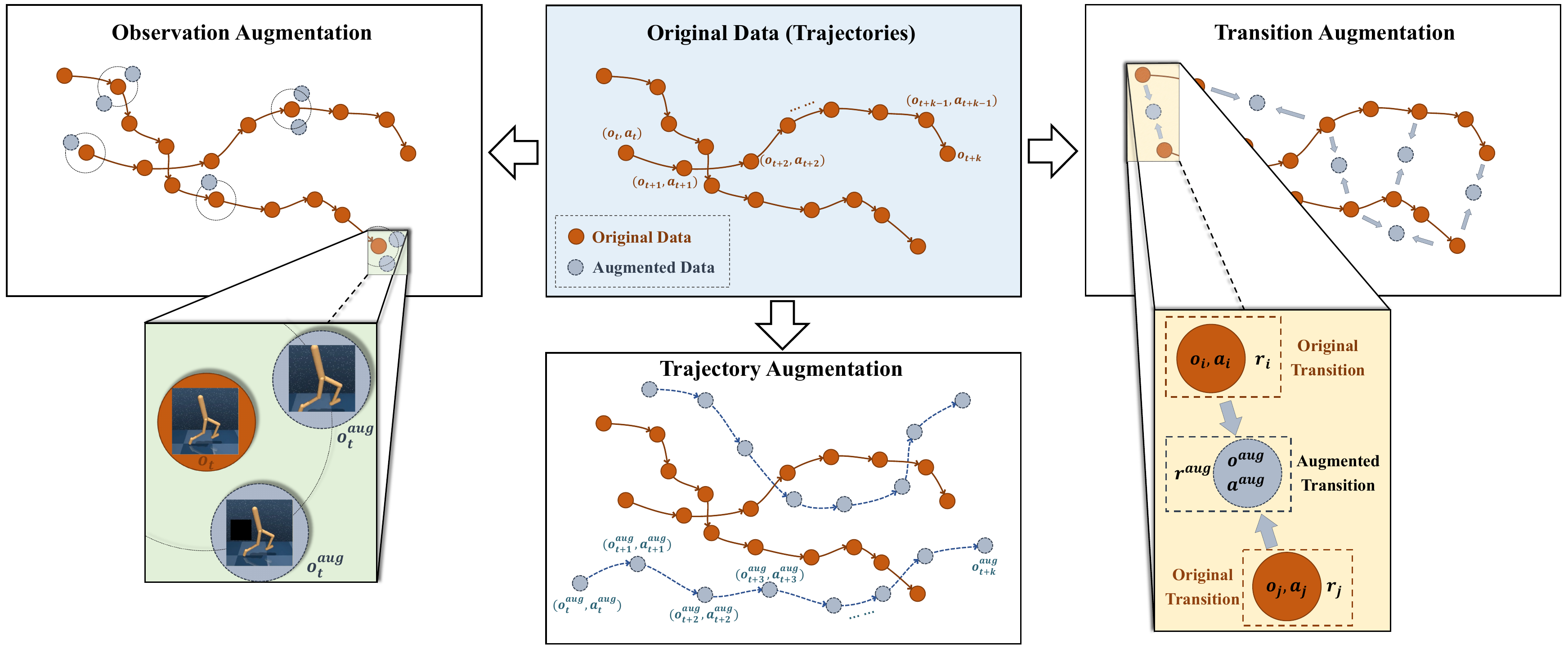}
\caption{
The comparison of different DA paradigms depending on the type of data augmented:
\textbf{observation augmentation} only generates synthetic observations $o_t^{aug}$;
\textbf{transition augmentation} augments observations together with supervision signals $(o_t^{a u g}, a_t^{a u g}, r_t^{a u g})$; and
\textbf{trajectory augmentation} generates virtual trajectories $(o_t^{a u g}, a_t^{{aug }}, o_{t+1}^{{aug }}, a_{t+1}^{a u g}, \ldots, o_{t+k}^{a u g})$.} 
\label{Overview of Section3} 
\end{figure}

\vspace{0.5\baselineskip}
Based on the type of data being augmented, we categorize the DA approaches in visual RL into three main types, as illustrated in Figure~\ref{Overview of Section3}
The first category, \textbf{observation augmentation}, involves transforming the given observations while keeping other transition factors (e.g., actions and rewards) unchanged, similar to label-preserving perturbations in SL. In Section~\ref{Observation Augmentation}, we detail various methods for employing DA on observations, which include not only diverse classical image manipulations directly applied to observation inputs but also several examples of DA in the feature space.
The other two types, \textbf{transition augmentation} and \textbf{trajectory augmentation}, specifically take into account the unique properties of RL to broaden the scope of augmentation. In Section~\ref{Transition Augmentation}, we introduce transition augmentation, which enhances observations along with supervision signals, such as rewards. Finally, in Section~\ref{Trajectory Augmentation}, we explore trajectory augmentation, focusing on generating synthesized sequential trajectories.

\vspace{0.5\baselineskip}
In addition to summarizing techniques for augmenting different data types, this section will also introduce three advanced DA techniques that enhance the diversity of DA and improve its overall effectiveness.
\textbf{Automatic augmentation} aims to automatically select the optimal DA type based on the specific task (Section~\ref{Automatic Selection of Data Augmentation}), and \textbf{task-aware augmentation} (Section~\ref{Context-aware Data Augmentation}) focuses on providing data diversity while preserving critical information within the data. Furthermore, in light of recent advancements in generative AI, contemporary research has explored the use of technologies such as GANs and diffusion models for data generation; we will discuss this \textbf{generative augmentation} approach in Section~\ref{Generative Augmentation}.

\subsection{Observation Augmentation}
\label{Observation Augmentation}
\begin{figure}[ht]
\centering
\includegraphics[width=\textwidth]{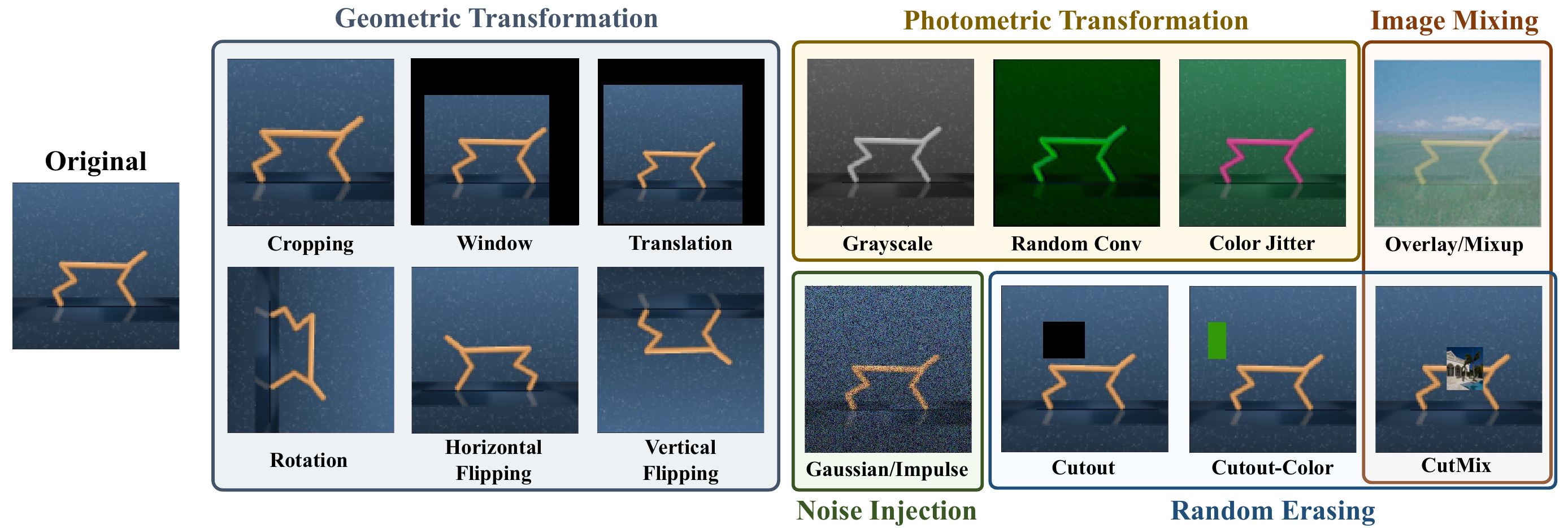}
\caption{Visualized examples of observation augmentation via classical image manipulations.} 
\label{aug} 
\end{figure}

A typical observation augmentation approach is to apply the classical image manipulations to the observed images; most such manipulations were originally proposed for computer vision applications.
Following the taxonomy of~\cite{image_aug_survey}, we identify five categories of image manipulations:
geometric transformations (Section~\ref{Geometric Transformation}),
photometric transformations (Section~\ref{Photometric Transformation}),
noise injections (Section~\ref{Noise Injection}),
random erasing (Section~\ref{Random Erasing})
and image mixing (Section~\ref{Mixing Images}).
Figure~\ref{aug} shows a list of the visualized examples.

\subsubsection{Geometric Transformations}
\label{Geometric Transformation}
Geometric transformations are generally employed as optimality-invariant or label-preserving transformations~\cite{image_aug_survey} to overcome data shortages during the training process.
{\textbf{Random cropping}} is an effective preprocessing technique for improving data efficiency; it works on image data with mixed width and height dimensions by locating a random central patch in each frame with a specific dimensionality~\cite{drq, drq_v2}.
In many visual RL scenarios, such as robotic manipulation tasks, the vital regions are often positioned at the centers of the images, and cropping can remove irrelevant edge pixels to simplify the learning process~\cite{rad}.
Similar to cropping, the {\textbf{window}} transformation selects a random region and masks out the cropped part of the image, while {\textbf{translation}} renders the image with a larger frame and randomly moves the image within that frame.

\vspace{0.5\baselineskip}
Other forms of geometric transformation have also been introduced in visual RL scenarios.
For example, {\textbf{rotation}} involves rotating an image right or left by $r$ degrees, where $r$ is randomly selected from a range~\cite{rad};
{\textbf{flipping}} obtains additional data by flipping the observations horizontally or vertically.
Although these techniques have been proven effective in computer vision tasks such as ImageNet~\cite{image_aug_survey}, visual RL tasks are sensitive to angle information.
In such a scenario, transformations such as rotation and flipping may produce erroneous results without properly adjusting the corresponding actions.

\subsubsection{Photometric Transformations}
\label{Photometric Transformation}
In real-world applications, the colors of objects and backgrounds may vary due to conditions such as lighting and weather~\cite{AIReview_image_augmentation}.
The intuition behind photometric transformations is to simulate these color variations to prevent overfitting on the training data~\cite{overfitting_1, overfitting_2}.
Overfitting in visual RL is especially problematic due to the spurious correlations between task-irrelevant features and the agent’s policy, which can severely damage its test performance~\cite{observational_overfitting}.
Based on the prior knowledge concerning the variations between the training and test environments, photometric transformations aim to better generalize the agent’s policy to unseen visual environments.
For instance, {\textbf{grayscale}} simply converts images from RGB to grayscale~\cite{rad}, while {\textbf{color jitter}} varies the features of images that are commonly used in DA, including brightness, contrast, and saturation~\cite{autoaugment}.
A common way to perform jitter in the color space is to convert images from RGB to HSV and add noise to the HSV channels~\cite{rad}.
Furthermore, \textbf{random convolution} has been introduced to mitigate visual bias that can negatively impact the performance of convolutional neural networks (CNNs)~\cite{randfm}. This technique augments image colors by passing the input observations through a randomly initialized single-layer convolutional network, which produces output layers of the same dimensions as the input.

\subsubsection{Noise Injection}
\label{Noise Injection}
Adding noise to images can help CNNs learn robust features in computer vision tasks~\cite{noise},
and recent studies~\cite{svea, secant} also attempted to exploit this mechanism in visual RL to obtain robust state representations.
In practice, distortion can be introduced by adding {\textbf{Gaussian}} noise~\cite{rad} or {\textbf{impulse}} (salt-and-pepper) noise~\cite{secant}.

\subsubsection{Image Mixing}
\label{Mixing Images}
This type of methods is commonly used in computer vision tasks to improve a model’s robustness and generalization ability~\cite{why_mixup}.
Among the different versions of mixing, {\textbf{Overlay/Mixup}}~\cite{mixup} trains a neural network on the convex combinations of samples and their labels.
In visual RL, there are two ways to leverage the Mixup mechanism.
First, we can combine two observations and their supervision signals, which will be discussed in Section~\ref{Transition Augmentation}.
Alternatively, we can mix RL observations and other images randomly sampled from another dataset while the supervision signals of the observations remain fixed.
For example, SECANT~\cite{secant} linearly blends an observation with a distracting image $I$ as $f(o)=\alpha o + (1-\alpha) I$, where $I$ is randomly sampled from the COCO~\cite{coco} image set.

\subsubsection{Random Erasing}
\label{Random Erasing}
As an analog of the dropout regularization, erasing prevents the network from overfitting by working in the input data space instead of the network structure space~\cite{Random_Erasing}.
{\textbf{Cutout}}~\cite{cutout} partially erases an image by randomly masking an $m \times n$ patch of the image.
Furthermore, {\textbf{Cutout-Color}} masks the patch with a random color.
As a combination of Cutout and Mixup, {\textbf{CutMix}}~\cite{cutmix} replaces the removed region with a patch from another image and the supervision signals of the original observation are preserved (in visual RL scenarios)~\cite{secant}. 

\subsubsection{Feature Space Augmentation}
\label{Feature space augmentation}
Instead of applying DA in the input space, another effective approach to augment observations is to perform transformations in the feature space~\cite{feature_space_augmentation}. 
The feature space, also known as the latent or embedding space, refers to an abstract domain that encodes meaningful internal representations from the original high-dimensional data. 

\vspace{0.5\baselineskip}
The first approach to applying feature space augmentation typically involves autoencoders, which map input images to the latent feature space and then reconstruct the images based on the augmented features in that space. Common augmentations in the latent space include Gaussian noise addition and linear interpolation~\cite{modals}, which can generate more diverse datasets than traditional transformations in many supervised tasks~\cite{latent_space_interpolation, Auto-encoder-based_generative}.
However, despite a few studies using autoencoders to design reconstruction-based auxiliary tasks to aid representation learning~\cite{sac_ae, imagined_goals}, the use of autoencoders for generating high-quality augmented data has yet to be fully explored in visual RL scenarios.

\vspace{0.5\baselineskip}
Another approach is to extract representations from the lower layers of a CNN and directly augment the latent data without reconstructing the high-dimensional images~\cite{image_aug_survey}. 
For instance, MixStyle~\cite{mixstyle} adopts style mixing in the bottom layers to simulate various visual styles~\cite{style_transfer}, and has demonstrated strong cross-domain generalization performance on benchmarks like CoinRun~\cite{CoinRun}.
More recently, CLOP~\cite{clop} was introduced as an innovative augmentation technique that swaps pixel positions in the feature maps after the deepest convolutional layer while maintaining channel consistency. 
Experimental results show that CLOP significantly improves generalization performance without requiring additional representation learning tasks, due to the high-level abstract features contained in the deepest neural network layer.

\subsection{Transition Augmentation}\label{Transition Augmentation}
\begin{figure}[ht] 
\centering 
\includegraphics[width=\textwidth]{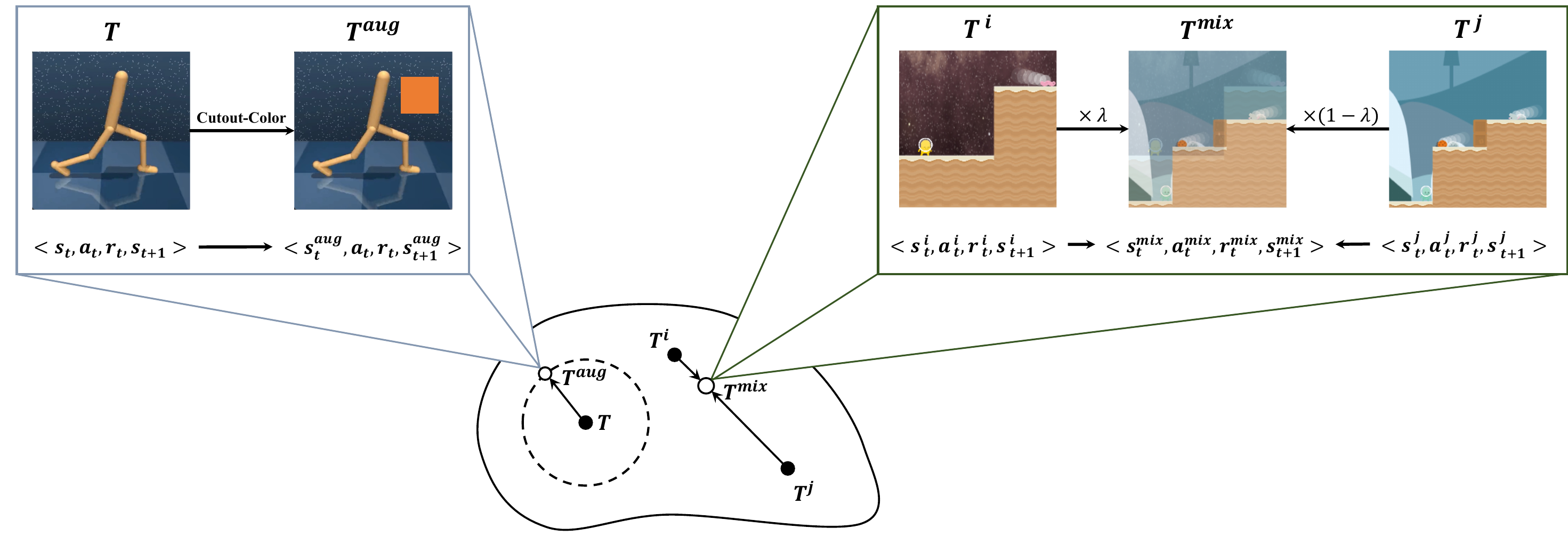} 
\caption{Comparison of observation augmentation and transition augmentation: observation augmentation via local perturbation (left, Cutout-Color~\cite{cutout}) and transition augmentation through joint interpolation of observations and supervision signals (right, MixReg~\cite{mixreg}).} 
\label{mixreg}
\end{figure}

As shown in Figure~\ref{mixreg}, augmenting $s_t$ with fixed supervision signals (e.g., the reward $r_t$ and action $a_t$) can be regarded as a form of local perturbation of the corresponding transition, representing a key example of observation augmentation discussed in Section~\ref{Observation Augmentation}. 
To ensure the validity of the augmented transition $<s_t^{aug}, a_t, r_t, s_{t+1}^{aug}>$, the augmented observation $s_t^{aug}$ must remain within a close range of the original observation $s_t$. 
Hence, local perturbation is inherently limited in terms of increasing data diversity, which is a common issue faced by all observation augmentations.

An intuitive solution is to apply interpolation across different data points instead of performing a local perturbation on each individual data point.
Inspired by Mixup~\cite{mixup} and CutMix~\cite{cutmix}, MixReg~\cite{mixreg} convexly combines two observations and their supervision signals to generate augmented data.
For example, let $y_i$ and $y_j$ denote the signals for states $s_i$ and $s_j$, respectively, which can be the reward or state values. 
After interpolating the observations by $\tilde{s}=\lambda s_{i}+(1-\lambda) s_{j}$, MixReg introduces mixture regularization in a similar manner via $\tilde{y}=\lambda y_{i}+(1-\lambda) y_{j}$, which helps learn more effective representations and smoother policies.

\subsection{Trajectory Augmentation}\label{Trajectory Augmentation}
Since observation or transition augmentation cannot directly enrich the trajectories encountered during training, to further improve the sample efficiency,
PlayVirtual~\cite{playvirtual} augments the actions to generate synthesized trajectories under a self-supervised cycle consistency constraint.

\begin{figure}[ht] 
\centering 
\includegraphics[width=\textwidth]{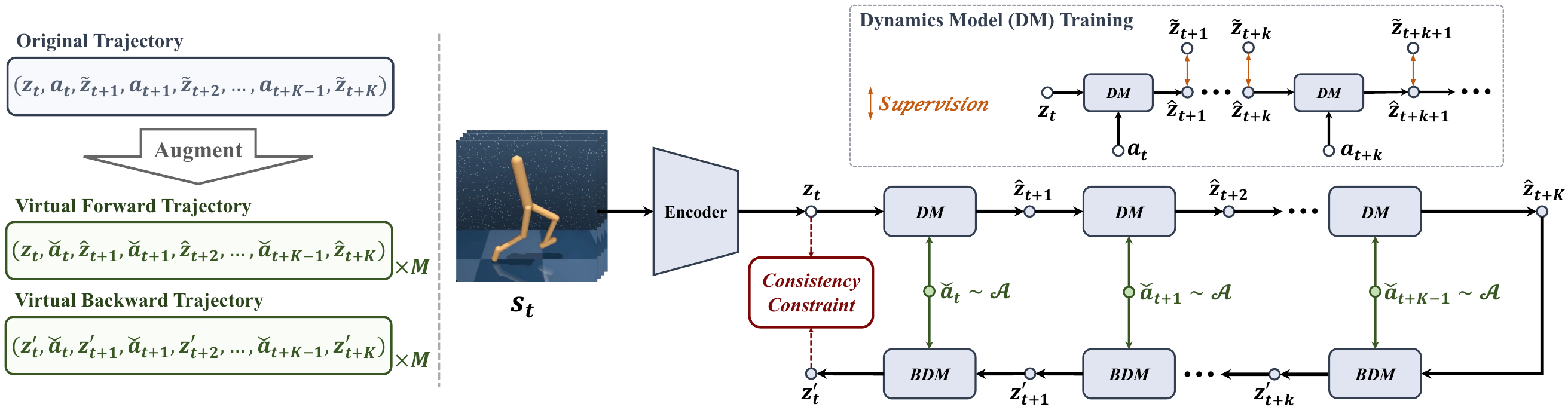} 
\caption{Data flow and architecture of PlayVirtual~\cite{playvirtual} as an example of trajectory augmentation.} 
\label{PlayVirtual}
\end{figure}

\vspace{0.5\baselineskip}
In Figure~\ref{PlayVirtual}, PlayVirtual operates entirely in the latent space after encoding the input observation $s_t$ into a low-dimensional state representation $z_t$.
Following the dynamics model (DM) in SPR~\cite{spr}, PlayVirtual introduces a backward dynamics model (BDM) to predict the backward transition dynamics $(z_{t+1},a_t) \longrightarrow z_t$ to build a loop with the forward trajectory.
During the training process, the DM is supervised by the original trajectory information, whereas the BDM is constrained by the cycle consistency between $z_t$ and $z^{\prime}_t$.
Further discussion on how to train the dynamics models with the auxiliary loss will be provided in Section~\ref{Policy Regularization Explicitly with Auxiliary Loss}.
After obtaining the effective DM and BDM, PlayVirtual can generate diverse synthesized trajectories by randomly sampling/augmenting $M$ sets of actions in the action space $\mathcal{A}$ and then calculating the state information.
Experimental studies confirmed that regularizing feature representation learning with cycle-consistent synthesized trajectories is the key to PlayVirtual's success.

\subsection{Automatic Augmentation}\label{Automatic Selection of Data Augmentation}

Automatic augmentation is receiving increasing attention due to the demand for task-specific augmentations~\cite{auto1,auto2,autoaugment}.
For example, although random cropping is one of the most effective augmentation techniques for improving sample efficiency on many benchmarks, such as DMControl-500k~\cite{rad,drq} and Procgen~\cite{drac}, the induced generalization ability improvement heavily depends on the specific choice of augmentation strategy.
In general, different tasks benefit from different augmentations, and selecting the most appropriate DA method often requires expert knowledge. Consequently, it is crucial to develop methods that can automatically identify the most effective augmentation techniques. Research in visual RL remains in its early stages~\cite{drac}, and we highlight some promising approaches below.

\paragraph{{Upper Confidence Bound (UCB):}}
The task of selecting an appropriate augmentation from a given set can be formulated as a multi-armed bandit problem where the action space is the set of available transformations $F=\{f_1,f_2,\dots,f_n\}$.
The UCB~\cite{ucb} is a popular solution for the multi-armed bandit problem that considers both exploration and exploitation.
Recently, UCB-DrAC~\cite{drac} and UCB-RAD~\cite{UCB-RAD} were proposed to achieve automatic augmentation in visual RL.
The experiment results suggest that UCB-based automatic augmentations can effectively improve the agent's generalization capabilities.

\paragraph{{Meta Learning:}}
Meta learning offers an alternative solution to automatic augmentation and can be implemented in two ways~\cite{drac}:
$(1)$ training a meta learner, such as RL$^2$\cite{rl2}, to automatically select an augmentation type before each update in a DA-based algorithm;
$(2)$ meta-learning the weights of a CNN to perturb observed images, a technique similar to model-agnostic meta learning (MAML)\cite{maml,meta}.
In practice, neither approach has yielded promising results, and designing expressive functions for automatic augmentation via meta learning remains a challenge.

\subsection{Context-Aware Augmentation}
\label{Context-aware Data Augmentation}


\begin{wrapfigure}{l}{0.47\textwidth}
\centering
\includegraphics[width=0.468\textwidth]{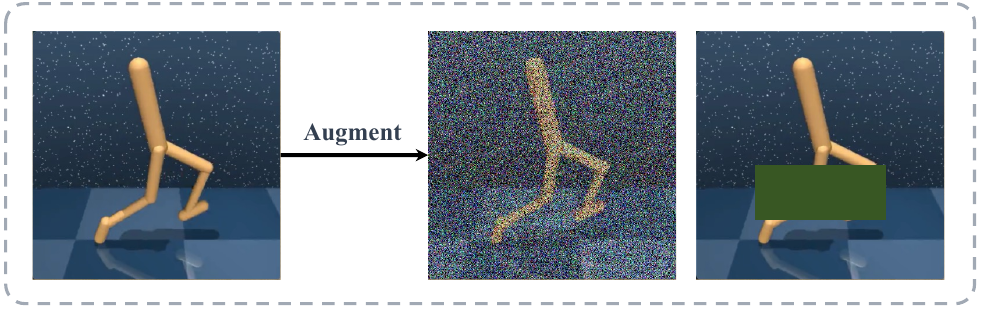}
\caption{Context-agnostic DA may distort critical information in observations.}
\label{Context-agnostic Augmentation}
\end{wrapfigure}
A notable limitation of existing DA techniques is their reliance on pixel-level image transformations that process each pixel without considering its contextual significance~\cite{tlda}.
In the context of visual RL, however, pixels within an observation typically exhibit differential relevance to the decision-making process~\cite{Visualizing_and_understanding, attention_augmented_agents}. 
As illustrated in Figure~\ref{Context-agnostic Augmentation}, context-agnostic augmentation techniques may inadvertently mask or alter critical regions in the original observation that are essential for decision making.
This disregard for context elucidates why the straightforward application of prior-based strong augmentation, despite its potential to improve generalization, can significantly impair both sample efficiency and training stability in visual RL~\cite{svea,tlda}.
This context-agnostic property explains why naively applying prior-based strong augmentation may severely damage both the sample efficiency and the training stability of visual RL, despite their potential to improve generalization~\cite{svea,tlda}.
Consequently, incorporating context awareness into augmentation techniques is essential for enhancing the effectiveness of DA while minimizing its potential drawbacks. Currently, two viable approaches have been proposed to advance context-aware augmentation:

\begin{enumerate}
\item \textbf{Introducing human guidance.}
Human-in-the-loop RL (HIRL)~\cite{HIRL} is a general paradigm that leverages human guidance to assist the RL process.
EXPAND~\cite{expand} introduces a human saliency map to mark the importance levels of different regions, and it only perturbs the irrelevant regions.
Saliency maps contain human domain knowledge, allowing context information to be embedded into the augmentation.
\item \textbf{Excavating task relevance.}
In visual RL, the contextual information can be extracted from the task relevance of each pixel, making it possible to directly determine its task relevance to achieve context-aware augmentation.
Task-aware Lipschitz DA (TLDA)~\cite{tlda} explicitly defines the task relevance by computing the Lipschitz constants produced when perturbing corresponding pixels.
Regions with large Lipschitz constants are crucial for the current task decision, and these regions will subsequently be protected from augmentation.
\end{enumerate}

Context-aware augmentation forms the foundation for semantic-level DA, which aims to apply targeted operations to different semantic contexts within observations~\cite{keepaugment, tlda}. In Section~\ref{Semantic-level Data Augmentation}, we will further discuss semantic-level DA as a challenging yet pivotal direction for future research.

\subsection{Generative Augmentation}
\label{Generative Augmentation}

Despite the remarkable success of leveraging generative models for data augmentation in computer vision tasks~\cite{gan-based_data_augmentation, GAN_based_data_augmentation2, yang2023diffusion, GANDA_1, GANDA_2}, the application of VAEs or GANs to generate synthetic data for reinforcement learning has not only failed to achieve comparable performance but may even lead to detrimental effects~\cite{imre2021investigation, zhu2023diffusion}.
This limitation in visual RL remained unresolved until the recent emergence of diffusion models. 
A series of successful works leveraging diffusion models to generate high-quality synthetic data for visual RL have gradually emerged, marking a significant advancement in this field.

\begin{enumerate}
\item \textbf{Generative Augmentation for Observations.}
ROSIE~\cite{yu2023scaling} and GenAug~\cite{GenAug} leverage text-guided diffusion models to augment observations in robotic control tasks while preserving the corresponding actions, representing an advanced approach to observation augmentation using generative models.
Trained on massive online datasets, the diffusion models employed for DA can zero-shot create realistic images of many different objects and scenes.
This capability significantly expands the diversity of augmented observations beyond what is achievable with traditional DA transformations.
\item \textbf{Generative Augmentation for Transitions.}
In contrast to approaches that solely generate observations, another line of research focuses on modeling the entire transition, simultaneously synthesizing novel action and corresponding reward labels.
Within this paradigm, SynthER~\cite{lu2024synthetic} directly trains diffusion models using either offline datasets or online replay buffers, subsequently generating samples for policy improvement. Advancing this concept further, MTDIFF~\cite{he2024diffusion} transcends the limitations of single-task scenarios by leveraging diffusion models to consolidate knowledge from multi-task datasets and augment data for novel tasks.
The success of SynthER~\cite{lu2024synthetic} and MTDIFF~\cite{he2024diffusion} demonstrates the significant potential of leveraging synthetic data to enhance visual RL performance.
\end{enumerate}

Overall, recent studies and analyses indicate that data generated by diffusion models surpasses that of traditional DA methods in both diversity and accuracy~\cite{lu2024synthetic}. 
This clearly demonstrates the capability of advanced generative models to produce novel, diverse, and dynamically accurate data. 
Such high-quality synthetic data can be effectively utilized by policies to enhance both the sample efficiency and generalization ability of visual RL algorithms.
Furthermore, the text-controllable nature of current generative models enables them to serve as effective tools for semantically meaningful samples~\cite{GenAug}. 
This capability holds promise for achieving genuine semantic-level manipulation of training data, presenting a crucial direction for future research.

\subsection{Remarks}

Data augmentation, as a data-centric approach, has demonstrated remarkable success in visual RL tasks, significantly enhancing both sample efficiency and generalization ability. 
This section provides a comprehensive review of various approaches addressing "How to augment data in visual RL". 
The key main takeaways can be summarized as follows:

\begin{enumerate}
\item In contrast to DA in supervised learning scenarios, visual RL tasks offer a wider varying range of manipulable data types~\cite{mixreg, playvirtual}.
Among these, observation augmentation has gained the most widespread application due to its ease of implementation~\cite{drq_v2}. 
However, with the advancement of generative models, more complex yet diverse transition and trajectory augmentation techniques show potential for achieving novel breakthroughs.
\item To date, the most effective and widely adopted DA approaches for enhancing sample efficiency in visual RL are considered to be spatial perturbations and minor scaling of observations~\cite{Learning_Better_with_Less}. 
This finding notably diverges from conclusions in other domains, primarily due to the distinct underlying mechanisms through which DA significantly improves sample efficiency in visual RL~\cite{ma2023revisiting}. 
We will discuss these mechanisms in depth in Section~\ref{Theoretic Frameworks for Data Augmentation}.
\item The robust generalization ability of visual RL agents during deployment largely depends on the diversity of training data. 
Consequently, both traditional strong augmentation techniques such as Color Jitter and advanced generative augmentation methods need to provide sufficient data richness to narrow the generalization gap~\cite{he2024diffusion}. 
In this context, ensuring training stability and maintaining the consistency of augmented data with environment dynamics become crucial considerations~\cite{tlda, svea}.
\item The key to advancing DA in visual RL lies in automatically generating optimal and context-aware augmented data. Recent advancements in generative AI have opened up new possibilities for leveraging pre-trained generative models to produce novel data that is not only rich and diverse but also maintains the constraints of RL dynamics.
\end{enumerate}

%% file: textfile/4_How_to_Leverage_Augmented_Data_in_Visual_RL.tex
\newpage
\section{How to Leverage Augmented Data in Visual RL?}\label{How to leverage}

Next, we discuss how to exploit the augmented data in visual RL.
To ease the discussion, we divide the application scenarios where DA plays a vital role into three cases.

\begin{figure}[ht] 
\centering 
\includegraphics[width=\textwidth]{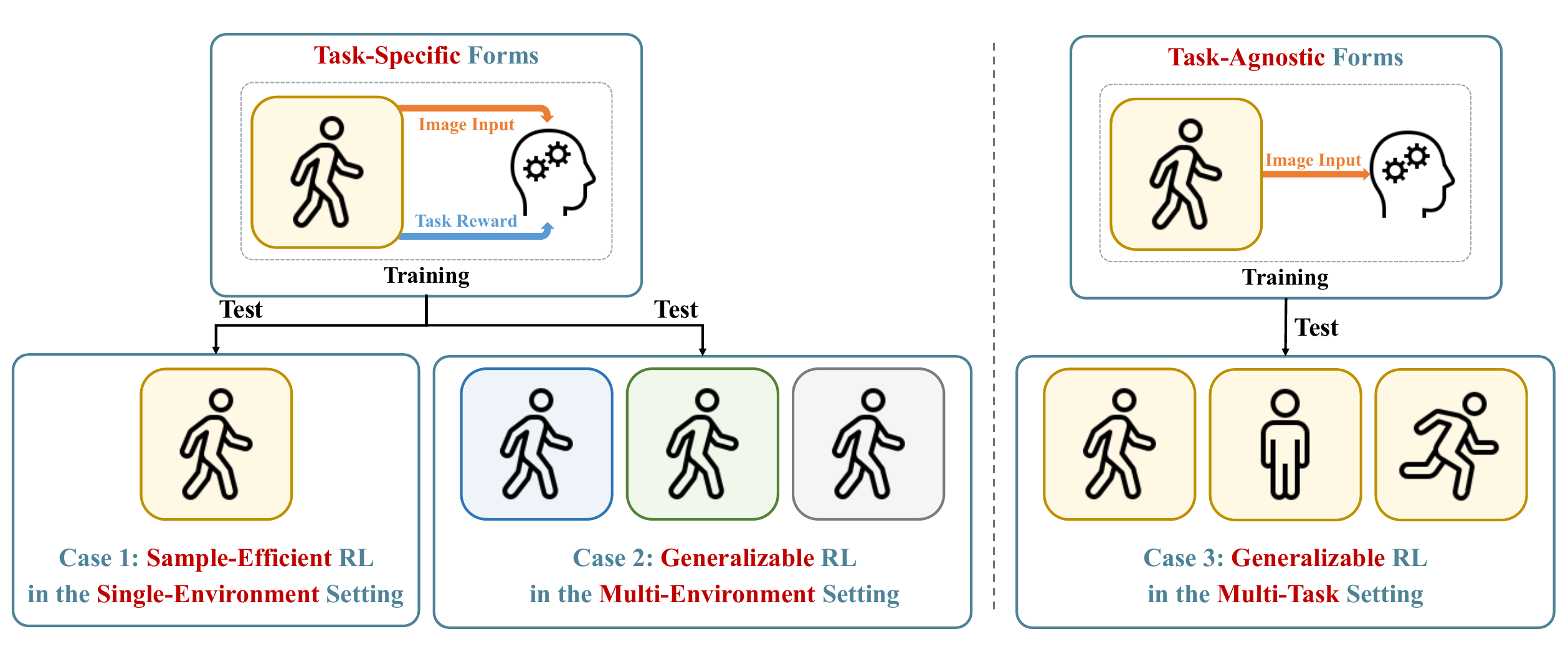} 
\vspace{-\baselineskip}
\caption{Three representative scenarios highlighting the critical role of DA. In the context of single-task environments, DA enhances sample efficiency during training and improves generalization ability during deployment. Furthermore, DA contributes to training task-agnostic representations, facilitating superior generalization and adaptation across multiple tasks.} 
\label{Overview of Section4} 
\end{figure}

\vspace{-0.5\baselineskip}
\begin{enumerate}
\item[{Case} 1] {\textbf{Sample-efficient RL in the single-environment setting.} }
Agents are trained and evaluated within a fixed environment, commonly referred to as the single-environment setting~\cite{sac_ae}.
The primary objective is to attain satisfactory performance with minimal interactions within the environment~\cite{drq, drq_v2, Learning_Better_with_Less}.
\item[{Case} 2] {\textbf{Generalizable RL in the multi-environment setting.}}
Agents are tested in unseen environments after interacting with the training environments~\cite{secant}.
Since RL agents tend to overfit the training environment~\cite{study_on_overfitting}, generalizing the learned policies to unseen environments remains challenging even when only visual appearances are altered~\cite{svea, SADA}.
\item[{Case} 3] {\textbf{Generalizable RL in the multi-task setting.}}
Agents in the multi-task setting aim to adapt to different tasks.
Traditional end-to-end RL algorithms heavily rely on task-specific rewards, making them unsuitable for other tasks~\cite{atc}.
Recent studies have attempted to address this limitation by pretraining cross-task representations in a task-agnostic manner, thereby enabling agents to swiftly adapt to multiple downstream tasks~\cite{urlb}.
\end{enumerate}

In Figure~\ref{Overview of Section4}, RL agents are trained with task-specific rewards in Case $1$ and Case $2$, where DA is implemented as an implicit regularization penalty when enlarging the training set (Section~\ref{Policy Regularization Implicitly}).
However, the effect of implicit regularization is limited~\cite{spr}, and many studies have attempted to design auxiliary losses to exploit the potential of DA (Section~\ref{Policy Regularization Explicitly with Auxiliary Loss}).
Some studies have also aimed to decouple representation learning from policy optimization to attain more generalizable policies~\cite{secant} (Section~\ref{Task-Specific Representation Decoupled from Policy Optimization}).
Finally, the related works belonging to Case 3, referred to as task-agnostic representation approaches using unsupervised learning, are introduced in Section~\ref{Task-Agnostic Representation using Unsupervised Learning}.

\subsection{Implicit Policy Regularization}
\label{Policy Regularization Implicitly}
DNNs are capable of learning complex representational spaces, which is essential for tackling intricate learning tasks.
However, the model capacity required to capture such high-dimensional representations makes these techniques difficult to optimize and prone to overfitting~\cite{survey_of_regularization}.
Moreover, the complexity of visual RL is further aggravated by the need to jointly learn representations and policies directly from high-dimensional observations based on sparse reward signals~\cite{sac_ae, rad}.
As a result, it is difficult for agents to distinguish the task-relevant (reward-relevant) features from high-dimensional observations, and they may mistakenly correlate rewards with spurious features~\cite{observational_overfitting}.
To solve these issues, researchers have conducted a series of studies to develop effective regularization techniques, which can prevent overfitting and improve generalization by incorporating the inductive biases of model parameters~\cite{survey_of_regularization}.

\vspace{0.5\baselineskip}
In RL, a myriad of techniques have been proposed as regularizers such as $L^p$-norm regularization~\cite{ddpg}, batch normalization~\cite{bn_rl}, weight decay~\cite{CoinRun} and dropout~\cite{SNI_IB}.
Among them, $L^p$-norm regularization explicitly includes regularization terms as additional constraints, and is referred to as
explicit regularization~\cite{observational_overfitting}.
Conversely, weight decay and dropout aim to tune the optimization process without affecting the loss function, making them implicit regularization strategies~\cite{SNI_IB}.
Additionally, DA has been prevalent in the deep learning community as a data-driven technique~\cite{image_aug_survey, instead_of_explicit_regularization}.
Furthermore, increasing efforts have been devoted to the theoretical underpinnings behind DA~\cite{A_Thousand_Samples, 0224_dac, Feature_Manipulation, 0308_Regularising_for_invariance, rethinkingDAinCL} to explain its regularization effects, including the derivation of an explicit regularizer to simulate the behaviors of DA~\cite{A_Thousand_Samples}.

\vspace{0.5\baselineskip}
The initial and naive practice of DA is to expand the training set with augmented (synthesized) samples~\cite{content_from_style}.
This practice incorporates prior-based human knowledge into the data instead of designing explicit penalty terms or modifying the optimization procedure.
Hence, it is often classified as a type of implicit regularization, formulated as the empirical risk minimization on augmented data (DA-ERM)~\cite{0224_dac} in SL tasks:
\begin{equation}
\widehat{h}^{d a-e r m} \triangleq \underset{h \in \mathcal{H}}{\operatorname{argmin}} \sum_{i=1}^{N} l\left(h\left(\mathbf{x}_{i}\right), y_{i}\right)+\sum_{i=1}^{N} \sum_{j=1}^{\alpha} l\left(h\left(\mathbf{x}_{i, j}\right), y_{i}\right)
\end{equation}
where $\left(\mathbf{x}{i}, y_i\right)$ represents the $i^{\text{th}}$ original training sample ($\mathbf{x}{i} \in \mathcal{X}$ denotes the input feature, and $y_{i} \in \mathcal{Y}$ is its corresponding label);
$\mathbf{x}{i,j}$ signifies the $j^{\text{th}}$ augmented sample of $\mathbf{x}{i}$, which retains the corresponding label $y_{i}$;
$\alpha$ indicates the number of augmentations;
$l: \mathcal{Y} \times \mathcal{Y} \rightarrow \mathbb{R}$ is the loss function, and $h(\cdot)$ is the model to be optimized.

\vspace{0.5\baselineskip}
In the visual RL community, RAD~\cite{rad} and DrQ~\cite{drq} first leverage classical image transformation strategies such as cropping to augment the input observations via the implicit regularization paradigm.
In the original paper, DrQ is proposed with two distinct ways to regularize the Q-function.
On the one hand, it uses $K$ augmented observations from the original $s_i^{\prime}$ to obtain the target values for each transition tuple $(s_i,a_i,r_i,s_i^{\prime})$: 
\begin{equation}
y_{i}=r_{i}+\gamma \frac{1}{K} \sum_{k=1}^{K} Q_{\theta}(f(s_{i}^{\prime}, \nu_{i, k}^{\prime}), a_{i, k}^{\prime}), a_{i, k}^{\prime} \sim \pi(\cdot \mid f(s_{i}^{\prime}, \nu_{i, k}^{\prime}))
\label{drq1}
\end{equation}
where $f: \mathcal{S} \times \mathcal{T} \rightarrow \mathcal{S}$ is the augmentation function and $\nu$ is the parameter of $f(\cdot)$, which is randomly sampled from the set of all possible parameters $\mathcal{T}$.
Alternatively, DrQ generates $M$ different augmentations of $s_i$ to estimate the Q-function:
\begin{equation}
J_{Q}^{\mathrm{DrQ}}(\theta) = \frac{1}{N M} \sum_{i=1, m=1}^{N, M} ||Q_{\theta}(f(s_{i}, \nu_{i, m}), a_{i})-y_{i}||^{2}_2
\label{drq2}
\end{equation}

In the above, DrQ leverages DA for improved estimation without adding any penalty terms, which is a type of data-driven implicit regularization.
Since a sample can be defined as a tuple $(\mathbf{x}_{i},y_i)$ in SL or a transition $(s_i,a_i,r_i,s_i^{\prime})$ in RL, the optimization objective of DrQ can be rewritten as:
\begin{equation}
J_{Q}^{\mathrm{DrQ}}(\theta) = \frac{1}{N M K} \sum_{i=1}^{N}\sum_{m=1}^{M}\sum_{k=1}^{K} l(f(s_{i}, \nu_{i, m}) ,a_i,r_i, f(s_{i}^{\prime}, \nu_{i, k}^{\prime}))
\label{drq3}
\end{equation}
where $l(s_i,a_i,r_i,s_i^{\prime}) = ||Q_{\theta}(s_i,a_i) - (r_i+\gamma Q_{\theta}(s_i^{\prime},a_i^{\prime}))||^2_2$ is the loss function, and $a_{i}^{\prime} \sim \pi(\cdot \mid s_{i}^{\prime})$.
RAD~\cite{rad} can be regarded as a specific form of DrQ with $K=1$ and $M=1$; it is a plug-and-play module that can be plugged into any RL method (on-policy methods such as PPO~\cite{ppo} and off-policy methods such as SAC~\cite{sac}) without making any changes to the underlying algorithm.
RAD has also highlighted the generalization benefits of DA on OpenAI Procgen~\cite{procgen}.

\vspace{0.5\baselineskip}
Since RAD and DrQ directly optimize the RL objective on multiple augmented observation views without any auxiliary losses, they can be viewed as implicit approaches for ensuring consistency and invariance among the augmented views.
Building on DrQ, DrQ-v2~\cite{drq_v2} makes several algorithmic adjustments, such as switching the baseline from SAC to DDPG and employing a larger replay buffer, which has resulted in significantly improved sample efficiency. The success of DrQ-v2 demonstrates that when utilizing weak augmentation to achieve sample-efficient visual RL algorithms, the method of implicit regularization can effectively harness the benefits of DA.

\begin{figure}[ht]
    \centering
    \includegraphics[width=\textwidth]{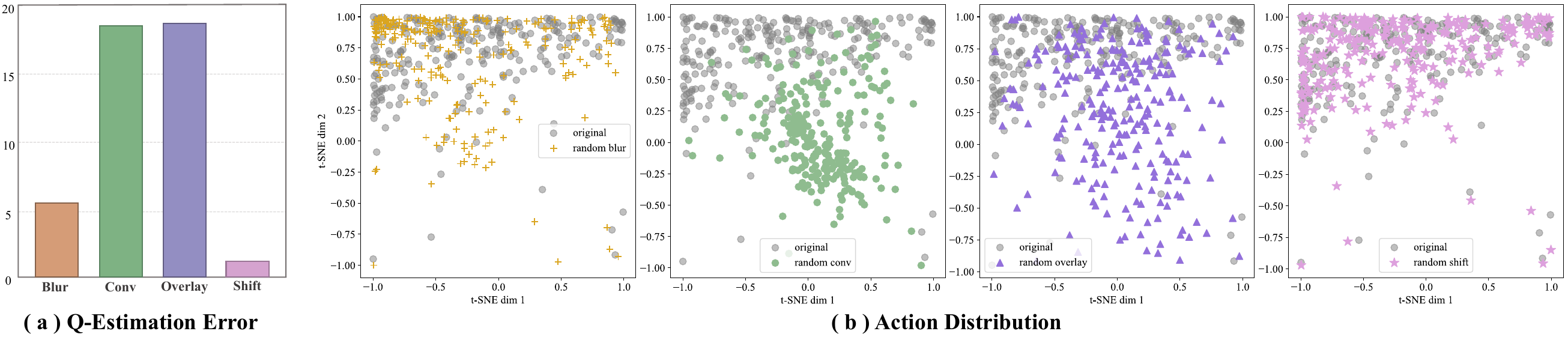}
    \caption{\textbf{$Q$-estimation errors and action distributions for augmented and original data.}
    (a) Mean absolute $Q$-estimation errors of the converged DrQ~\cite{drq} agents for the same observations before and after augmentation (copied from~\cite{svea}).
    (b) Action distributions between the augmented and original data.
    We use t-distributed stochastic neighbor embedding (t-SNE) to show the high-dimensional actions employed by the same converged DrQ agent.}
    \label{action and q} 
\end{figure}

However, later studies found that implicit regularization with cropping exhibits poor generalization performance in unseen environments~\cite{drac,svea}.
As discussed in Section~\ref{Data Augmentation in Visual RL},
optimality-invariant transformations (represented by cropping) cannot provide sufficient visual diversity for reducing the generalization gap.
Furthermore, although prior-based strong augmentations such as color jitter have the potential to improve generalization, they may induce large $Q$-estimation errors and action distribution shifts, as shown in Figure~\ref{action and q}.
Hence, implicit regularization approaches with prior-based strong augmentations (e.g., random convolution and overlay) may make the RL optimization process fragile and unstable~\cite{svea, tlda}.
This poses a \textit{dilemma} in visual RL:
diverse augmentation is necessary to improve an agent’s generalization ability, but excessive data variations may damage the stability of RL~\cite{secant}.

\vspace{0.5\baselineskip}
SVEA~\cite{svea} aims to enhance the stability of RL optimization with DA~\cite{drq}.
It consists of two main components.
First, SVEA uses only original data copies to estimate $Q$-targets to avoid erroneous bootstrapping caused by DA, where $y_{i}=r_{i}+\gamma  Q_{\theta}(s_{i}^{\prime}, a_{i}^{\prime})$, $a_{i}^{\prime} \sim \pi(\cdot \mid s_{i}^{\prime})$.
Second, SVEA formulates a modified $Q$-objective to estimate the $Q$-value over both augmented and original copies of the observations, which can be expressed in a modified ERM form as follows:
\begin{equation}
\begin{aligned}
J_{Q}^{\mathrm{SVEA}}(\theta) &= \alpha \sum_{i=1}^{N} ||Q_{\theta}(s_{i}, a_{i})-y_{i}||^{2}_2 + \beta  \sum_{i=1}^{N} \sum_{m=1}^{M}  ||Q_{\theta}(f(s_{i}, \nu_{i, m}), a_{i})-y_{i}||^{2}_2\\
&= \alpha \sum_{i=1}^{N} l(s_{i},a_i,r_i, s_{i}^{\prime}) + \beta \sum_{i=1}^{N} \sum_{m=1}^{M} l(f(s_{i}, \nu_{i, m}) ,a_i,r_i, s_{i}^{\prime})
\end{aligned}
\label{f_svea}
\end{equation}

\begin{figure}[b!]
    \centering
    \includegraphics[width=0.75\textwidth]{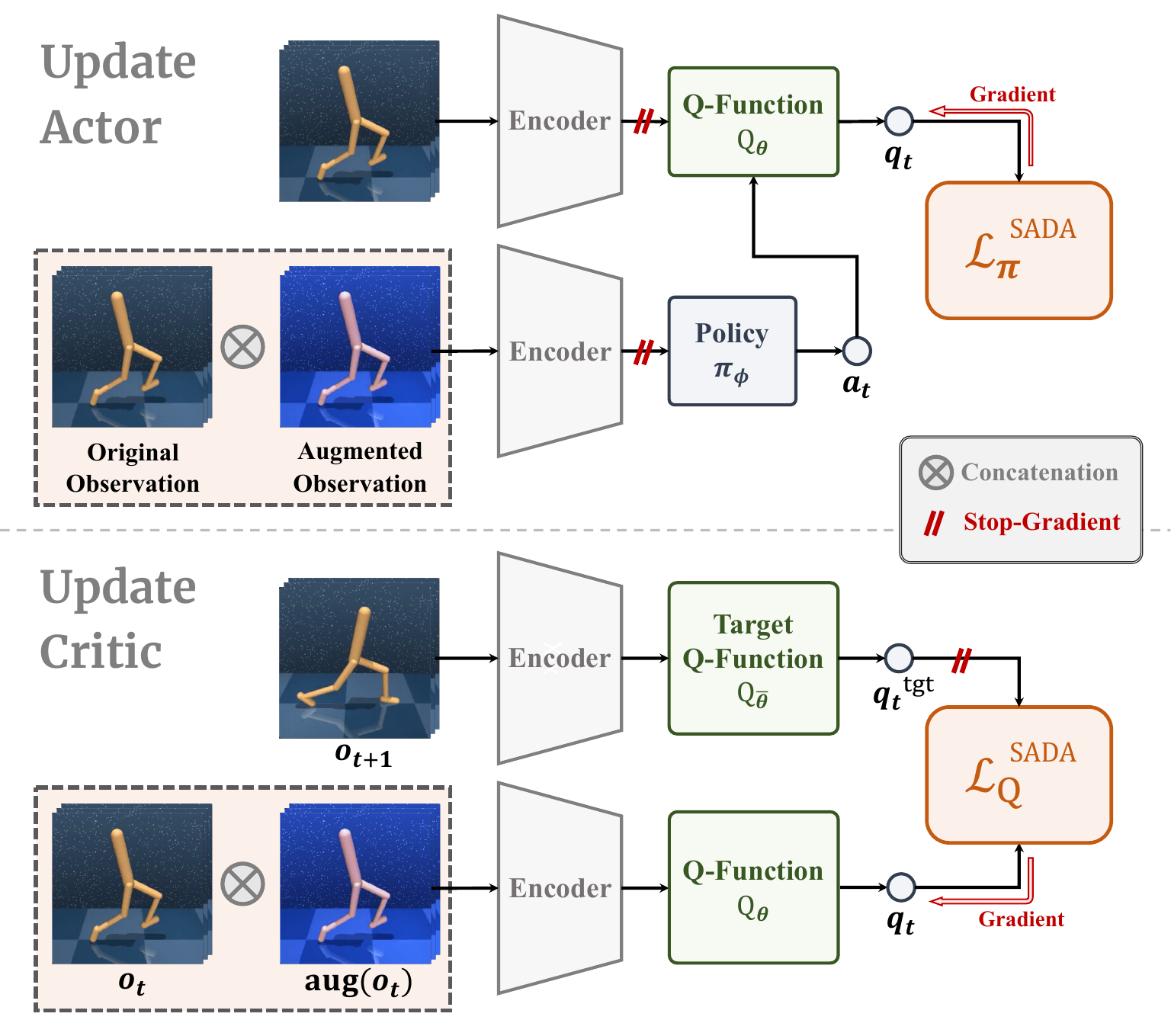}
    \caption{The workflow of \textbf{S}tabilized \textbf{A}ctor-Critic under \textbf{D}ata \textbf{A}ugmentation (SADA).}
    \label{SADA} 
\end{figure}

For actor-critic algorithms, SVEA employs strong augmentation exclusively during critic updates, with no augmentation applied during actor updates. SVEA assumes that the encoder’s output embedding can become fully invariant to input augmentations. Under this assumption, an actor trained solely on unaugmented observations can indirectly achieve robustness to augmented inputs via a shared actor-critic encoder.
While this assumption holds for scenarios where the differences between test and training environments are limited to photometric changes, it fails when geometric augmentations are necessary for more complex generalization tasks. 

\vspace{0.5\baselineskip}
To overcome this limitation, SADA~\cite{SADA} enhances the use of DA as implicit regularization to accommodate a broader range of augmentations. Instead of augmenting only the critic inputs, SADA carefully augments both actor and critic inputs to prevent training instabilities. As shown in the Figure~\ref{SADA}, (1) during actor updates, only the policy input is augmented while the Q-function input remains unaugmented; (2) during critic updates, only the online Q-function input is augmented while the target Q-function input remains unaugmented; and (3) components are jointly optimized using both augmented and unaugmented data.

\vspace{0.5\baselineskip}
SVEA~\cite{svea} and SADA~\cite{SADA} have significantly enhanced the generalization ability of visual RL algorithms. They achieve this by carefully designing pipelines that use strong augmentations without adding auxiliary losses or altering RL update mechanisms. This demonstrates the potential of implicit regularization to leverage DA for better generalization. However, it is crucial to meticulously design the use of strong augmentations to avoid training instabilities. In addition to carefully designing implicit regularization, other approaches have been explored to harness DA for improving both sample efficiency and generalization while avoiding adverse effects on training stability. These approaches include designing auxiliary tasks (Section~\ref{Policy Regularization Explicitly with Auxiliary Loss}) and decoupling representation learning from policy optimization (Section~\ref{Task-Specific Representation Decoupled from Policy Optimization}).

\subsection{Explicit Policy Regularization with Auxiliary Tasks}
\label{Policy Regularization Explicitly with Auxiliary Loss}
Visual RL relies on the state representation, but it remains challenging to directly infer the ideal representation from high-dimensional observations~\cite{aarl}.
A typical workflow involves designing auxiliary objectives to facilitate the representation learning process~\cite{bellemare2019geometric}, or improve sample efficiency~\cite{curl} or prevent observational overfitting~\cite{observational_overfitting}.
In general, an auxiliary task can be considered an additional cost function that the RL agent predicts and observes from the environment in a self-supervised manner~\cite{auxiliary_tasks_in_rl}. For example, the last layer of the network can be divided into multiple parts (heads), with each head dedicated to a specific task~\cite{cody, does_ssl}. These multiple heads then propagate errors back to the shared network layers, thereby forming the comprehensive representations required by all heads.

\vspace{0.5\baselineskip}
With the recent success in unsupervised learning, various auxiliary tasks have been designed to produce effective representations~\cite{auxiliary_tasks_in_rl, he2022reinforcement}.
Thus, it is natural to design additional losses to explicitly constrain an agent’s policy and value functions, which we will discuss in Section~\ref{Data Augmentation Consistency}.
Moreover, we introduce contrastive learning as a lower bound of mutual information in Section~\ref{Contrastive Learning} and future prediction objectives with a DM in Section~\ref{Future Prediction with Dynamics Model}.

\subsubsection{DA Consistency}
\label{Data Augmentation Consistency}
In contrast to simply inserting augmented data into the training dataset, DA consistency (DAC)~\cite{0224_dac} builds a regularization term to penalize the representation difference between the original sample $\phi_{h}\left(\mathbf{x}_{i}\right)$ and augmented sample $\phi_{h}\left(\mathbf{x}_{i, j}\right)$, under the assumption that similar samples should be close in the representation space:
\begin{equation}
\underset{h \in \mathcal{H}}{\operatorname{argmin}} \sum_{i=1}^{N} l\left(h\left(\mathbf{x}_{i}\right), y_{i}\right)+\underbrace{\lambda \sum_{i=1}^{N} \sum_{j=1}^{\alpha} \varrho\left(\phi_{h}\left(\mathbf{x}_{i}\right), \phi_{h}\left(\mathbf{x}_{i, j}\right)\right)}_{\text {DAC regularization }} 
\label{dac}
\end{equation}
where $\phi_{h}$ refers to the features extracted from the high-dimensional data, which can be viewed as the output of any layer in the DNN, and $\varrho$ is the metric function defined in the representation space, which can be the $\mathcal{L}_p$ norm or KL divergence.
As an unsupervised representation module, DAC regularization can be employed as an auxiliary task in any SL or RL algorithms to enforce the model to produce similar predictions on the original and augmented samples.
For example, SODA~\cite{soda} calculates the consistency loss by minimizing the $L^2$ norm between the features of the augmented and original observations in the latent space;
SIM~\cite{sim} produces a cross-correlation matrix between two embedding vector sets of the original and augmented observations, and designs an invariance loss term to ensure the invariance of data.

\vspace{0.5\baselineskip}
For RL tasks, it is also desirable to train the network to output the same policies and values for both original and augmented observations~\cite{drq}.
For example, DrAC~\cite{drac} employs two extra loss terms: $G_{\pi}$ for regularizing the policy by the KL divergence measure and $G_V$ for regularizing the value function using the mean-squared deviation:
\begin{equation}
G_{\pi}=K L\left[\pi_{\theta}(a \mid s) \mid \pi_{\theta}(a \mid f(s, \nu))\right],
\qquad
G_{V}=\left(V_{\phi}(s)-V_{\phi}(f(s, \nu))\right)^{2}
\label{drac1}
\end{equation}

The complete optimization objective of DrAC based on PPO is as follows:
\begin{equation}
J_{\mathrm{DrAC}}=J_{\mathrm{PPO}}-\alpha_{r}\left(G_{\pi}+G_{V}\right)
\label{drac3}
\end{equation}
where $\alpha_{r}$ is the weight of the regularization term, and both $G_{\pi}$ and $G_V$ can be added to the objective of any actor-critic algorithm.
By enforcing the DA consistency into the networks, specific transformations can be used to impose inductive biases relevant to the given task (e.g., invariance with respect to colors or translations)~\cite{drac, 0224_dac}.

\vspace{0.5\baselineskip}
Compared with implicit regularization techniques such as RAD and DrQ, DrAC employs two auxiliary consistency loss terms for explicitly regularizing the policy and the value function to ensure invariance.
Instead of directly optimizing the RL objective on multiple augmented views
of the observations, DAC regularization uses only the transformed observations $f(s, \nu)$ to compute the regularization losses $G_{\pi}$ and $G_V$.
Hence, DrAC can benefit from the regularizing effect of DA while mitigating the adverse effect on the RL objective~\cite{drac}.

\subsubsection{Contrastive Learning}
\label{Contrastive Learning}
Another type of auxiliary task closely related to DA is contrastive learning. 
As \textbf{mutual information (MI)} is often hard to estimate, it is practical to maximize the lower bound of MI through approaches using, for example, InfoNCE loss~\cite{cpc}) to train robust feature extractors~\cite{MI_maximization}.
Recent studies~\cite{curl,atc} have shown that contrastive learning can significantly improve the sample efficiency and generalization performance of visual RL~\cite{cody}.
Since contrastive learning only requires unlabeled data, it can not only be performed as auxiliary tasks together with RL objectives but also be leveraged to learn a task-agnostic representation, which we will discuss in Section~\ref{Task-Agnostic Representation using Unsupervised Learning}.

\vspace{0.5\baselineskip}
In visual RL, there are two types of contrastive learning for improving agents’ sample efficiency and generalization abilities~\cite{cody}.
The first class~\cite{curl,dribo,ADAT} focuses on \textbf{{maximizing the MI between different augmented versions of the same observation}} while minimizing the similarity between different observations.
It tends to further exploit the regularization ability of DA at the MI level~\cite{cody}.
However, simply maximizing the lower bound of MI may retain the task-irrelevant information~\cite{multi_view_information_bottleneck}, which needs to be eliminated based on the information bottleneck principle.
The second class~\cite{cpc,atc} aims to \textbf{{maximize the predictive MI between consecutive states}} by applying contrastive losses between an observation $o_t$ and the near-future observations $o_{t+k}$ over multiple time steps.
This technique encourages the encoder to extract the temporal correlations of the latent dynamics from the observations~\cite{atc}, and DA can be applied as the prior-based data preprocessing.

\paragraph{\textbf{Maximizing Multi-view MI:}}
In self-supervised representation learning, feature extractors can be trained by maximizing the MI between different augmented views of the original data~\cite{MI_maximization}, and this approach has also been extended to the domain of visual RL~\cite{curl, CCLF}.

\begin{figure}[ht]
    \centering
    \includegraphics[width=0.85\textwidth]{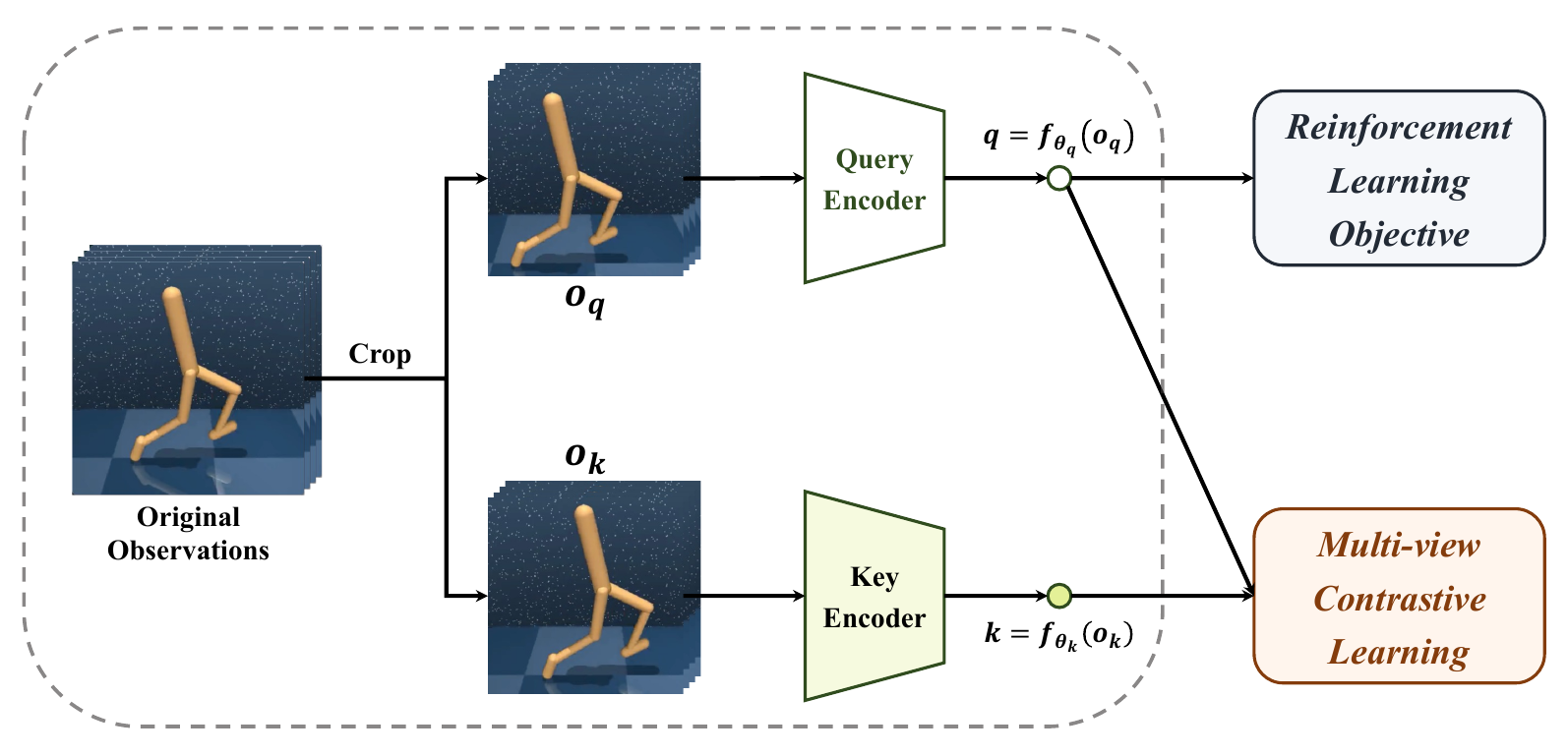} 
    \caption{The workflow of \textbf{c}ontrastive \textbf{u}nsupervised representations for \textbf{RL} (CURL).
    } 
    \label{curl} 
\end{figure}

CURL\cite{curl} is the first general framework for combining multi-view contrastive learning and DA in visual RL.
It builds an auxiliary contrastive task to learn useful state representations by maximizing the MI between the different augmented views of the same observations to improve the transformation invariance of the learned embedding.
In Figure~\ref{curl}, the contrastive representation is jointly trained with the RL objective, and the latent encoder receives gradients from both the contrastive learning objective and the RL objective.

\vspace{0.5\baselineskip}
A key component of contrastive learning is the selection of positive and negative samples relative to an anchor, and CURL uses instance discrimination rather than patch discrimination~\cite{curl}.
Specifically, the anchor and positive observations are two different augmentations of the same observation, while the negative samples come from other observations in the minibatch.
The contrastive learning task in CURL aims to maximize the MI between the anchor and the positives while minimizing the MI between the anchor and the negatives.

\vspace{0.5\baselineskip}
Following the setting of momentum contrast (MoCo)~\cite{moco}, CURL applies DA twice to generate queries and key observations, which are then encoded by the query encoder and key encoder, respectively.
The query observations $o_q$ are treated as the anchor, while the key observations $o_k$ contain the positives and negatives.
During the gradient update step, only the query encoder is updated, while the key encoder weights are set to the exponential moving average (EMA) of the query weights~\cite{moco}.
CURL employs the bilinear inner product $\operatorname{sim}(q, k)=q^{T} W k$ to measure the agreement between query-key pairs, where $W$ is a learned parameter matrix.
Then, it uses the InfoNCE loss~\cite{cpc} to build an auxiliary loss function as follows:
\begin{equation}
\mathcal{L}_\mathrm{InfoNCE}=\log \frac{\exp (q^{T} W k_{+})}{\exp (q^{T} W k_{+})+\sum_{i=0}^{K-1} \exp (q^{T} W k_{i})}
\label{infoloss}
\end{equation}
where $\{k_{0},k_{0},\dots,k_{K-1}\}$ are the keys of the dictionary and $k_{+}$ denotes a positive key.
The InfoNCE loss can be interpreted as the log-loss of a $K$-way softmax classifier whose label is $k_{+}$~\cite{MI_maximization}.

\vspace{0.5\baselineskip}
Many subsequently developed contrastive multi-view coding methods also employ the InfoNCE bound to maximize the MI between two embeddings that result from different augmentations.
For example, DRIBO~\cite{dribo} aims to maximize the InfoNCE loss $\hat{I}_{\psi}({o}_{t}^{(1)}, {o}_{t}^{(2)})$, where $\psi$ represents the learnable parameters.
Furthermore, ADAT~\cite{ADAT} selects the positive observations with the same action type and the negatives with other actions so that more positives can be produced.
CCLF~\cite{CCLF} introduces a curiosity appraisal module to select the most informative augmented observations for enhancing the effect of multi-view contrastive learning.

\begin{figure}[ht]
    \centering
    \includegraphics[width=0.8\textwidth]{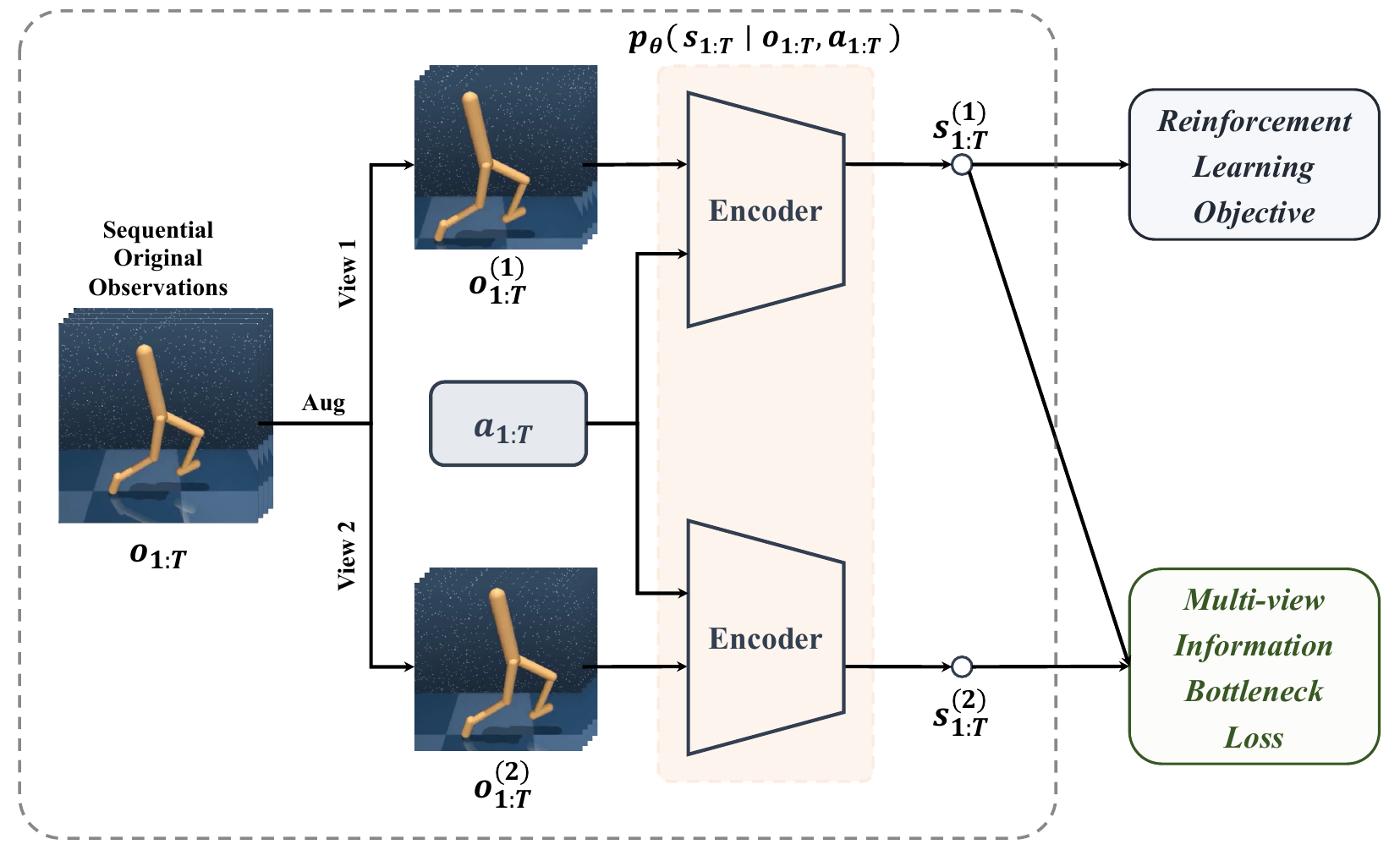} 
    \caption{The workflow of \textbf{d}eep \textbf{R}L via multi-view \textbf{i}nfomration \textbf{bo}ttleneck (DRIBO).} 
    \label{DRIBO Architecture} 
\end{figure}

Although maximizing the similarity between augmented versions of the same observation is valuable for state representation~\cite{curl,cody}, maximizing the lower-bound of MI may inevitably retain some task-irrelevant information, limiting the generalization abilities of agents.
To tackle this issue, DRIBO~\cite{dribo} uses contrastive learning combined with a multi-view information bottleneck (MIB)-based auxiliary objective to learn representations that contain only task-relevant information that is predictive of the future while eliminating task-irrelevant information.

\vspace{0.5\baselineskip}
The assumption of DRIBO is that a desired representation for RL should facilitate the prediction of future states and discard excessive, task-irrelevant information from visual observations.
In Figure~\ref{DRIBO Architecture}, the augmented observations share the same task-relevant information, while any information not shared by them is regarded as being task-irrelevant~\cite{dribo}.
In practice, the task-relevant MI can be maximized by InfoNCE, as in CURL.
With the information bottleneck principle, DRIBO constructs a relaxed Lagrangian loss to obtain a sufficient representation with minimal task-irrelevant information, and the task-irrelevant minimization term is upper-bounded by:
\begin{equation}
\mathcal{L}_{\mathrm{SKL}} = D_{\mathrm{SKL}}(p_{\theta}(\boldsymbol{s}_{t}^{(1)} \mid \boldsymbol{o}_{t}^{(1)}, \boldsymbol{s}_{t-1}^{(1)}, \boldsymbol{a}_{t-1}) \mid\mid p_{\theta}(\boldsymbol{s}_{t}^{(2)} \mid \boldsymbol{o}_{t}^{(2)}, \boldsymbol{s}_{t-1}^{(2)}, \boldsymbol{a}_{t-1}))
\label{dribo_loss}
\end{equation}
where $D_{\mathrm{SKL}}$ represents the symmetrized KL divergence based on the probability densities of $\boldsymbol{s}_{t}^{(1)}$ and $\boldsymbol{s}_{t}^{(2)}$ obtained using the encoder.
Experiments have shown that DRIBO yields significantly improved generalization and robustness on the DeepMind control suite~\cite{dmc} and Procgen~\cite{procgen} benchmarks.

\paragraph{\textbf{Maximizing Temporal Predictive MI:}}
Another popular strategy for representation learning is to learn a compact predictive coding to predict future states or information, which can also be combined with DA.
The first approach is to directly minimize the prediction error between the true future states and the predicted future states via a dynamic transition model, which will be discussed in Section~\ref{Future Prediction with Dynamics Model}.
Another approach is to maximize the lower bound of the MI between the embeddings of consecutive time steps to induce predictive representations without relying on a generative model.

\vspace{0.5\baselineskip}
CPC~\cite{cpc} and ST-DIM~\cite{st_dim} use temporal contrastive losses to maximize the MI between the previous state embedding and a future embedding several time steps later, but they both do not leverage DA to transform the observations.
Recently, ATC~\cite{atc}, CCFDM~\cite{ccfdm} and CoDy~\cite{cody} apply DA to regularize the observations obtained prior to encoding, imposing an inductive bias on information not relevant to the agent.
For example, $\mathcal{L}_{\mathrm{TMI}}$ in CoDy~\cite{cody} aims to maximize the InfoNCE bound on the temporal MI between the embedding of the current state and action and the true embedding of the next state to increase the linearity of the latent dynamics.
In practice, this approach first randomly draws a minibatch of transitions $(s_{t}, a_{t}, s_{t+1})$ from the replay buffer. 
Then, it obtains a minibatch of positive sample pairs $(z_{t}^{1}, c_{t}, z_{t+1})$ by feeding $s_{t}, a_{t}$ and $s_{t+1}$ into their corresponding encoders. 
For a given positive sample pair $(z_{t}^{1}, c_{t}, z_{t+1})$, it constructs negative samples by replacing $z_{t+1}$ with all features $z_{t+1}^{*}$ from other sample pairs $(z_{t}^{1^{*}}, c_{t}^{*}, z_{t+1}^{*})$ in the same minibatch.
Furthermore, M-CURL~\cite{M-CURL} utilizes a bidirectional transformer to reconstruct the features of masked observations using information from surrounding observations. It then captures temporal dependencies by minimizing the contrastive loss between the reconstructed features and the original features.

\subsubsection{Future Prediction with a DM}
\label{Future Prediction with Dynamics Model}
\begin{figure}[ht]
    \centering
    \includegraphics[width=0.98\textwidth]{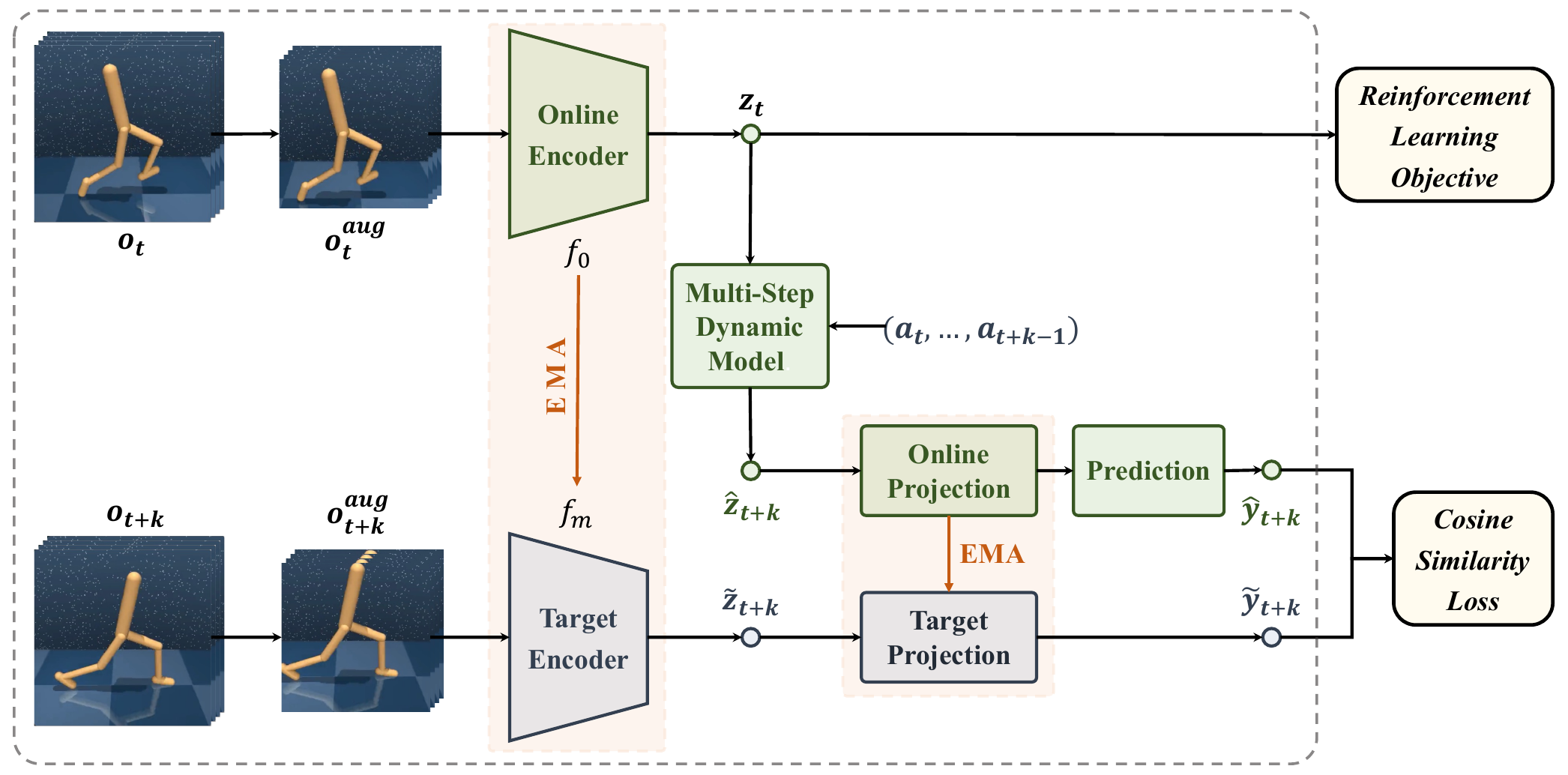} 
    \caption{The workflow of \textbf{s}elf-\textbf{p}redictive \textbf{r}epresentations (SPR).} 
    \label{SPR} 
\end{figure}

The motivation of future prediction tasks is to encourage state representations to be predictive of future states given the current state and future action sequence~\cite{cody,Predictive_1,Predictive_2}.
Instead of maximizing the MI between the current state and the future state using the InfoNCE loss~\cite{cpc,st_dim,atc}, SPR~\cite{spr} produces state representations by minimizing the prediction error between the \textbf{true future states} and the \textbf{predicted future states} using an explicit multi-step DM.
As shown in Figure~\ref{SPR}, this approach also incorporates DA into the future prediction task, which enforces consistency across different views of each observation.

\vspace{0.5\baselineskip}
The DM $h(\cdot, \cdot)$ operates entirely in the latent space to predict the transition dynamics $\left(\mathbf{z}_{t}, a_{t}\right) \rightarrow \mathbf{z}_{t+1}$, where $\mathbf{z}_{t}=f\left(\mathbf{o}_{t}\right)$ is encoded by the feature encoder $f(\cdot)$ of the current input observation $o_t$.
The prediction loss is computed by summing up the differences (errors) between the predicted representations $\hat{\mathbf{z}}_{t+1: t+K}$ and the observed representations $\tilde{\mathbf{z}}_{t+1: t+K}$:
\begin{equation}
\mathcal{L}_{\mathrm{pred}}=\sum_{k=1}^{K} d\left(\hat{\mathbf{z}}_{t+k}, \tilde{\mathbf{z}}_{t+k}\right)
=-\sum_{k=1}^{K}\left(\frac{\tilde{\mathbf{z}}_{t+k}}{\left\|\tilde{\mathbf{z}}_{t+k}\right\|_{2}}\right)^{\top}\left(\frac{\hat{\mathbf{z}}_{t+k}}{\left\|\hat{\mathbf{z}}_{t+k}\right\|_{2}}\right)
\label{pred_loss}
\end{equation}
where the latent representation $\hat{\mathbf{z}}_{t+1: t+K}$ is computed \textit{iteratively} as $\hat{\mathbf{z}}_{t+k+1} \triangleq h\left(\hat{\mathbf{z}}_{t+k}, a_{t+k}\right)$, starting from $\hat{\mathbf{z}}_{t} \triangleq \mathbf{z}_{t} \triangleq f_{o}\left(o_{t}\right)$,
and $\tilde{\mathbf{z}}_{t+k} \triangleq f_{m}\left(o_{t+k}\right)$ is computed by the target encoder $f_{m}$, whose parameters are the EMAs of the parameters of the online encoder $f_{o}$.
Combined with DA, SPR improves the agent's sample efficiency and results in superior performance with limited iterations on Atari Games and the DeepMind control suite~\cite{spr}.

\vspace{0.5\baselineskip}
PlayVirtual~\cite{playvirtual} is an extension of SPR that introduces cycle consistency to generate augmented virtual trajectories for achieving enhanced data efficiency.
Following the DM in SPR~\cite{spr}, PlayVirtual~\cite{playvirtual} proposes a BDM for backward state prediction to build a cycle/loop with a forward trajectory.
Given a DM $h(\cdot, \cdot)$, a BDM $b(\cdot, \cdot)$, the current state representation $\mathbf{z}_t$, and a sequence of actions $a_{t:t+K}$, a forward trajectory and the corresponding backward trajectory can be generated to form a synthesized trajectory:
\begin{equation}
\begin{aligned}
\textit{Forward} &: \hat{\mathbf{z}}_{t}=\mathbf{z}_{t}, \hat{\mathbf{z}}_{t+k+1}=h\left(\hat{\mathbf{z}}_{t+k}, \mathbf{a}_{t+k}\right), \text { for } k=0,1, \cdots, K-1 \\
\textit{Backward} &: \mathbf{z}_{t+K}^{\prime}=\hat{\mathbf{z}}_{t+K}, \mathbf{z}_{t+k}^{\prime}=b\left(\mathbf{z}_{t+k+1}^{\prime}, \mathbf{a}_{t+k}\right), \text { for } k=K-1, K-2, \cdots, 0
\end{aligned}
\label{cyc_loss_1}
\end{equation}

Since cycle consistency can be enforced by constraining the distance between the starting state $\mathbf{z}_{t}$ and the ending state $\mathbf{z}_{t}^{\prime}$ in the loop, appropriate synthesized training trajectories can be obtained by augmenting actions.
In practice, the cycle consistency loss can be calculated by randomly sampling $M$ sets of actions from the action space $\mathcal{A}$:
\begin{equation}
\mathcal{L}_\mathrm{cyc}=\frac{1}{M} \sum_{m=1}^{M} d_{\mathcal{M}}\left(\mathbf{z}_{t}^{\prime}, \mathbf{z}_{t}\right)
\label{cyc_loss}
\end{equation}
where $d_{\mathcal{M}}$ is the distance metric over the latent space $\mathcal{M}$.
The performance of PlayVirtual~\cite{playvirtual} can be explained from two aspects.
First, the generated trajectories can help the agent "see" more flexible experiences.
Second, enforcing the trajectory with the cycle consistency constraint can further regularize the feature representation learning process.

\subsection{Task-Specific Representation Decoupled from Policy Optimization}
\label{Task-Specific Representation Decoupled from Policy Optimization}
Utilizing DA as an implicit~\cite{rad,drq,clop} or explicit regularization approach with purposefully designed auxiliary tasks~\cite{curl,spr,cody}, the sample efficiency of visual RL has been significantly improved, resulting in performance comparable to state-based algorithms on several benchmarks~\cite{drq_v2}.
However, training \textit{{generalizable RL agents}} that are robust against irrelevant environmental variations remains a challenging task.
Similar challenges in SL tasks, such as image classification, can be addressed by strong augmentations that heavily distort the input images, such as Mixup~\cite{mixup} and CutMix~\cite{cutmix}.
However, since the training process of RL is vulnerable to excessive data variations, a naive application of DA may severely damage the training stability~\cite{svea, secant}.

\begin{figure}[ht]
    \centering
    \includegraphics[width=\textwidth]{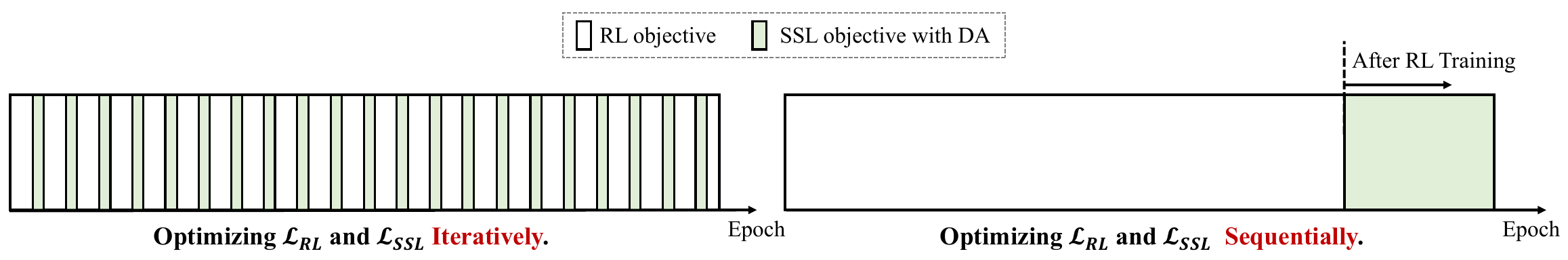} 
    \caption{Different strategies for decoupling policy optimization and representation learning.} 
    \label{different combinations} 
\end{figure}

\vspace{0.5\baselineskip}
This poses a dilemma: aggressive augmentations are necessary for achieving good generalization in the visual domain~\cite{augmix}, but injecting heavy DA into the optimization of an RL objective may cause deterioration in both the sample efficiency and the training stability~\cite{tlda}.
Recent works~\cite{soda,secant} argued that this is mainly due to the conflation of two objectives: policy optimization and representation learning.
Hence, an intuitive idea is to decouple the training data flow by using nonaugmented or weakly augmented data for RL optimization while using strongly augmented data for representation learning.
As shown in Figure~\ref{different combinations}, two strategies are available for achieving
the decoupling goal:
$(1)$ dividing the training data into two streams to separately optimize $\mathcal{L}_{RL}$ and $\mathcal{L}_{SSL}$; and {\textbf{iteratively}} updating the model parameters by the two objectives~\cite{soda};
$(2)$ optimizing the RL objective $\mathcal{L}_{RL}$ first and then {\textbf{sequentially}} leveraging DA combined with SSL objective $\mathcal{L}_{SSL}$ for knowledge distillation~\cite{secant}.

\paragraph{Optimizing $\mathcal{L}_{RL}$ and $\mathcal{L}_{SSL}$ Iteratively:}
This strategy aims to divide the training data into two data streams and only uses the nonaugmented or weakly augmented data for the RL training process; it leverages strong augmentations under prior-based diversity assumptions to optimize the self-supervised representation objective and enhance the generalization ability of the model.
In practice, this technique can be performed by iteratively optimizing the RL objective $\mathcal{L}_{RL}$ and the self-supervised representation objective $\mathcal{L}_{SSL}$ in combination with DA to update the network parameters.
For example, SODA~\cite{soda} maximizes the MI between the latent representations of augmented and nonaugmented data as the auxiliary objective  $\mathcal{L}_\mathrm{SODA}$, and \textit{\textbf{continuously alternates}} between optimizing $\mathcal{L}_{RL}$ with nonaugmented data and $\mathcal{L}_\mathrm{SODA}$ with augmented data.
While a policy is learned only from nonaugmented data, SODA still substantially benefits from DA through representation learning~\cite{soda}.

\begin{figure}[ht]
    \centering
    \includegraphics[width=0.98\textwidth]{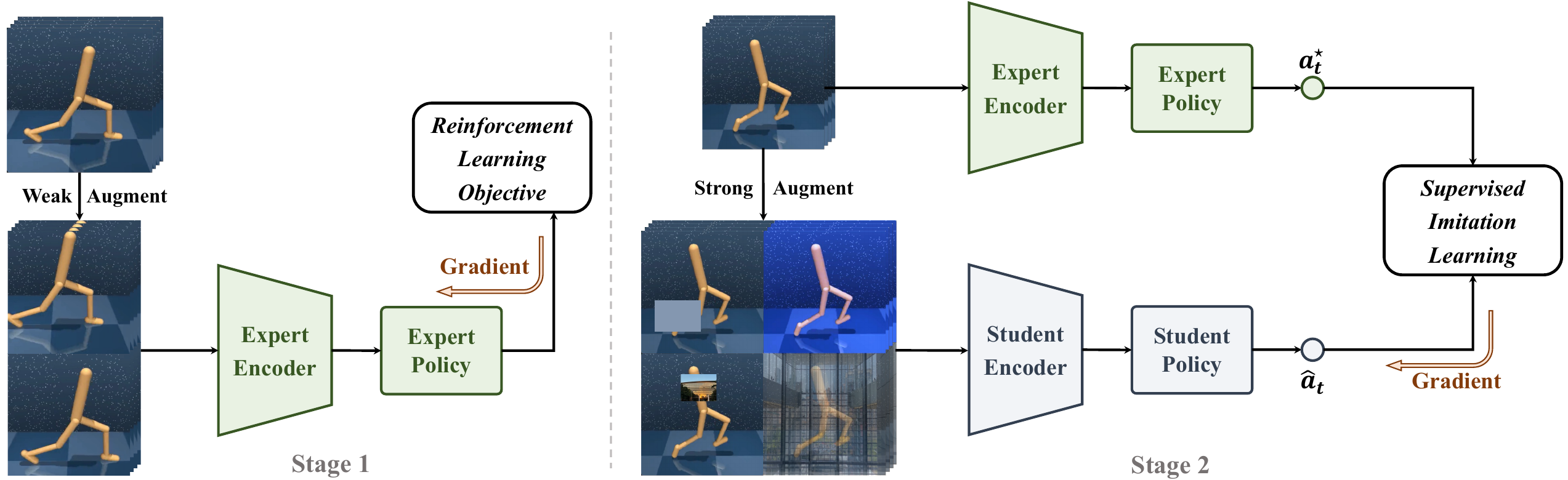} 
    \caption{The workflow of \textbf{s}elf \textbf{e}xpert \textbf{c}loning for \textbf{a}daptation to \textbf{n}ovel \textbf{t}est-env (SECANT).} 
    \label{SECANT} 
\end{figure}

\paragraph{Optimizing $\mathcal{L}_{RL}$ and $\mathcal{L}_{SSL}$ Sequentially:}
This is a two-stage training strategy, which first trains a sample-efficient agent using weak augmentations, and then enhances the state representation by auxiliary self-supervised learning or imitation learning with strong augmentations. 
For example, SECANT~\cite{secant} first trains a sample-efficient expert with random cropping (weak augmentation).
In the second stage, a student network learns a generalizable policy by mimicking the behavior of the expert at every time step but with a crucial difference: the expert produces the ground-truth actions from unmodified observations, while the student learns to predict the same actions from heavily corrupted observations, as shown in Figure~\ref{SECANT}.
The student optimizes the imitation objective by performing gradient descent on a supervised regression loss: $\mathcal{L}\left(o ; \theta_{s}\right)=\left\|\pi_{s}(f(o))-\pi_{e}(o)\right\|_{F}$, which has better training stability than the RL loss.
Furthermore, conducting policy distillation through strong augmentations can greatly remedy overfitting so that robust representations can be acquired without sacrificing policy performance.

\vspace{0.5\baselineskip}

\subsection{Task-Agnostic Representation Using Unsupervised Learning}
\label{Task-Agnostic Representation using Unsupervised Learning}
Unsupervised/self-supervised pretraining, a framework that trains models without supervision, has achieved remarkable success in various domains~\cite{mae,gpt3,TNNLS_pretrain} and can efficiently solve downstream tasks through fine-tuning.
Similarly, it is also reasonable to learn an {\textbf{unsupervised pretrained RL agent}} that can quickly adapt to diverse test tasks in a zero-shot or few-shot manner~\cite{apt, VRL3}.
Furthermore, some recent studies~\cite{atc} argued that the visual representations of standard end-to-end RL methods heavily rely on task-specific rewards, making them ineffective for other tasks.
To overcome this limitation, the environment can be first explored in {\textbf{a task-agnostic fashion to learn its visual representations}} without any task-specific rewards, and specific downstream tasks can subsequently be efficiently solved~\cite{atc,proto}.
Another key application of task-agnostic representation considers {\textbf{multi-task settings}} where, with the same or similar visual scenes, different downstream tasks are defined by corresponding reward functions.
For instance, the \textit{Walker} domain in the DeepMind control suite~\cite{dmc} consists of multiple tasks, including \textit{standing}, \textit{walking forward}, \textit{flipping backward}, etc.

\vspace{0.5\baselineskip}
Two strategies are available for learning an encoder that maps a high-dimensional input to a compact representation in a task-agnostic fashion.
The first approach is to design {\textbf{unsupervised representation tasks}}, as in Section~\ref{Policy Regularization Explicitly with Auxiliary Loss}.
Second, we can {\textbf{maximize the intrinsic rewards}} to encourage meaningful behaviors in the absence of external rewards, which are derived from self-supervised forms such as the particle-based entropy and curiosity~\cite{VISR,apt,proto,aps,sgi}.
With the ability of DA to promote prior discrimination, many unsupervised pretraining studies combine DA with other auxiliary tasks to learn more meaningful representations.
For example, ATC~\cite{atc} applies random cropping combined with contrastive learning as the task-agnostic representation tasks, while APT~\cite{apt} and SGI~\cite{sgi} leverage DA to design self-predictive tasks.

\subsection{Remarks}

As a data-centric method, DA is independent of specific RL baseline algorithms and can smoothly integrate with various techniques and training paradigms.
When applied selectively to observations without altering other aspects of the algorithm, such as loss functions and training methodologies, DA acts as a form of implicit regularization. 
Conversely, incorporating auxiliary tasks to create a joint loss function while performing DA represents explicit regularization. In some auxiliary tasks, such as DA consistency regularization and multi-view contrastive learning, DA is an indispensable component. In other tasks, such as future prediction, DA serves as an enhancement.
Furthermore, to mitigate training instability induced by strong augmentation, the decoupling of visual representation learning from policy optimization is receiving growing attention. This approach proves advantageous for both task-specific representations and general representations that are not tied to specific tasks.

\vspace{0.5\baselineskip}
The optimal approaches for maximizing the effectiveness of DA vary significantly across different scenarios. Based on recent studies, the most effective approaches to leveraging augmented data in the following three contexts are:

\begin{enumerate}
\item {\textbf{Sample-efficient RL in the single-environment setting:}}
Weak augmentations, such as random shift, are considered the optimal type of DA for enhancing sample efficiency during training. The success of DrQ~\cite{drq} and DrQ-v2~\cite{drq_v2} demonstrates that using DA solely as an implicit regularization technique can significantly improve an agent’s sample efficiency. 
Surprisingly, unlike in supervised learning, where DA typically provides incremental enhancements to training efficiency, DA can decisively influence the sample efficiency of visual RL algorithms~\cite{Learning_Better_with_Less}. 
Moreover, recent studies have demonstrated that DA alone can achieve more efficient performance compared to the meticulous design of self-supervised learning tasks~\cite{does_ssl} or the pre-training of representation encoders using additional data~\cite{hansen2022pre}. 
This unique phenomenon may be attributed to the different mechanisms through which DA facilitates sample-efficient visual RL, as opposed to traditional computer vision tasks. Further elaboration on this will be provided in Section~\ref{Data Augmentation in Supervised Learning vs. in Reinforcement Learning}.

\item {\textbf{Generalizable RL in the multi-environment setting:}}
Strong augmentations based on priors, such as Color Jitter, are crucial for enhancing the generalization ability of visual RL. The key to leveraging strong augmentations lies in balancing their effectiveness while avoiding severe training instability. This trade-off will be discussed in detail in Section~\ref{Trade-off between the Instability and Generalization Ability Caused by DA}. Although SVEA~\cite{svea} and SADA~\cite{SADA} have demonstrated that implicit regularization can handle strong augmentations through careful data flow design, employing specific representation tasks to utilize DA for generalization improvement is more conducive to maintaining training stability~\cite{spr}. Additionally, decoupling representation learning from policy optimization has shown greater potential and should be considered the best practice for leveraging DA in this context~\cite{secant, soda}.

\item {\textbf{Generalizable RL in the multi-task setting:}}
Combining DA with unsupervised learning is an effective strategy for training general visual representations, enabling agents to quickly adapt to multiple tasks. As the pre-training followed by fine-tuning paradigm gains popularity, there is growing interest in pre-training task-agnostic representations and subsequently developing efficient policies for specific tasks. Beyond training task-agnostic encoders from visual RL task observations using DA and unsupervised learning, leveraging diverse OOD vision data from other domains has shown greater potential for efficient adaptation to downstream tasks~\cite{VRL3,RRL,mvp,PVR,R3M}. In Section~\ref{The Role of Visual RL and DA in the Age of Foundation Models}, we will discuss the significance and necessity of DA in the context of pre-trained visual RL paradigms.

\end{enumerate}

%% file: textfile/5_Experimental_Evaluation.tex
\newpage
\section{Experimental Evaluation}
\label{Experimental Evaluation}
This section provides a systematic empirical evaluation of the methods in visual RL that leverage DA.
First in Section~\ref{Typical Benchmarks in Related Works}, we introduce the commonly used benchmarks for evaluating the sample efficiency and generalization ability of agents.
Then in Section~\ref{Sample Efficiency Evaluation} and Section~\ref{Zero-shot Generalization Evaluation}, we present the experimental results of representative RL techniques using DA in comparison with those of other baselines to demonstrate the effectiveness of DA and identify the pros and cons of these methods.

\subsection{Representative Benchmarks}
\label{Typical Benchmarks in Related Works}

\subsubsection{Benchmarks for Sample Efficiency Evaluating in Visual RL}

\paragraph{\textbf{Atari Games}~\cite{atari}}
This suite of games is widely used by both state-based and image-based discrete control algorithms for sample-constrained evaluations~\cite{dqn2013}.
While RL algorithms can achieve superhuman performance on Atari games, they are still far less efficient than human learners, especially in image-based cases~\cite{curl}.
In the sample-efficient \textit{\textbf{Atari-100k}} setting, only 100k interactions (400k frames with frame-skip=4) are available.
The performance of an agent on a game is measured by its human-normalized score (HNS), defined as $\frac{S_{A}-S_{R}}{S_{H}-S_{R}}$ where $S_{A}$ is the agent's score; $S_{R}$ is the score of a random play; $S_{H}$ is the expert human score.

\vspace{0.5\baselineskip}
\paragraph{\textbf{DeepMind Control Suite}~\cite{dmc}} 
This is a continuous control benchmark suite for evaluating visual RL algorithms.
It presents a variety of challenging tasks, including bipedal balancing, locomotion, contact forces, and goal reaching, with both sparse and dense reward signals.
Previous studies usually measured the data efficiency and performance of their algorithms on the DeepMind control suites with 100k (for measuring learning speed) and 500k (for measuring overall performance) environment steps, which are referred to as \textit{\textbf{DMControl-100k}} and \textit{\textbf{DMControl-500k}}, respectively.
DeepMind control suite is also a proper testbed for multi-task settings, as different tasks often involve the same domain.
For example, the walker domain contains running, walking, standing and many other tasks, allowing agents to transfer learned policies to other tasks with similar visual observations.

\subsubsection{Benchmarks for Generalization Evaluating in Visual RL}\label{gen ben}
\begin{figure}[ht] 
\centering 
\includegraphics[width=0.95\textwidth]{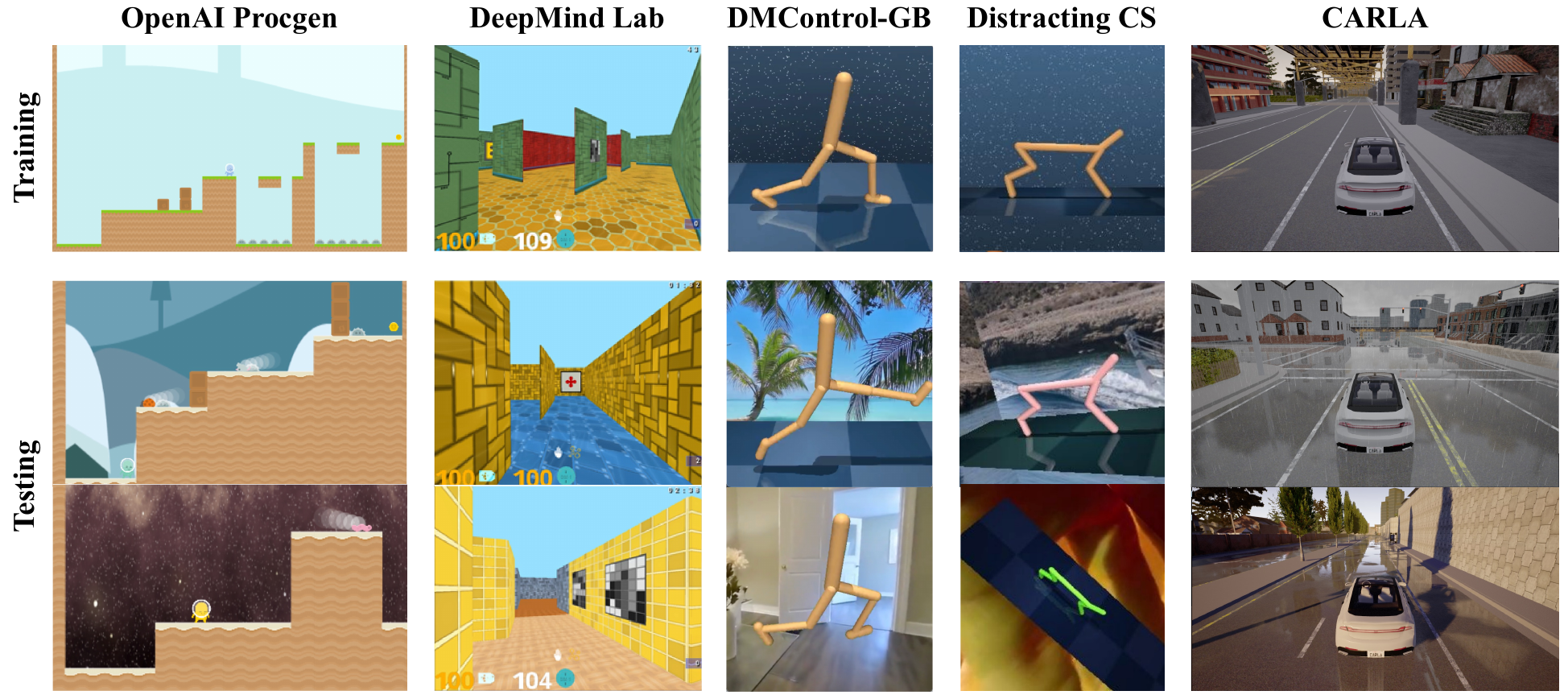} 
\caption{Typical benchmarks used to evaluate an agent’s generalization ability in visual RL.} 
\label{benchmark} 
\end{figure}

Although Atari Games and the DeepMind control suite are suitable for benchmarking the sample efficiency of visual RL agents, they are not applicable for investigating the generalization abilities of these agents~\cite{rad}.
Generally, measuring the generalization ability of an agent requires variations between the training environment and the test environment, including state-space variations (the initial state distribution), dynamics variations (the transition function), visual variations (the observation function), and reward function variations~\cite{Generalisation_survey}.
In particular, DA-based techniques focus on zero-shot generalization to unseen environments with similar high-level goals and dynamics but different layouts and visual properties~\cite{svea, clop, p3o}.
Figure~\ref{benchmark} shows the representative benchmarks for evaluating the agent's generalization ability in visual RL.

\paragraph{\textbf{OpenAI Procgen}~\cite{procgen}}
This is a suite of game-like environments where different levels feature varying visual attributes.
Different combinations of the game levels can be used to separately construct training and test environments.
Agents are only allowed to be trained on limited levels and are evaluated on unseen levels with different backgrounds or layouts~\cite{mixreg, clop}.

\paragraph{\textbf{DeepMind Lab}~\cite{deepmindlab}}
This is a first-person 3D maze environment in which various objects are placed in the rooms.
As a measure of their generalization ability, agents are trained to collect objects in a fixed map layout and tested in unseen environments that differ only in terms of their walls and floors (i.e., the variational contexts)~\cite{randfm}.

\paragraph{\textbf{DeepMind Control Suite Variants}~\cite{DistractingCS,soda,Generalisation_survey}}
Since the original DeepMind control suite is not applicable for studying generalization, a number of variants have been proposed in recent years.
Most of them, such as DMControl-GB~\cite{soda}, DMControl-Remastered~\cite{DMC_Remastered} and Natural Environments~\cite{natural}, focus on visual generalization by changing the colors or styles of the background and floors.
Furthermore, the Distracting Control Suite (DCS)~\cite{DistractingCS} features a broader set of variations, including background style and camera pose variations.

\paragraph{\textbf{CARLA}~\cite{carla}}
This is a realistic driving simulator where the agent’s goal is to drive as far as possible in 1000 time steps without colliding into 20 other moving vehicles or barriers~\cite{secant}.
Learning directly from the rich observations in this scenario is challenging since diverse types of task-irrelevant distractors (e.g., lighting conditions, shadows, realistic rain, clouds, etc.) are available around the agent, which increases the difficulty of extracting control-related features~\cite{li2024normalization}.

\subsection{Sample Efficiency Evaluation}
\label{Sample Efficiency Evaluation}
To measure the sample efficiency, we report the results on three common benchmarks: Atari-100k~\cite{atari}, DMControl-100k and DMControl-500k~\cite{dmc}.

\subsubsection{Atari-100k}

In Table~\ref{Sample Efficiency Evaluation in Atari},
the results of a random player (Random) and an expert human player (Human) are copied from~\cite{duelingDQN} as baselines.
Other scores are copied from their original papers~\cite{drq, CCLF, ADAT, M-CURL, spr, playvirtual}.
The results show that augmenting the observations as implicit regularization is effective, boosting the performance in terms of the median HNS from $5.8\%$ (Efficient DQN) to $26.8\%$ (DrQ).
Moreover, appropriate auxiliary tasks such as contrastive learning~\cite{curl,M-CURL,CCLF,ADAT} and future prediction representation~\cite{spr,CVPR22_LVT,playvirtual} can further yield improved sample efficiency.
Among them, SPR~\cite{spr} achieves the highest mean HNS value ($70.4\%$) with its future prediction module, while
PlayVirtual~\cite{playvirtual} achieves the highest median HNS value ($47.2\%$) with the trajectory augmentation.

\begin{table}[ht]
\centering
\caption{\textbf{Evaluation of Sample Efficiency on Atari-100k.} We report the scores and the mean and median HNSs achieved by different methods on Atari-100k. The results are copied from the original works~\cite{drq, CCLF, ADAT, M-CURL, spr, playvirtual}.}
\label{Sample Efficiency Evaluation in Atari}
\renewcommand{\arraystretch}{1.15}
\begin{small}
\setlength{\tabcolsep}{2pt}
\resizebox{\columnwidth}{!}{
\begin{tabular}{lcccccccccc}
\toprule
\multirow{2}*{\textbf{Game}} & \multirow{2}*{\textbf{Human}} & \multirow{2}*{\textbf{Random}} & \textbf{DQN} & \textbf{CURL} & \textbf{CCLF} & \textbf{ADAT} & \textbf{DrQ} & \textbf{M-CURL} & \textbf{SPR} & \textbf{PlayVirtual}\\
~ & ~ & ~ & \cite{dqn2013} & \cite{curl} & \cite{CCLF} & \cite{ADAT} & \cite{drq} & \cite{M-CURL} & \cite{spr} & \cite{playvirtual} \\
\midrule
Alien & $7127.7$ & $227.8$ & $558.1$ & $558.2$ & $ 920.0 $ & $ 1029.7 $ & $771.2$ & $ \textcolor[RGB]{192,0,0}{\bm{1151.6}} $ & $801.5$ & $947.8$ \\
Amidar & $1719.5$ & $5.8$ & $63.7$ & $142.1$ & $ 154.7 $ & $ 147.3 $ & $102.8$ & $ \textcolor[RGB]{192,0,0}{\bm{182.2}} $ & $176.3$ & $165.3$ \\
Assault & $742.0$ & $222.4$ & $589.5$ & $600.6$ & $ 612.4 $ & $ \textcolor[RGB]{192,0,0}{\bm{749.4}} $ & $452.4$ & $ 613.5 $ & $571.0$ & $702.3$ \\
Asterix & $8503.3$ & $210.0$ & $341.9$ & $734.5$ & $ 708.8 $ & $864$ & $603.5$ & $ 738.1 $ & $\textcolor[RGB]{192,0,0}{\bm{977.8}}$ & $933.3$  \\
Bank Heist & $753.1$ & $14.2$ & $74.0$ & $131.6$ & $ 36.0 $ & $ 164 $ & $168.9$ & $ 220 $ & $\textcolor[RGB]{192,0,0}{\bm{380.9}}$ & $245.9$ \\
Battle Zone & $37187.5$ & $2360.0$ & $4760.8$ & $14870.0$ & $ 5775.0 $ & $ 21240 $ & $12954.0$ & $ \textcolor[RGB]{192,0,0}{\bm{21600}} $ & $16651.0$ & $13260.0$ \\
Boxing & $12.1$ & $0.1$ & $-1.8$ & $1.2$ & $ 7.4 $ & $ 0.4 $ & $6.0$ & $ 5.9 $ & $35.8$ & $\textcolor[RGB]{192,0,0}{\bm{38.3}}$ \\
Breakout & $30.5$ & $1.7$ & $7.3$ & $4.9$ & $ 2.7 $ & $ 4.5 $ & $16.1$ & $ 5.7 $ & $17.1$ & $\textcolor[RGB]{192,0,0}{\bm{20.6}}$ \\
Chopper Command & $7387.8$ & $811.0$ & $624.4$ & $1058.5$ & $ 765.0 $ & $ 1106 $ & $780.3$ & $ \textcolor[RGB]{192,0,0}{\bm{1138.9}} $ & $974.8$ & $922.4$ \\
Crazy Climber & $35829.4$ & $10780.5$ & $5430.6$ & $12146.5$ & $ 7845.0 $ & $ 21240 $ & $20516.5$ & $ 20781.2 $ & $\textcolor[RGB]{192,0,0}{\bm{42923.6}}$ & $23176.7$ \\
Demon Attack & $1971.0$ & $152.1$ & $403.5$ & $817.6$ & $ \textcolor[RGB]{192,0,0}{\bm{1360.9}} $ & $ 851.9 $ & $1113.4$ & $ 864.4 $ & $545.2$ & $1131.7$ \\
Freeway & $29.6$ & $0.0$ & $3.7$ & $26.7$ & $ 22.6 $ & $ \textcolor[RGB]{192,0,0}{\bm{29.7}} $ & $9.8$ & $ 28.9 $ & $24.4$ & $16.1$ \\
Frostbite & $4334.7$ & $65.2$ & $202.9$ & $1181.3$ & $ 1401.0 $ & $ 1943.2 $ & $331.1$ & $ \textcolor[RGB]{192,0,0}{\bm{2342.2}} $ & $1821.5$ & $1984.7$  \\
Gopher & $2412.5$ & $257.6$ & $320.8$ & $669.3$ & $ \textcolor[RGB]{192,0,0}{\bm{814.7}} $ & $ 601.2 $ & $636.3$ & $ 453.8 $ & $715.2$ & $684.3$ \\
Hero & $30826.4$ & $1027.0$ & $2200.1$ & $6279.3$ & $ 6944.5 $ & $ 7259.2 $ & $3736.3$ & $ 7360.6 $ & $7019.2$ & $\textcolor[RGB]{192,0,0}{\bm{8597.5}}$  \\
Jamesbond & $302.8$ & $29.0$ & $133.2$ & $471.0$ & $ 308.8 $ & $ \textcolor[RGB]{192,0,0}{\bm{635.7}} $ & $236.0$ & $ 436.2 $ & $365.4$ & $394.7$  \\
Kangaroo & $3035.0$ & $52.0$ & $448.6$ & $872.5$ & $ 650.0 $ & $ 956.9 $ & $940.6$ & $ 1691.4 $ & $\textcolor[RGB]{192,0,0}{\bm{3276.4}}$ & $2384.7$ \\
Krull & $2665.5$ & $1598.0$ & $2999.0$ & $\textcolor[RGB]{192,0,0}{\bm{4229.6}}$ & $ 3975.0 $ & $ 3502.9 $ & $4018.1$ & $ 3240.9 $ & $3688.9$ & $3880.7$ \\
Kung Fu Master & $22736.3$ & $258.5$ & $2020.9$ & $14307.8$ & $ 12605.0 $ & $ \textcolor[RGB]{192,0,0}{\bm{19146}} $ & $9111.0$ & $ 17645.6 $ & $13192.7$ & $14259.0$ \\
Ms Pacman & $6951.6$ & $307.3$ & $872.0$ & $1465.5$ & $ 1397.5 $ & $ 1075 $ & $960.5$ & $ \textcolor[RGB]{192,0,0}{\bm{1758.9}} $ & $1313.2$ & $1335.4$ \\
Pong & $14.6$ & $-20.7$ & $-19.4$ & $-16.5$ & $ -17.3 $ & $ -15.1 $ & $-8.5$ & $ -8.9 $ & $-5.9$ & $\textcolor[RGB]{192,0,0}{\bm{-3.0}}$ \\
Private Eye & $69571.3$ & $24.9$ & $351.3$ & $218.4$ & $ 100.0 $ & $ \textcolor[RGB]{192,0,0}{\bm{388}} $ & $-13.6$ & $ 321.6 $ & $124.0$ & $93.9$  \\
Qbert & $13455.0$ & $163.9$ & $627.5$ & $1042.4$ & $ 953.8 $ & $ 1578 $ & $854.4$ & $ 1785 $ & $669.1$ & $\textcolor[RGB]{192,0,0}{\bm{3620.1}}$ \\
Road Runner & $7845.0$ & $11.5$ & $1491.9$ & $5661.0$ & $ 11730.0 $ & $ 12508 $ & $8895.1$ & $ 12320 $ & $\textcolor[RGB]{192,0,0}{\bm{14220.5}}$ & $13534.0$  \\
Seaquest & $42054.7$ & $68.4$ & $240.1$ & $384.5$ & $ 550.5 $ & $ 251.6 $ & $301.2$ & $ 481.1 $ & $\textcolor[RGB]{192,0,0}{\bm{583.1}}$ & $527.7$  \\
Up N Down & $11693.2$ & $533.4$ & $2901.7$ & $2955.2$ & $ 3376.3 $ & $ 3597.8 $ & $3180.8$ & $ 4399.5 $ & $\textcolor[RGB]{192,0,0}{\bm{28138.5}}$ & $10225.2$  \\
\midrule
Mean HNS ($ \% $) & $100$ & $0$ & $13.7$ & $38.1$ & $ 38.2 $ & $ 47.2 $ & $35.7$ & $ 46.6 $ & $\textcolor[RGB]{192,0,0}{\bm{70.4}}$ & $63.7$ \\
Median HNS ($ \% $) & $100$ & $0$ & $5.8$ & $17.5$ & $ 18.1 $ & $20.6$ & $26.8$ & $ 34.0 $ & $41.5$ & $\textcolor[RGB]{192,0,0}{\bm{47.2}}$ \\
\midrule
\# Superhuman & N/A & 0 & $1$ & $2$ & $3$ & $6$ & $2$ & $ 3 $ & $\textcolor[RGB]{192,0,0}{\bm{7}}$ & $4$ \\
\# SOTA & N/A & 0 & $0$ & $1$ & $2$ & $5$ & $0$ & $ 6 $ & $\textcolor[RGB]{192,0,0}{\bm{7}}$ & $5$ \\
\bottomrule
\end{tabular}
}
\end{small}
\end{table}

\subsubsection{DMControl-100k and DMControl-500k}
Compared with Atari games, the tasks in the DeepMind control suite~\cite{dmc} are more complex and challenging.
We first report the performance of the underlying SAC algorithm~\cite{sac} based on state and image inputs, referred to as Pixel SAC and State SAC in Table~\ref{Sample Efficiency Evaluation in DMControl Suite} (copied from~\cite{curl}), respectively, followed by the results of SAC-AE~\cite{sac_ae}.
Since State SAC operates on low-dimensional state-based features instead of pixels, it approximates the upper bounds of sample efficiency in these environments for image-based agents.
Similar to the case of Atari-100k, DrQ~\cite{drq} achieves significant improvements over the underlying SAC algorithm~\cite{sac}, which is unable to complete these tasks.
Combining auxiliary tasks with DA provides improved performance and potential for training sample-efficient agents.
For example, based on SPR~\cite{spr}, recent studies have achieved superior performance by introducing cycle consistency constraints for more diverse trajectories (PlayVirtual~\cite{playvirtual}) or curiosity modules for better exploration (CCFDM~\cite{ccfdm}).
\begin{table}[ht]
\centering
\caption{\textbf{Evaluation of Sample Efficiency on the DeepMind Control Suite.} The reported scores (means and standard deviations) are achieved by different methods on DMControl-100k and DMControl-500k.
The results are copied from their original works with 10 random seeds~\cite{sac_ae,curl,drq,spr,playvirtual,CCLF,ccfdm,cody}.
}
\label{Sample Efficiency Evaluation in DMControl Suite}
\begin{small}
\setlength{\tabcolsep}{4pt}
\resizebox{\columnwidth}{!}{
\begin{tabular}{lcccccccccc|c}
\toprule
{\textbf{DMControl}} & \textbf{Pixel} & \textbf{SAC-AE} & \textbf{CURL} & \textbf{DrQ} & \textbf{SPR} &  \textbf{CCLF} & \textbf{CoDy} & \textbf{MLR} & \textbf{CCFDM} & \textbf{PlayVirtual} & {\textbf{State}} \\
{\textbf{100k}} & \textbf{SAC} & \textbf{\cite{sac_ae}} & \textbf{\cite{curl}} & \textbf{\cite{drq}} & \textbf{\cite{spr}} & \textbf{\cite{CCLF}} & \textbf{\cite{cody}} & \textbf{\cite{mlr}} & \textbf{\cite{ccfdm}} & \textbf{\cite{playvirtual}} & {\textbf{SAC}} \\
\midrule
\textbf{Finger,} & \normalsize$179$ & \normalsize$747$ & \normalsize$767$ & \normalsize$901$ & \normalsize$868$ & \normalsize$\textcolor[RGB]{192,0,0}{\bm{944}}$ & \normalsize$887$ & \normalsize$907$ & \normalsize$880$ & \normalsize$915$ &\normalsize$811$ \\
\textbf{Spin} & \footnotesize$\pm166$ & \footnotesize$\pm130$ & \footnotesize$\pm56$ & \footnotesize$\pm104$ & \footnotesize$\pm143$ & \footnotesize$\textcolor[RGB]{192,0,0}{{\pm42}}$ & \footnotesize$\pm39$ & \footnotesize$\pm58$& \footnotesize$\pm142$ & \footnotesize$\pm49$ & \footnotesize$\pm46$ \\
\textbf{Cartpole,} & \normalsize$419$ & \normalsize$276$ & \normalsize$582$ & \normalsize$759$ & \normalsize$799$ & \normalsize$799$  & \normalsize$784$ & \normalsize$806$& \normalsize$785$ &\normalsize$\textcolor[RGB]{192,0,0}{\bm{816}}$ & \normalsize$835$ \\
\textbf{Swingup} & \footnotesize$\pm40$ & \footnotesize$\pm38$ & \footnotesize$\pm146$ & \footnotesize$\pm92$ & \footnotesize$\pm42$ &  \footnotesize$\pm61$ & \footnotesize$\pm18$ & \footnotesize$\pm48$ & \footnotesize$\pm87$ & \footnotesize$\textcolor[RGB]{192,0,0}{{\pm36}}$ & \footnotesize$\pm22$ \\
\textbf{Reacher,} & \normalsize$145$ & \normalsize$225$ & \normalsize$538$ & \normalsize$601$ & \normalsize$638$ & \normalsize$738$ & \normalsize$624$ & \normalsize$\textcolor[RGB]{192,0,0}{\bm{866}}$ & \normalsize$811$ & \normalsize$785$ & \normalsize$746$ \\
\textbf{Easy} & \footnotesize$\pm30$ & \footnotesize$\pm164$ & \footnotesize$\pm233$ & \footnotesize$\pm213$ & \footnotesize$\pm269$ & \footnotesize$\pm99$ & \footnotesize$\pm42$ & \footnotesize$\textcolor[RGB]{192,0,0}{{\pm103}}$ & \footnotesize${\pm220}$ & \footnotesize$\pm142$ & \footnotesize$\pm25$ \\
\textbf{Cheetah,} & \normalsize$197$ & \normalsize$252$ & \normalsize$299$ & \normalsize$344$ & \normalsize$467$ & \normalsize$317$ & \normalsize$323$ & \normalsize$\textcolor[RGB]{192,0,0}{\bm{482}}$ & \normalsize$274$ & \normalsize$474$ & \normalsize$616$ \\
\textbf{Run} & \footnotesize$\pm15$ & \footnotesize$\pm173$ & \footnotesize$\pm48$ & \footnotesize$\pm67$ & \footnotesize$\pm36$ & \footnotesize$\pm38$ & \footnotesize$\pm29$ & \footnotesize$\textcolor[RGB]{192,0,0}{{\pm38}}$ & \footnotesize$\pm98$ & \footnotesize$\pm50$ & \footnotesize$\pm18$ \\
\textbf{Walker,} & \normalsize$42$ & \normalsize$395$ & \normalsize$403$ & \normalsize$612$ & \normalsize$398$ & \normalsize$648$ & \normalsize$\textcolor[RGB]{192,0,0}{\bm{673}}$ & \normalsize$643$ & \normalsize$634$ & \normalsize$460$ & \normalsize$891$ \\
\textbf{Walk} & \footnotesize$\pm12$ & \footnotesize$\pm58$ & \footnotesize$\pm24$ & \footnotesize$\pm164$ & \footnotesize$\pm165$ & \footnotesize$\pm110$ & \footnotesize$\textcolor[RGB]{192,0,0}{{\pm94}}$ & \footnotesize$\pm114$ & \footnotesize$\pm132$ & \footnotesize$\pm173$ & \footnotesize$\pm82$ \\
\textbf{Ball in cup,} & \normalsize$312$ & \normalsize$338$ & \normalsize$769$ & \normalsize$913$ & \normalsize$861$ & \normalsize$914$ & \normalsize$948$ & \normalsize$933$ & \normalsize$\textcolor[RGB]{192,0,0}{\bm{962}}$ &\normalsize$926$ & \normalsize$746$ \\
\textbf{Catch} & \footnotesize$\pm63$ & \footnotesize$\pm196$ & \footnotesize$\pm43$ & \footnotesize$\pm53$ & \footnotesize$\pm233$ & \footnotesize$\pm20$ & \footnotesize$\pm6$ & \footnotesize$\pm16$ & \footnotesize$\textcolor[RGB]{192,0,0}{{\pm28}}$ & \footnotesize$\pm31$ & \footnotesize$\pm91$ \\
\midrule
\midrule
\textbf{{500k}} &  &  &  &  &  &  &  &  &  &  &  \\
\midrule
\textbf{Finger,} & \normalsize$179$ & \normalsize$914$ & \normalsize$926$ & \normalsize$938$ & \normalsize$924$ & \normalsize$\textcolor[RGB]{192,0,0}{\bm{974}}$ & \normalsize$937$ & \normalsize$973$ & \normalsize$906$ & \normalsize$963$ & \normalsize$811$ \\
\textbf{Spin} & \footnotesize$\pm166$ & \footnotesize$\pm107$ & \footnotesize$\pm45$ & \footnotesize$\pm103$ & \footnotesize$\pm132$ & \footnotesize$\textcolor[RGB]{192,0,0}{{\pm6}}$ & \footnotesize$\pm41$ & \footnotesize$\pm31$ & \footnotesize$\pm152$ & \footnotesize$\pm40$ & \footnotesize$\pm46$ \\
\textbf{Cartpole,} & \normalsize$419$ & \normalsize$730$ & \normalsize$841$ & \normalsize$868$ & \normalsize$870$ & \normalsize$869$ & \normalsize$869$ & \normalsize$872$ & \normalsize$\textcolor[RGB]{192,0,0}{\bm{975}}$ & \normalsize$865$ & \normalsize$835$ \\
\textbf{Swingup} & \footnotesize$\pm40$ & \footnotesize$\pm152$ & \footnotesize$\pm45$ & \footnotesize$\pm10$ & \footnotesize$\pm12$ & \footnotesize$\pm9$ & \footnotesize$\pm4$ & \footnotesize$\pm5$ & \footnotesize$\textcolor[RGB]{192,0,0}{{\pm38}}$ & \footnotesize$\pm11$ & \footnotesize$\pm22$ \\
\textbf{Reacher,} & \normalsize$145$ & \normalsize$601$ & \normalsize$929$ & \normalsize$942$ & \normalsize$925$ & \normalsize$941$ & \normalsize$957$ & \normalsize$957$ & \normalsize$\textcolor[RGB]{192,0,0}{\bm{973}}$ & \normalsize$942$ & \normalsize$746$ \\
\textbf{Easy} & \footnotesize$\pm30$ & \footnotesize$\pm135$ & \footnotesize$\pm44$ & \footnotesize$\pm71$ & \footnotesize$\pm79$ & \footnotesize$\pm48$ & \footnotesize$\pm16$ & \footnotesize$\pm41$ & \footnotesize$\textcolor[RGB]{192,0,0}{{\pm36}}$ & \footnotesize$\pm66$ & \footnotesize$\pm25$ \\
\textbf{Cheetah,} & \normalsize$197$ & \normalsize$544$ & \normalsize$518$ & \normalsize$660$ & \normalsize$716$ & \normalsize$588$ & \normalsize$656$ & \normalsize$674$ & \normalsize$552$ & \normalsize$\textcolor[RGB]{192,0,0}{\bm{719}}$ & \normalsize$616$ \\
\textbf{Run} & \footnotesize$\pm15$ & \footnotesize$\pm50$ & \footnotesize$\pm28$ & \footnotesize$\pm96$ & \footnotesize$\pm47$ & \footnotesize$\pm22$ & \footnotesize$\pm43$ & \footnotesize$\pm37$ & \footnotesize$\pm130$ & \footnotesize$\textcolor[RGB]{192,0,0}{{\pm51}}$ & \footnotesize$\pm18$ \\
\textbf{Walker,} & \normalsize$42$ & \normalsize$858$ & \normalsize$902$ & \normalsize$921$ & \normalsize$916$ & \normalsize$936$ & \normalsize$\textcolor[RGB]{192,0,0}{\bm{943}}$ & \normalsize$939$ & \normalsize$929$ & \normalsize$928$ & \normalsize$891$ \\
\textbf{Walk} & \footnotesize$\pm12$ & \footnotesize$\pm82$ & \footnotesize$\pm43$ & \footnotesize$\pm45$ & \footnotesize$\pm75$ & \footnotesize$\pm23$ & \footnotesize$\textcolor[RGB]{192,0,0}{{\pm17}}$ & \footnotesize$\pm10$ & \footnotesize$\pm68$ & \footnotesize$\pm30$ & \footnotesize$\pm82$ \\
\textbf{Ball in cup,} & \normalsize$312$ & \normalsize$810$ & \normalsize$959$ & \normalsize$863$ & \normalsize$963$ & \normalsize$961$ & \normalsize$970$ & \normalsize$964$ & \normalsize$\textcolor[RGB]{192,0,0}{\bm{979}}$ & \normalsize$967$ & \normalsize$746$ \\
\textbf{Catch} & \footnotesize$\pm63$ & \footnotesize$\pm121$ & \footnotesize$\pm27$ & \footnotesize$\pm9$ & \footnotesize$\pm8$ & \footnotesize$\pm9$ & \footnotesize$\pm4$ & \footnotesize$\pm14$ & \footnotesize$\textcolor[RGB]{192,0,0}{{\pm17}}$ & \footnotesize$\pm5$ & \footnotesize$\pm91$ \\
\bottomrule
\end{tabular}
}
\end{small}
\end{table}

\subsection{Zero-Shot Generalization Evaluation}
\label{Zero-shot Generalization Evaluation}

In this subsection, we report the studies conducted on two benchmarks representing two different types of generalization: Procgen~\cite{procgen} for level generalization in arcade games, and DMControl-GB~\cite{soda} for vision generalization in robot control tasks.

\subsubsection{Level Generalization on Procgen}
In Table~\ref{Generalization Evaluation in Procgen}, the results of RAD~\cite{rad} and DrAC~\cite{drac} are based on their most suitable augmentation types for different environments, and UCB-DrAC selects the most suitable type of DA as a multi-armed bandit problem.
Based on the comparison of RAD~\cite{rad} and its underlying PPO algorithm~\cite{ppo}, it is evident that appropriate augmentations are beneficial in almost every environment.
Additionally, explicitly regularizing the policy and value functions after performing augmentations (as in DrAC~\cite{drac}) leads to further improvements.
The outstanding results of CLOP~\cite{clop} and DRIBO~\cite{dribo} highlight the remarkable potential of subtly designed representation learning methods to distinguish task-relevant information from task-irrelevant information.
\begin{table}[ht]
\renewcommand{\arraystretch}{1.3}
\centering
\caption{\textbf{Evaluation of Generalization Ability on Procgen.} Agents are trained on the first 200 levels of each game and evaluated on unseen levels. The scores are copied from the original papers on UCB-DrAC~\cite{drac} and DRIBO~\cite{dribo}. The mean and standard deviation values are calculated with 10 random seeds.}
\begin{small}
\label{Generalization Evaluation in Procgen}
\setlength{\tabcolsep}{4pt}
\resizebox{\columnwidth}{!}{
\begin{tabular}{lcccccccc}
\toprule
\multirow{2}*{\textbf{Game}} & \textbf{PPO} & \textbf{RandFM} & \textbf{MixReg} & \textbf{RAD} & \textbf{DrAC} & \textbf{UCB-DrAC} & \textbf{CLOP} &\textbf{DRIBO}  \\
~ & \cite{ppo} & \cite{randfm} & \cite{mixreg} & \cite{rad} & \cite{drac} & \cite{drac} & \cite{clop} & \cite{dribo} \\
\midrule
\textbf{BigFish} & \normalsize$4.0 $\footnotesize$\pm 1.2$ & \normalsize$0.6 $\footnotesize$\pm 0.8$ & \normalsize$ 7.1$\footnotesize$\pm1.6 $ & \normalsize$9.9 $\footnotesize$\pm 1.7$ & \normalsize$8.7 $\footnotesize$\pm 1.4$ & \normalsize$9.7$\footnotesize$ \pm 1.0$ & \normalsize$\textcolor[RGB]{192,0,0}{\bm{19.2}}$\footnotesize$\textcolor[RGB]{192,0,0}{\bm{\pm 4.6}}$ & \normalsize$10.9$\footnotesize$\pm1.6$ \\
\textbf{StarPilot} & \normalsize$24.7$\footnotesize$ \pm 3.4$ & \normalsize$8.8 $\footnotesize$\pm 0.7$  & \normalsize$ 32.4$\footnotesize$\pm1.5 $ & \normalsize$33.4 $\footnotesize$\pm 5.1$ & \normalsize$29.5$\footnotesize$ \pm 5.4$ & \normalsize$30.2 $\footnotesize$\pm 2.8$ & \normalsize$\textcolor[RGB]{192,0,0}{\bm{40.9}}$\footnotesize$\textcolor[RGB]{192,0,0}{\bm{\pm 1.7}}$ & \normalsize$36.5 $\footnotesize$\pm 3.0$ \\
\textbf{FruitBot} & \normalsize$26.7 $\footnotesize$\pm 0.8$ & \normalsize$24.5 $\footnotesize$\pm 0.7$ & \normalsize$ 27.3$\footnotesize$\pm0.8 $ & \normalsize$27.3$\footnotesize$\pm 1.8$ & \normalsize$28.2$\footnotesize$\pm 0.8$ & \normalsize$28.3$\footnotesize$\pm 0.9$ & \normalsize$29.8$\footnotesize$\pm 0.3$ & \normalsize$\textcolor[RGB]{192,0,0}{\bm{30.8}}$\footnotesize$\textcolor[RGB]{192,0,0}{\bm{ \pm 0.8}}$ \\
\textbf{BossFight} & \normalsize$7.7$\footnotesize$\pm 1.0$ & \normalsize$1.7$\footnotesize$\pm 0.9$ & \normalsize$ 8.2$\footnotesize$\pm0.7 $ & \normalsize$7.9$\footnotesize$\pm 0.6$ & \normalsize$7.5$\footnotesize$\pm 0.8$ & \normalsize$8.3$\footnotesize$\pm 0.8$ & \normalsize$9.7$\footnotesize$\pm 0.1$ & \normalsize$\textcolor[RGB]{192,0,0}{\bm{12.0}}$\footnotesize$\textcolor[RGB]{192,0,0}{\bm{ \pm 0.5}}$ \\
\textbf{Ninja} & \normalsize$5.9$\footnotesize$\pm 0.7$ & \normalsize$6.1$\footnotesize$\pm 0.8$ & \normalsize$ 6.8$\footnotesize$\pm0.5 $ & \normalsize$6.9$\footnotesize$\pm 0.8$ & \normalsize$7.0$\footnotesize$\pm 0.4$ & \normalsize$6.9$\footnotesize$\pm 0.6$ & \normalsize$5.8$\footnotesize$\pm 0.4$ & \normalsize$\textcolor[RGB]{192,0,0}{\bm{9.7 }}$\footnotesize$\textcolor[RGB]{192,0,0}{\bm{\pm 0.7}}$ \\
\textbf{Plunder} & \normalsize$5.0$\footnotesize$\pm 0.5$ & \normalsize$3.0$\footnotesize$\pm 0.6$ & \normalsize$ 5.9$\footnotesize$\pm0.5 $ & \normalsize$8.5$\footnotesize$\pm 1.2$ & \normalsize$\textcolor[RGB]{192,0,0}{\bm{9.5}}$\footnotesize$\textcolor[RGB]{192,0,0}{\bm{ \pm 1.0}}$ & \normalsize$8.9$\footnotesize$\pm 1.0$ & \normalsize$5.4$\footnotesize$\pm 0.7$& \normalsize$5.8$\footnotesize$\pm 1.0$ \\
\textbf{CaveFlyer} & \normalsize$5.1$\footnotesize$\pm 0.9$ & \normalsize$5.4$\footnotesize$\pm 0.8$ & \normalsize$ 6.1$\footnotesize$\pm0.6 $ & \normalsize$5.1$\footnotesize$\pm 0.6$ & \normalsize$6.3$\footnotesize$\pm 0.8$ & \normalsize$5.3$\footnotesize$\pm 0.9$ & \normalsize$5.0$\footnotesize$\pm 0.3$& \normalsize$\textcolor[RGB]{192,0,0}{\bm{7.5}}$\footnotesize$\textcolor[RGB]{192,0,0}{\bm{ \pm 1.0}}$ \\
\textbf{CoinRun} & \normalsize$8.5$\footnotesize$\pm 0.5$ & \normalsize$9.3$\footnotesize$\pm 1.4$ & \normalsize$ 8.6$\footnotesize$\pm0.3 $ & \normalsize$9.0$\footnotesize$\pm 0.8$ & \normalsize$8.8$\footnotesize$\pm 0.2$ & \normalsize$8.5$\footnotesize$\pm 0.6$ & \normalsize$\textcolor[RGB]{192,0,0}{\bm{9.6}}$\footnotesize$\textcolor[RGB]{192,0,0}{\bm{ \pm 0.1}}$ & \normalsize$9.2$\footnotesize$\pm 0.7$ \\
\textbf{Jumper} & \normalsize$5.8$\footnotesize$\pm 0.5$ & \normalsize$5.3$\footnotesize$\pm 0.6$ & \normalsize$ 6.0$\footnotesize$\pm0.3 $ & \normalsize$6.5$\footnotesize$\pm 0.6$ & \normalsize$6.6$\footnotesize$\pm 0.4$ & \normalsize$6.4$\footnotesize$\pm 0.6$ & \normalsize$5.6$\footnotesize$\pm 0.2$ & \normalsize$\textcolor[RGB]{192,0,0}{\bm{8.4}}$\footnotesize$\textcolor[RGB]{192,0,0}{\bm{\pm 1.6}}$ \\
\textbf{Chaser} & \normalsize$5.0$\footnotesize$\pm 0.8$ & \normalsize$1.4$\footnotesize$\pm 0.7$ & \normalsize$ 5.8$\footnotesize$\pm1.1 $ & \normalsize$5.9$\footnotesize$\pm 1.0$ & \normalsize$5.7$\footnotesize$\pm 0.6$ & \normalsize$6.7$\footnotesize$\pm 0.6$ & \normalsize$\textcolor[RGB]{192,0,0}{\bm{8.7}}$\footnotesize$\textcolor[RGB]{192,0,0}{\bm{\pm 0.2}}$ & \normalsize$4.8$\footnotesize$\pm 0.8$ \\
\textbf{Climber} & \normalsize$5.7$\footnotesize$\pm 0.8$ & \normalsize$5.3$\footnotesize$\pm 0.7$ & \normalsize$ 6.9$\footnotesize$\pm0.7 $ & \normalsize$6.9$\footnotesize$\pm 0.8$ & \normalsize$7.1$\footnotesize$\pm 0.7$ & \normalsize$6.5$\footnotesize$\pm 0.8$ & \normalsize$7.4$\footnotesize$\pm 0.3$ & \normalsize$\textcolor[RGB]{192,0,0}{\bm{8.1}}$\footnotesize$\textcolor[RGB]{192,0,0}{\bm{\pm 1.6}}$ \\
\textbf{DodgeBall} & \normalsize$\textcolor[RGB]{192,0,0}{\bm{11.7}}$\footnotesize$\textcolor[RGB]{192,0,0}{\bm{\pm 0.3}}$ & \normalsize$0.5$\footnotesize$\pm 0.4$ & \normalsize$ 1.7$\footnotesize$\pm0.4 $ & \normalsize$2.8$\footnotesize$\pm 0.7$ & \normalsize$4.3$\footnotesize$\pm 0.8$ & \normalsize$4.7$\footnotesize$\pm 0.7$ & \normalsize$7.2$\footnotesize$\pm1.2$ & \normalsize$3.8$\footnotesize$\pm 0.9$ \\
\textbf{Heist} & \normalsize$2.4$\footnotesize$\pm 0.5$ & \normalsize$2.4$\footnotesize$\pm 0.6$ & \normalsize$ 2.6$\footnotesize$\pm0.4 $ & \normalsize$4.1$\footnotesize$\pm 1.0$ & \normalsize$4.0$\footnotesize$\pm 0.8$ & \normalsize$4.0$\footnotesize$\pm 0.7$ & \normalsize$4.5$\footnotesize$\pm 0.2$ & \normalsize$\textcolor[RGB]{192,0,0}{\bm{7.7}}$\footnotesize$\textcolor[RGB]{192,0,0}{\bm{\pm 1.6}}$ \\
\textbf{Leaper} & \normalsize$4.9$\footnotesize$\pm 0.7$ & \normalsize$6.2$\footnotesize$\pm 0.5$ & \normalsize$ 5.3$\footnotesize$\pm1.1 $ & \normalsize$4.3$\footnotesize$\pm 1.0$ & \normalsize$5.3$\footnotesize$\pm 1.1$ & \normalsize$5.0$\footnotesize$\pm 0.3$ & \normalsize$\textcolor[RGB]{192,0,0}{\bm{9.2}}$\footnotesize$\textcolor[RGB]{192,0,0}{\bm{\pm 0.2}}$ & \normalsize$5.3$\footnotesize$\pm 1.5$ \\
\textbf{Maze} & \normalsize$5.7$\footnotesize$\pm 0.6$ & \normalsize$8.0$\footnotesize$\pm 0.7$ & \normalsize$ 5.2$\footnotesize$\pm0.5 $ & \normalsize$6.1$\footnotesize$\pm 1.0$ & \normalsize$6.6$\footnotesize$\pm 0.8$ & \normalsize$6.3$\footnotesize$\pm 0.6$ & \normalsize$5.9$\footnotesize$\pm 0.2$ & \normalsize$\textcolor[RGB]{192,0,0}{\bm{8.5}}$\footnotesize$\textcolor[RGB]{192,0,0}{\bm{\pm 1.6}}$ \\
\textbf{Miner} & \normalsize$8.5$\footnotesize$\pm 0.5$ & \normalsize$7.7$\footnotesize$\pm 0.6$ & \normalsize$ 9.4$\footnotesize$\pm0.4 $ & \normalsize$9.4$\footnotesize$\pm 1.2$ & \normalsize$\textcolor[RGB]{192,0,0}{\bm{9.8}}$\footnotesize$\textcolor[RGB]{192,0,0}{\bm{\pm 0.6}}$ & \normalsize$9.7$\footnotesize$\pm 0.7$ & \normalsize$\textcolor[RGB]{192,0,0}{\bm{9.8}}$\footnotesize$\textcolor[RGB]{192,0,0}{\bm{\pm 0.3}}$ & \normalsize$\textcolor[RGB]{192,0,0}{\bm{9.8}}$\footnotesize$\textcolor[RGB]{192,0,0}{\bm{\pm 0.9}}$\\
\bottomrule
\end{tabular}
}
\end{small}
\end{table}

\begin{figure}[ht] 
\centering 
\includegraphics[width=\textwidth]{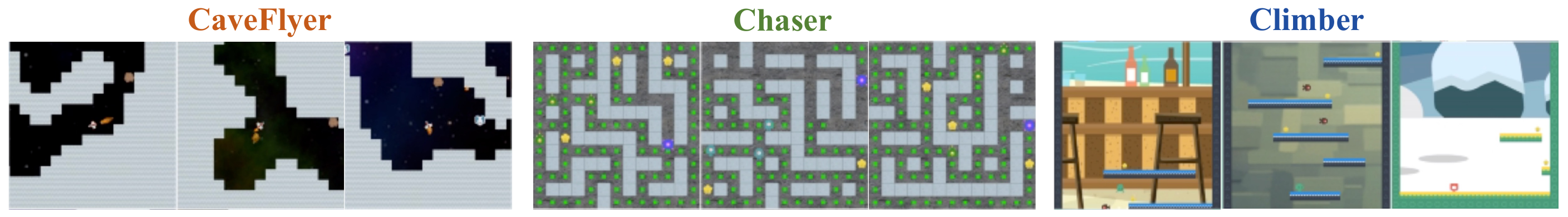} 
\caption{\textbf{Examples of three different game structures.}
As shown in Table~\ref{Best augmentation type of DrAC}, the most effective DA for Climber, CaveFlyer, and Chaser are color jitter, rotation, and cropping, respectively.} 
\label{ProcGen} 
\end{figure}

Table~\ref{Best augmentation type of DrAC} shows the best augmentation type for each game (copied from the original paper of DrAC~\cite{drac}).
Random cropping achieves the best performance on 9 out of 16 instances, while 5 out of 16 game environments benefit significantly from photometric transformations, including color jitter and random convolution.
For a detailed understanding of the connection between the properties of environments and augmentation types, Figure~\ref{ProcGen} suggests that the visual differences between the training environment and the test environment act as a major factor when determining the best augmentation type.
For example, the background styles of Climber vary significantly across different levels, and manipulating the color and other photometric factors is intuitively beneficial to generalization.

By contrast, the different levels of the maze game Chaser share similar visual information but exhibit increasing difficulty.
Consequently, applying photometric augmentations is likely to fail in this setting, which is consistent with the experimental results.
In such cases, the appropriate augmentation type is usually random cropping, which is beneficial to sample efficiency and contributes to improving the generalization performance.
In addition, CaveFlyer is uniquely friendly with the rotation augmentation, which is often destructive in other games. 
A closer check of the game shows that the major regions of the observations (except the gray areas) feature different positions and angles, and rotation can effectively narrow down the differences among them.

\begin{table}[ht]
\centering
\caption{\textbf{Best augmentation types for DrAC~\cite{drac} in different games.}
The original experiments~\cite{drac} investigate a set of eight transformations: cropping, grayscale, Cutout, Cutout-Color, flipping, rotation, random convolution and color jitter (all of them are shown in Figure~\ref{aug}).}
\label{Best augmentation type of DrAC}
\begin{small}
\setlength{\tabcolsep}{2pt}
\resizebox{\columnwidth}{!}{
\begin{tabular}{lcccccccc}
\toprule \textbf{Game} & BigFish & StarPilot & FruitBot & BossFight & Ninja & Plunder & \textcolor[RGB]{197,90,25}{\textbf{CaveFlyer}} & CoinRun \\
\midrule 
\textbf{Aug Type} & Crop & Crop & Crop & Flip & Color Jitter & Crop & \textcolor[RGB]{197,90,25}{\textbf{Rotate}} & Random Conv\\
\midrule
\midrule \textbf{Game} & Jumper & \textcolor[RGB]{84,130,53}{\textbf{Chaser}} & \textcolor[RGB]{31,78,161}{\textbf{Climber}} & DodgeBall & Heist & Leaper & Maze & Miner \\
\midrule \textbf{Aug Type} & Random Conv & \textcolor[RGB]{84,130,53}{\textbf{Crop}} & \textcolor[RGB]{31,78,161}{\textbf{Color Jitter}} & Crop & Crop & Crop & Crop & Color Jitter \\
\bottomrule
\end{tabular}
}
\end{small}
\end{table}
\vspace{-0.5\baselineskip}

\subsubsection{Vision Generalization on DMControl-GB}
As a variant of the DeepMind control suite~\cite{dmc}, DMControl-GB~\cite{soda} aims to evaluate the generalization ability of an agent by changing the image color or replacing the background with another image set (Figure~\ref{benchmark}).
A comparison of performance levels achieved in seen environments (Table~\ref{Sample Efficiency Evaluation in DMControl Suite}) and unseen environments (Table~\ref{Generalization Evaluation in DMControl-GB}) shows that although DrQ~\cite{drq} is prominent in terms of sample efficiency, the diversity derived from the naive application of cropping is limited, and a significant generalization gap is induced when this approach is transferred to unseen environments.
To provide sufficient visual diversity for generalization, it is necessary to use strong augmentations such as random convolution and overlay, as indicated by the results of follow-up studies.
\begin{table}[ht]
\centering
\caption{\textbf{Evaluation of Generalization Ability on DMControl-GB.} The scores (means and standard deviations) are obtained by conducting training in a fixed environment and performing evaluations in two unseen test environments with random colors (top) and natural video backgrounds (bottom).
}
\label{Generalization Evaluation in DMControl-GB}
\begin{small}
\setlength{\tabcolsep}{4pt}
\resizebox{\columnwidth}{!}{
\begin{tabular}{lccccccccccc}
\toprule
\textbf{DMControl-GB} & \textbf{SAC} & {\textbf{CURL}} & {\textbf{RAD}} & {\textbf{PAD}}& {\textbf{DrQ}} & {\textbf{SVEA}} & {\textbf{SODA}} & {\textbf{SIM}} & {\textbf{TLDA}} & {\textbf{SECANT}} & {\textbf{PIE-G}}  \\
(Random Colors) & \cite{sac} & \cite{curl}& \cite{rad} & \cite{pad} & \cite{drq} & \cite{svea} & \cite{soda} & \cite{sim} & \cite{tlda} & \cite{secant} & \cite{pieg} \\
\midrule
\textbf{Walker,} & \normalsize$365$ & \normalsize$662$ & \normalsize$644$ & \normalsize$797$ & \normalsize$770$ & \normalsize$942$ & \normalsize$930$ & \normalsize$940$ & \normalsize$\textcolor[RGB]{192,0,0}{\bm{947}}$ & \normalsize$939$ & \normalsize$941$ \\
\textbf{Stand} & \footnotesize$\pm 79$ & \footnotesize$\pm 54$ & \footnotesize$\pm 88$ & \footnotesize$\pm 46$ & \footnotesize$\pm 71$ & \footnotesize$\pm 26$ & \footnotesize$\pm 12$ & \footnotesize$\pm 2$ & \footnotesize$\textcolor[RGB]{192,0,0}{{\pm 26}}$ & \footnotesize$\pm 7$  & \footnotesize$\pm 35$\\
\textbf{Walker,} & \normalsize$144$ & \normalsize$445$ & \normalsize$400$ & \normalsize$468$ & \normalsize$520$ & \normalsize$760$ & \normalsize$697$ & \normalsize$803$ & \normalsize$823$ & \normalsize$856$ & \normalsize$\textcolor[RGB]{192,0,0}{\bm{884}}$ \\
\textbf{Walk} & \footnotesize$\pm 19$ & \footnotesize$\pm 99$ & \footnotesize$\pm 61$ & \footnotesize$\pm 47$ & \footnotesize$\pm 91$ & \footnotesize$\pm 145$ & \footnotesize$\pm 66$ & \footnotesize$\pm 33$ & \footnotesize$\pm 58$ & \footnotesize$\pm 31$ & \footnotesize$\textcolor[RGB]{192,0,0}{{\pm 20}}$ \\
\textbf{Cartpole,} & \normalsize$248$ & \normalsize$454$ &\normalsize$590$ & \normalsize$837$ &\normalsize$630$ & \normalsize$837$ & \normalsize$831$ & \normalsize$841 $ & \normalsize$760 $& \normalsize$\textcolor[RGB]{192,0,0}{\bm{866}}$ & \normalsize$749$ \\
\textbf{Swingup} & \footnotesize$\pm 24$ & \footnotesize$\pm 110$ & \footnotesize$\pm 53$ & \footnotesize$ \pm 63$ & \footnotesize$\pm 52$ & \footnotesize$ \pm 23$ & \footnotesize$\pm 21$ & \footnotesize$\pm 13$ & \footnotesize$\pm 60$& \footnotesize$\textcolor[RGB]{192,0,0}{{ \pm 15}}$ & \footnotesize$ \pm 46$\\
\textbf{Ball in cup,} & \normalsize$151$ & \normalsize$231$ & \normalsize$541$ & \normalsize$563$ & \normalsize$365$ & \normalsize$\textcolor[RGB]{192,0,0}{\bm{961}}$ & \normalsize$949$ & \normalsize$953$ & \normalsize$932$ & \normalsize$958$ & \normalsize$\textcolor[RGB]{192,0,0}{\bm{964}}$\\
\textbf{Catch} & \footnotesize$\pm 36$ & \footnotesize$\pm 92$ & \footnotesize$\pm 29$ & \footnotesize$\pm 50$ & \footnotesize$\pm 210$ & \footnotesize$\textcolor[RGB]{192,0,0}{{\pm 7}}$ & \footnotesize$\pm 19$ & \footnotesize$\pm 7$ & \footnotesize$\pm 32$ & \footnotesize$\pm 7$ & \footnotesize$\textcolor[RGB]{192,0,0}{{\pm 7}}$\\
\textbf{Finger,} & \normalsize$504$ & \normalsize$691$ & \normalsize$667$ & \normalsize$803$ & \normalsize$776$ & \normalsize$\textcolor[RGB]{192,0,0}{\bm{977}}$ & \normalsize$793$ & \normalsize$960$ & \multirow{2}*{\normalsize$-$} & \normalsize$910$ & \multirow{2}*{\normalsize$-$}\\
\textbf{Spin} & \footnotesize$\pm 114$ & \footnotesize$\pm 12$ & \footnotesize$\pm 154$ & \footnotesize$\pm 72$ & \footnotesize$\pm 134$ & \footnotesize$\textcolor[RGB]{192,0,0}{{\pm 5}}$ & \footnotesize$\pm 128$ & \footnotesize$\pm 6$ & ~ & \footnotesize$\pm 115$ & ~\\
\textbf{Cheetah,} & \normalsize$133$ & \multirow{2}*{\normalsize$-$} & \multirow{2}*{$-$} & \normalsize$159$ & \normalsize$100$ & \normalsize$273$ & \normalsize$294$ & \multirow{2}*{\normalsize$-$} & \normalsize$371$ & \normalsize$\textcolor[RGB]{192,0,0}{\bm{582}}$ & \normalsize$364$ \\
\textbf{Run} & \footnotesize$\pm 26$ & ~ & ~ & \footnotesize$\pm 28$ & \footnotesize$\pm 27$ & \footnotesize$\pm 23$ & \footnotesize$\pm 34$ & ~ & \footnotesize$\pm 51$ & \footnotesize$\textcolor[RGB]{192,0,0}{{\pm 64}}$ & \footnotesize$\pm 40$\\
\midrule
\midrule
(Natural Videos) &  &  & & & &  &  & & & & \\
\midrule
\textbf{Walker,} & \normalsize$274$ & \normalsize$852$ & \normalsize$745$ & \normalsize$935$ & \normalsize$873$ & \normalsize$961$ & \normalsize$955$ & \normalsize$963$ & \normalsize$\textcolor[RGB]{192,0,0}{\bm{973}}$ & \normalsize$932$ & \normalsize$957$ \\
\textbf{Stand} & \footnotesize$\pm 39$ & \footnotesize$\pm 75$ & \footnotesize$\pm 146$ & \footnotesize$\pm 20$ & \footnotesize$\pm 83$ & \footnotesize$\pm 8$ & \footnotesize$\pm 13$ & \footnotesize$\pm 5$ & \footnotesize$\textcolor[RGB]{192,0,0}{{\pm 6}}$ & \footnotesize$\pm 15$ & \footnotesize$\pm 12$ \\
\textbf{Walker,} & \normalsize$104$ & \normalsize$556$ & \normalsize$606$ & \normalsize$717$ & \normalsize$682 $ & \normalsize$819$ & \normalsize$768$ & \normalsize$861$ & \normalsize$\textcolor[RGB]{192,0,0}{\bm{873}}$ & \normalsize$842$ & \normalsize$\textcolor[RGB]{192,0,0}{\bm{870}}$\\
\textbf{Walk} & \footnotesize$\pm 14$ & \footnotesize$\pm 133$ & \footnotesize$\pm 63$ & \footnotesize$\pm 79$ & \footnotesize$\pm 89$ & \footnotesize$\pm 71$ & \footnotesize$\pm 38$ & \footnotesize$\pm 33$ & \footnotesize$\textcolor[RGB]{192,0,0}{{\pm 34}}$ & \footnotesize$\pm 47$ & \footnotesize$\textcolor[RGB]{192,0,0}{\bm{\pm 22}}$\\
\textbf{Cartpole,} & \normalsize$204$ & \normalsize$404$ & \normalsize$373$ & \normalsize$521$ & \normalsize$485$ & \normalsize$\textcolor[RGB]{192,0,0}{\bm{782}}$ & \normalsize$758$ & \normalsize$770 $ & \normalsize$671$ & \normalsize$752$ & \normalsize$597$\\
\textbf{Swingup} & \footnotesize$\pm 20$ & \footnotesize$\pm 67$ & \footnotesize$\pm 72$ & \footnotesize$\pm 76$ & \footnotesize$\pm 105$ & \footnotesize$\textcolor[RGB]{192,0,0}{{\pm 27}}$ & \footnotesize$\pm 62$ & \footnotesize$\pm 13$ & \footnotesize$\pm 57$ & \footnotesize$\pm 38$ & \footnotesize$\pm 61$\\
\textbf{Ball in cup,} & \normalsize$172$ & \normalsize$316$ & \normalsize$481$ & \normalsize$436$ & \normalsize$318$ & \normalsize$871$ & \normalsize$875$ & \normalsize$820$ & \normalsize$887$ & \normalsize$903$ & \normalsize$\textcolor[RGB]{192,0,0}{\bm{922}}$ \\
\textbf{Catch} & \footnotesize$\pm 46$ & \footnotesize$\pm 119$ & \footnotesize$\pm 26$ & \footnotesize$\pm 55$ & \footnotesize$\pm 157$ & \footnotesize$\pm 106$ & \footnotesize$\pm 56$ & \footnotesize$\pm 135$ & \footnotesize$\pm 58$ & \footnotesize$\pm 49$ & \footnotesize$\textcolor[RGB]{192,0,0}{{\pm 20}}$ \\
\textbf{Finger,} & \normalsize$276$ & \normalsize$502$ & \normalsize$400$ & \normalsize$691$ & \normalsize$533$ & \normalsize$808$ & \normalsize$695$ & \normalsize$81$ & \multirow{2}*{\normalsize$-$} & \normalsize$\textcolor[RGB]{192,0,0}{\bm{861}}$ & \normalsize$837$ \\
\textbf{Spin} & \footnotesize$\pm 81$ & \footnotesize$\pm 19$ & \footnotesize$\pm 64$ & \footnotesize$\pm 80$ & \footnotesize$\pm 119$ & \footnotesize$\pm 33$ & \footnotesize$\pm 94$ & \footnotesize$\pm 38$ & ~ & \footnotesize$\textcolor[RGB]{192,0,0}{{\pm 102}}$ & \footnotesize$\pm 107$ \\
\textbf{Cheetah,} & \normalsize$80$ & \multirow{2}*{\normalsize$-$} & \multirow{2}*{\normalsize$-$} & \normalsize$206$ & \normalsize$102$ & \normalsize$292$ & \normalsize$229$ & \multirow{2}*{\normalsize$-$} & \normalsize$356$ & \normalsize$\textcolor[RGB]{192,0,0}{\bm{428}}$ & \normalsize$287$ \\
\textbf{Run} & \footnotesize$\pm 19$ & ~ & ~ & \footnotesize$\pm 34$ & \footnotesize$\pm 30$ & \footnotesize$\pm 32$ & \footnotesize$\pm 29$ & ~ & \footnotesize$\pm 52$ & \footnotesize$\textcolor[RGB]{192,0,0}{{\pm 70}}$ & \footnotesize$\pm 20$\\
\bottomrule
\end{tabular}
}
\end{small}
\end{table}

SVEA~\cite{svea} and TLDA~\cite{tlda} both significantly outperform DrQ~\cite{drq} by focusing on stabilizing the training process when leveraging strong augmentation to optimize the representation and policy together.
Another way to improve generalization is to decouple the unsupervised representation learning and the RL optimization process, either in an iterative manner (e.g., SODA~\cite{soda} and SIM~\cite{sim}) or in a sequential manner (e.g., SECANT~\cite{secant}).
Moreover, pre-trained encoders from off-the-shelf image datasets such as PIE-G~\cite{pieg} from ImageNet~\cite{imagenet} also show attractive potential to provide generalizable representations in downstream tasks.

%% file: textfile/6_Discussion_and_future_works.tex
\newpage
\section{Discussion and Future Works}
\label{Discussion of future works}
DA techniques have substantially improved the sample efficiency and generalization abilities of visual RL methods; however, many challenges remain to be addressed.
In this section, we elaborate on these points and highlight key directions for future research, encompassing the opportunities, limitations, and underlying mechanisms of leveraging DA in visual RL.

\subsection{Towards Semantic-Level DA}
\label{Semantic-level Data Augmentation}
It can be considered as a kind of feature manipulation technique that alters the relative contributions of task-relevant and task-irrelevant features in the gradient update steps of the utilized network~\cite{Feature_Manipulation}.
In this context, an ideal (albeit theoretical) DA method would operate at the semantic level, possessing the capability to precisely identify features pertinent to the current label or task while effectively perturbing irrelevant information. 

\vspace{0.5\baselineskip}
However, this assumption underlying label-preserving transformations in SL and optimality-invariant transformations in RL proves challenging to uphold in practice, particularly when implementing pixel-level augmentations.
The fundamental limitation of pixel-level approaches becomes evident when considering their mechanism: such augmentations, which aim to transform each pixel in a context-agnostic manner, inherently struggle to discriminate between task-relevant and task-irrelevant information ~\cite{tlda}. 
This indiscriminate modification of pixels often results in the inadvertent alteration of critical task-relevant features, thereby compromising the efficacy of DA techniques, particularly in the domain of visual RL.
Therefore, a promising avenue for advancing DA techniques lies in the development of semantic-level augmentation strategies. These strategies offer a more sophisticated approach compared to conventional pixel-level manipulations.

\vspace{0.5\baselineskip}
Several recent studies have attempted to move towards semantic-level DA by focusing on preserving task-relevant information.
For example, EXPAND~\cite{expand} and TLDA~\cite{tlda}, as discussed in Section~\ref{Context-aware Data Augmentation}, propose methods to prevent augmentation of salient or sensitive areas in observed images, thereby maintaining critical visual information.
In the broader computer vision community, KeepAugment~\cite{keepaugment} uses a saliency map to identify the key regions and then preserves these informative regions during augmentation to produce reliable training samples.
While not directly a DA technique, VAI~\cite{vai} employs unsupervised keypoint detection and visual attention mechanisms, combined with a reconstruction loss, to compel the encoder to embed only the foreground information of the input image. 
This method effectively introduces an inductive bias based on the assumption that key information controlling the objective in observations typically resides in the foreground, thereby prioritizing task-relevant features.

\vspace{0.5\baselineskip}
Recent advancements in pre-trained generative models have opened new avenues for achieving semantic-level augmentation~\cite{GenAug}. 
The robust generative capabilities of large multi-modal models offer a promising DA approach that can be controlled through prompts. 
For instance, these models can be instructed to "replace the background of a robotic arm with an outdoor scene while keeping the arm itself unchanged". 
Such prompt-driven DA transformations have the potential to generate diverse augmented data while retaining semantic invariance.
The application of generative augmentation in visual RL is still in its nascent stages. 
This emerging field presents a fertile ground for further exploration, particularly in developing methods that can leverage the semantic understanding of these models to produce task-relevant augmentations. 
Future research could focus on optimizing prompt engineering for RL-specific tasks, investigating the impact of generated data on policy learning, and developing techniques to ensure the generated augmentations align with the underlying dynamics of the RL environment.

\subsection{Trade-off between Training Stability and Generalization}
\label{Trade-off between the Instability and Generalization Ability Caused by DA}
In practice, DrQ~\cite{drq, drq_v2} and RAD~\cite{rad} that leverage weak DA such as random cropping as implicit regularization methods can yield significantly improved sample efficiency during training, while a noticeable generalization gap is observed when these approaches are transferred to unseen environments~\cite{soda,svea,secant}.
Furthermore, more diverse augmentations, such as color jitter, have the potential to improve generalization but tend to result in unstable optimization and poor sample efficiency~\cite{rad,svea}.
Therefore, a dilemma of balancing between stability and generalization is persistent when applying DA in visual RL.
This challenge is particularly acute due to the inherently fragile nature of the optimization process in RL~\cite{secant}.

\vspace{0.5\baselineskip}
This dilemma is frequently attributed to the conflation of policy optimization and representation learning in current end-to-end visual RL algorithms~\cite{secant, soda}. Consequently, a logical approach is to decouple these processes, independently learning a robust representation and a competent policy, as elaborated in Section~\ref{Task-Specific Representation Decoupled from Policy Optimization} and Section~\ref{Task-Agnostic Representation using Unsupervised Learning}.
Such decoupling facilitates the application of heavy augmentations to improve generalization while simultaneously employing weak augmentations to maintain satisfactory sample efficiency~\cite{soda,secant}. This strategy effectively addresses the stability-generalization trade-off.
Moreover, the dilemma is exacerbated by the limitations of pixel-level augmentation techniques. These methods, when applied intensively, risk inadvertently destroying critical features, further complicating the balance between effective augmentation and preservation of essential information.

\vspace{0.5\baselineskip}
Further insight into this dilemma can be gained through the lens of the bias-variance trade-off, a fundamental principle in machine learning~\cite{Overfitting_and_undercomputing}. 
Contemporary complex models, such as DNNs, typically exhibit low bias but high variance. 
Consequently, these models are prone to overfitting the training data, resulting in sub-optimal performance on unseen data.
DA addresses this issue by introducing increased diversity, thereby reducing variance and enhancing the model's generalization capabilities~\cite{study_on_overfitting}.
Although DA may mitigate the issue of overfitting, certain augmentation combinations can actually lead to underfitting, making the training process unstable and challenging~\cite{0308_Regularising_for_invariance, survey_of_regularization}.
Note that the issue of underfitting is more detrimental in RL, as its optimization process is more unstable than those of supervised tasks.

\vspace{0.5\baselineskip}
Overall, there are two potential paths to further balance training stability and generalization. Firstly, designing more effective augmented data generation methods to avoid corrupting task-relevant information, as discussed in Section~\ref{Semantic-level Data Augmentation}. Secondly, optimizing existing paradigms for leveraging DA, such as decoupling policy optimization and representation learning.

\subsection{Underlying Mechanisms of DA Efficacy in Visual RL}
\label{Theoretic Frameworks for Data Augmentation}
Given the widespread application and powerful impact of DA in visual RL, understanding its underlying mechanisms has become crucial. 
However, most works have simply employed DA as a basic component without delving deeper into `why DA works'~\cite{drq,rad,drq_v2}.
To gain a comprehensive understanding of DA's effectiveness in visual RL, this section begins by briefly reviewing existing studies on the underlying mechanisms of DA within the broader context of deep learning (Section~\ref{Prevailing Perspectives on DA Effectiveness in DL}). Subsequently, we will explore in depth the unique mechanisms of DA's efficacy in visual RL tasks in Section~\ref{Specialized Mechanisms of DA in Visual RL}.

\subsubsection{Prevailing Perspectives on DA Effectiveness in DL}
\label{Prevailing Perspectives on DA Effectiveness in DL}

In recent years, numerous efforts have been made to investigate the theoretical guarantees for DA from various perspectives. These guarantees provide researchers with valuable insights into the practical effects of such approaches~\cite{0308_Regularising_for_invariance,Feature_Manipulation,lecun_Class_Dependent}.
This section offers a concise overview of previous works on the theoretical foundations of DA, categorizing them into three main perspectives:
implicit regularization~\cite{survey_of_regularization,lecun_Class_Dependent,instead_of_explicit_regularization},
invariance learning~\cite{0308_Regularising_for_invariance,Learning_with_invariances}
and feature manipulation~\cite{Feature_Manipulation,understanding_the_feature_learning}.

\paragraph{Implicit Regularization vs. Explicit Regularization.}
Regularization is a fundamental technique in deep learning that aims to prevent overfitting and improve generalization abilities by constraining the complexity of a model~\cite{survey_of_regularization,ICML20_DL,clop}.
The regularization strategies of DA act on the training data instead of the model’s parameters and hence can be considered a type of implicit regularization approach instead of an explicit regularization technique that imposes constraints on the parameters, such as minimizing the $\mathcal{L}_2$ norms of the parameters~\cite{lecun_Class_Dependent}.
By keeping the parameter space intact, this data-driven regularization approach can maintain the model’s representational capacity while increasing its robustness~\cite{drq,instead_of_explicit_regularization}.
In fact, DA is more straightforward than explicit regularization that integrates the prior knowledge into objective functions, and neural networks can implicitly encode the attributes of DA without explicitly training towards these objectives~\cite{Do_CNNs, yang2023neumeta, dablain2023towards}.
Furthermore, attempts have also been made to derive explicit regularizers to describe the implicit regularization effect of DA~\cite{A_Thousand_Samples,hu2024revisiting}.

\paragraph{\textbf{Transformation Invariance.}}
Invariance is an essential property of all intelligent systems that makes them generalize effectively~\cite{learning_invariances}.
The purpose of DA is to constrain a model’s output to be invariant when applying task-irrelevant transformations to the input data~\cite{mixup, drac}.
It has been widely accepted that translation invariance is an inherent feature of CNNs~\cite{Do_CNNs}, whereas other types of transformation invariance, such as rotation invariance, must be induced by corresponding augmentations~\cite{Invariance_Learning_220222}.
The definition of DA assumes that semantics are invariant to data transformation~\cite{image_aug_survey, drq}, which implies that performing optimization with augmentation can result in implicit invariances.
Furthermore, the specific details of augmentations can be used to encode prior knowledge about task-specific or dataset-specific invariances~\cite{learning_invariances}.

\paragraph{\textbf{Feature Manipulation.}}
An alternative explanation of how DA works is derived from the perspective of feature manipulation~\cite{zou2023benefits, yang2022deep, yang2022factorizing}.
Learning meaningful features from high-dimensional data is empirically challenging, as critical features are often highly sparse and associated with spurious features such as dense noise.
In practice, this may result in the network's overfitting the noisy features instead of properly learning the critical features~\cite{Representation_Learning_A_Survey}.
First, by adjusting the relative contributions of the original data features, DA can effectively facilitate the incorporation of informative but hard-to-learn features into the learning process~\cite{Feature_Manipulation}.
Second, the latest research shows that leveraging DA in contrastive learning can decouple spurious features from the representations of positive samples.
By ignoring the decoupled features, the performance of networks may be boosted by focusing on the learning of resistant features~\cite{understanding_the_feature_learning}.

\subsubsection{Specialized Mechanisms of DA in Visual RL}
\label{Specialized Mechanisms of DA in Visual RL}

The underlying mechanisms of DA in visual RL partially align with those observed in other domains, as discussed in Section~\ref{Prevailing Perspectives on DA Effectiveness in DL}. 
For instance, applying random shift to input observations without modifying other algorithmic details can be viewed as a form of implicit regularization~\cite{Learning_Better_with_Less, hu2024revisiting}.
Additionally, the improvements in visual generalization achieved through DA can be attributed to the previously mentioned concepts of transform invariance and feature manipulation~\cite{Generalisation_survey}.
However, several remarkable phenomena unique to applying DA in visual RL cannot be adequately explained by mechanisms from other domains.

\begin{wrapfigure}{r}{0.4\textwidth}
\centering
\vspace{-\baselineskip}
\includegraphics[width=0.398\textwidth]{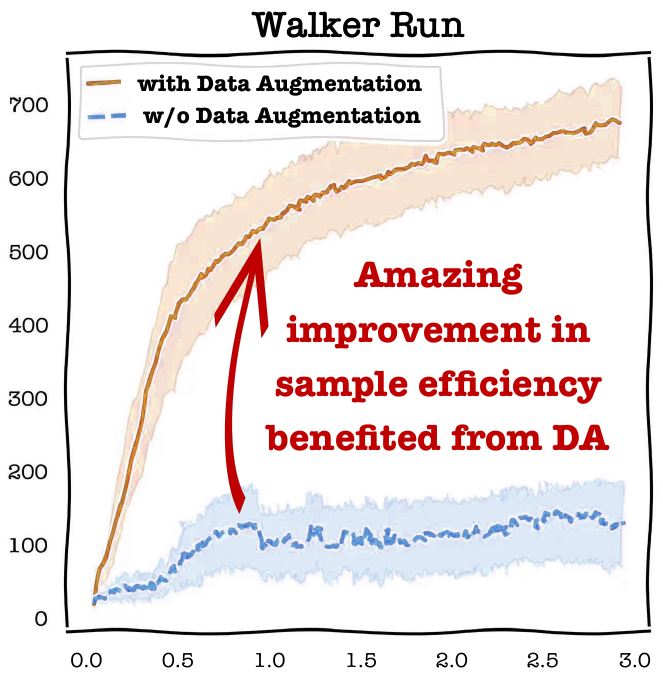}
\caption{Unlike in tasks from other domains, DA decisively enhances the training efficiency of visual RL.}
\label{Motivation}
\end{wrapfigure}

\vspace{0.5\baselineskip}
As illustrated by the blue training curve in Figure~\ref{Motivation}, visual RL agents (based on the DDPG algorithm~\cite{ddpg} in this example) fail to achieve an effective decision policy in the classic continuous control task \textit{Walker Run} from the DeepMind Control suite~\cite{dmc}.
Remarkably, merely applying simple random shift transformations to input observations, without any other modifications to the algorithm, results in superior performance~\cite{drq_v2}, as demonstrated by the orange curve in Figure~\ref{Motivation}.
This striking improvement in training efficiency for visual RL stands in stark contrast to the incremental performance gains typically observed when applying DA in other task domains~\cite{2022image_da_survey}. 
Such a significant disparity suggests that previous understandings of DA's mechanisms are insufficient to fully explain its role in visual RL tasks. 
Instead, this remarkable enhancement implies that DA must effectively overcome some critical bottleneck unique to visual RL that previously limited its training efficacy. 
This phenomenon underscores the need for a deeper investigation into the specific mechanisms by which DA contributes to the success of visual RL algorithms.

\vspace{0.5\baselineskip}
The most recent investigation~\cite{ma2023revisiting} has unveiled that DA's remarkable effectiveness in visual RL stems from its ability to mitigate the \textbf{plasticity loss} of deep neural networks. 
Plasticity, referring to the capacity of deep neural networks to continually learn from new data, gradually diminishes during training with non-stationary objectives~\cite{nikishin2022primacy, sokar2023dormant}. 
The inherent nature of DRL necessitates that agents continuously refine their policies through environmental interactions, resulting in intrinsically non-stationary data streams and optimization objectives~\cite{kumar2023maintaining}. 
This characteristic of DRL paradigms renders plasticity loss a critical impediment to achieving sample-efficient applications, as the networks must maintain their adaptability throughout the learning process~\cite{lyle2023understanding, nikishin2024deep}.
Compared to traditional state-based RL tasks, visual RL algorithms suffer from more severe plasticity loss due to increased task complexity and larger network architectures. 
It has been demonstrated through ingenious experiments~\cite{ma2023revisiting} that without DA, agents fail to train effectively due to catastrophic plasticity loss, while applying DA significantly alleviates this bottleneck. 
Furthermore, DA's efficacy in mitigating plasticity loss surpasses that of several interventions specifically designed for this purpose, such as Layer Normalization~\cite{lyle2024normalization}, Shrink $\&$ Perturb~\cite{ash2020warm}, and CReLU~\cite{abbas2023loss}.

\subsection{Unique Characteristics of DA in Visual RL versus Other Domains}
\label{Data Augmentation in Supervised Learning vs. in Reinforcement Learning}

Given the widespread application of DA across various domains in deep learning~\cite{mixup,alexnet}, a pertinent question naturally arises: 
\begin{center}
\textit{Does the implementation of DA in visual RL exhibit significant distinctions from its utilization in other domains, particularly in supervised and unsupervised vision tasks?}
\end{center} 
Indeed, there are substantial differences, which underscores the necessity of organizing a survey specifically focused on DA in visual RL.
This section will delineate the most salient differences in DA implementation between visual RL and other domains. Further exploration of these distinctions and the development of tailored DA techniques for the visual RL scenario represent crucial directions for future research.

\paragraph{Diverse Augmentable Data Types.}
Compared to typical supervised or unsupervised vision tasks, RL data is inherently more complex.
On one hand, RL data encompasses three distinct elements: state, action, and reward. On the other hand, RL involves long sequences of sequential decision-making data, known as trajectories. 
This complexity in data types enables a more diverse and flexible approach to transforming different data components in visual RL tasks. 
As introduced in Section~\ref{How to augment the data in Visual RL?}, when categorizing based on the type of data being augmented, visual RL incorporates at least three classes of augmentation: observation augmentation, transition augmentation, and trajectory augmentation.
While existing works predominantly employ the simplest and most direct approach of observation augmentation~\cite{rad,drq_v2}, the advancement of generative models such as diffusion techniques is likely to precipitate the emergence of more sophisticated yet diverse DA techniques in visual RL scenarios~\cite{he2024diffusion, zhu2023diffusion}. 
This trend towards more complex augmentation strategies promises to exploit the rich structure of RL data more fully, potentially leading to significant improvements in sample efficiency and generalization capabilities.

\paragraph{Distinctive Implementation Details.}
Due to the nonstationary nature of RL, there are significant differences in the optimal practices for applying DA in visual RL compared to other scenarios. 
Firstly, contrary to supervised vision tasks such as image classification, where heavy transformations like Mixup~\cite{mixup} and CutMix~\cite{cutmix} demonstrate notable advantages over traditional image transformations, in visual RL, random cropping has emerged as the most practical augmentation technique~\cite{rad,drac}. 
Consequently, the approach to harnessing augmented data must be thoughtfully designed to avoid potential destabilization of the optimization process while effectively exploiting the generalization capabilities induced by DA~\cite{tlda,Learning_Better_with_Less}.
Secondly, the timing of augmentation application is critical in RL~\cite{indaexda}, in contrast to supervised learning tasks~\cite{Time_matters_in_regularizing_deep_networks}, due to the heightened time sensitivity of augmentation in RL contexts. 
For example, optimality-invariant augmentations such as cropping should be implemented as early as possible to enhance sample efficiency and expedite the RL training process. 
Conversely, prior-based strong augmentations like color jitter may interfere with RL training, suggesting that these techniques might be more effectively employed after RL optimization for knowledge distillation~\cite{indaexda}.

\paragraph{Unique Underlying Mechanisms.}
As discussed in Section~\ref{Specialized Mechanisms of DA in Visual RL}, contrary to the general understanding of DA's mechanisms, it has been discovered that DA effectively addresses a bottleneck unique to RL: plasticity loss~\cite{ma2023revisiting}. 
This insight explains the remarkable extent to which DA enhances the sample efficiency of visual RL algorithms~\cite{drq_v2}. 
The distinctive mechanism through which DA operates in this context necessitates a paradigm shift in our approach to designing DA methods for visual RL. 
Specifically, future work must adopt a novel perspective in approaching DA for visual RL by focusing on mitigating plasticity loss, a consideration that has been largely overlooked in traditional DA approaches. 
This shift in focus represents a significant departure from conventional DA strategies and opens up new avenues for research in visual RL.

\subsection{The Role of Visual RL and DA in the Age of Foundation Models}
\label{The Role of Visual RL and DA in the Age of Foundation Models}

In recent years, foundation models, particularly large language models, have emerged rapidly, showcasing extraordinary intelligence and driving a new wave of AI innovation~\cite{dubey2024llama,firoozi2023foundation}. 
In this era of large foundation models, we must consider two key questions: 
\begin{center}
\textit{Does the classic RL paradigm, as represented by visual RL, still retain research value and necessity?}
\end{center} 
\begin{center}
\textit{Additionally, what role does DA play in this evolving landscape?}
\end{center} 
Clarifying the roles of visual RL and DA in the Age of Foundation Models is important for two reasons. 
First, it ensures that this survey's content remains pertinent and valuable in the current AI landscape. 
Second, it offers meaningful guidance for the future development of visual RL and DA. 
This section will address these two questions systematically.

\vspace{0.5\baselineskip}
Primarily, the classic RL paradigm, exemplified by visual RL, remains an indispensable component in achieving super-human decision-making intelligence.
Currently, there are two primary approaches to leveraging foundation models in decision-making tasks. 
\begin{itemize}
\item The first approach leverages pre-trained multi-modal large models for high-level perception and planning, harnessing their comprehensive understanding and reasoning capabilities to enhance strategic decision-making in complex environments~\cite{kim2024openvla, NEURIPS2023_4ec43957, li2024cog}. In this context, RL remains crucial for training the low-level control policy. This policy generates specific actions based on task-relevant features extracted by foundation models, enabling effective agent-environment interaction~\cite{ma2024survey, tang2024kalie}. Consequently, RL bridges the gap between the high-level perception and planning capabilities of foundation models and the concrete action execution required in real-world scenarios.
\item The second category of methods adopts the pre-training paradigm for data-driven offline policy training~\cite{prudencio2023survey, tarasov2024revisiting, lee2022offline}, drawing inspiration from the success of foundation models in other domains. In this paradigm, offline pre-training can be viewed as seeking an optimal initialization for online RL, while online fine-tuning is imperative for achieving high-level decision-making intelligence~\cite{kim2024unsupervised, nakamoto2024cal}.
This approach addresses two critical aspects.
Firstly, due to the inherent distribution shift between offline data and real-world scenarios, online RL is necessary to correct these biases~\cite{yue2022boosting}.
Secondly, only through exploration in online learning can the agent transcend the limitations of human-collected offline data and potentially surpass human-level intelligence~\cite{lambert2022challenges, wen2023towards}.
\end{itemize}

Furthermore, DA will undoubtedly continue to play a crucial role in visual RL and other RL tasks that utilize high-dimensional features as input.
Primarily, DA remains a direct and effective method for expanding datasets and incorporating human prior knowledge~\cite{he2024diffusion,zhu2023diffusion}. 
In decision-making tasks such as robotic control, the availability of training data is significantly more limited compared to language and vision tasks. Consequently, designing more powerful DA techniques is essential for training large decision models.
Moreover, as discussed in Section~\ref{Specialized Mechanisms of DA in Visual RL}, DA effectively mitigates plasticity loss during online RL training. 
Without DA, even when employing pre-trained visual encoders for feature extraction, visual RL algorithms would still suffer from catastrophic plasticity loss, impeding efficient training~\cite{pieg, ma2023revisiting}. 
As foundation models continue to evolve, RL is tasked with handling increasingly complex decision-making scenarios, and agents are required to possess continual fine-tuning capabilities.
This evolution in the field underscores the critical need for agents to retain adaptive capabilities, emphasizing the continued relevance and importance of DA research and application.

\subsection{Limitations of DA}
\label{Limitations of Data Augmentation}
While DA has demonstrated significant benefits in visual RL, as extensively discussed in the preceding sections, it is crucial to acknowledge its limitations and potential drawbacks to provide a comprehensive understanding of its applicability and effectiveness.

\begin{enumerate}
\item The applications of DA are highly task-specific and require extensive expert knowledge~\cite{drac}. In practice, DA's effectiveness depends on prior knowledge of the variations between training and test environments, allowing for reliable specification of appropriate augmentation techniques~\cite{Generalisation_survey}.
For example, in DMControl-GB, only visual settings such as background colors are varied in the test environments, and specific DA techniques, such as random convolution, can effectively capture these prior variations~\cite{svea,secant}.
\item DA effectively mitigates plasticity loss in single-task visual RL, ensuring efficient training~\cite{ma2023revisiting}. However, it proves insufficient for maintaining adequate plasticity in open-ended and continual RL scenarios, where significant plasticity degradation occurs over extended training periods despite its application~\cite{abbas2023loss, ahn2024reset}.
This limitation calls for more targeted interventions to maintain long-term plasticity in open-ended, multi-task RL, essential for developing adaptive foundational RL policies.
\item DA, which modifies observations after they are generated, provides versatile applicability without necessitating direct simulator manipulation~\cite{rad}.
This characteristic is particularly advantageous when the underlying simulation environment is inaccessible or unmodifiable. 
However, in scenarios where direct access to the simulator is available, domain randomization (DR) can generate more diverse and precise data~\cite{mehta2020active, chen2021understanding}. 
Consequently, in specific fields such as robot learning, DR may prove more effective than DA~\cite{horvath2022object, huber2024domain}.
\item Current DA techniques, primarily focused on image transformations of observations, are insufficient for generating truly diverse synthetic data. 
This limitation stems from their reliance on human-defined priors, which constrains the injection of novel, informative knowledge into the data~\cite{svea}.
However, the advent of increasingly powerful pre-trained generative models presents a promising avenue to overcome this bottleneck, potentially enabling DA to produce substantially richer training data~\cite{he2024diffusion}.
\end{enumerate}

%% file: textfile/7_Conclusion.tex
\newpage
\section{Conclusion}
\label{Conclusion}
In this paper, we present a comprehensive survey of DA in the paradigm of visual RL.
We first propose the High-Dimensional Contextual Markov Decision Process (HCMDP) as a general framework, elucidating the motivations for DA in improving sample efficiency and generalization.
The main body of this survey meticulously examines existing related works, structured around two central themes: how to augment data and how to effectively utilize the augmented data. 
Subsequently, experimental results from widely used benchmarks demonstrate the efficacy of these techniques in visual RL.
This survey also provides a list of current challenges and potential directions for future studies.
In the following, we present a few suggestions and insights that are intended to benefit the relevant communities.

\begin{enumerate}
\item Compact and robust representation is vital for acquiring \textit{sample-efficient} and \textit{generalizable} visual RL agents; therefore, it is necessary to apply appropriate representation learning strategies to tackle the specific challenges of visual RL (Section~\ref{Major Challenges in Visual RL}).
As a data-driven technique, DA is an essential component of representation learning and has great potential to be further explored (Section~\ref{Data Augmentation in Visual RL}).
\item 
To fully harness the potential of DA, two complementary aspects must be addressed: 
how to augment data (Section~\ref{How to augment the data in Visual RL?}) and how to effectively leverage augmented data (Section~\ref{How to leverage}).
The key to further advancing these aspects lies in two critical objectives: achieving semantic-level DA (Section~\ref{Semantic-level Data Augmentation}), and attaining an effective balance between training stability and generalization ability (Section~\ref{Trade-off between the Instability and Generalization Ability Caused by DA}). 
\item Beyond the common benefits such as regularization and feature manipulation that DA provides across all deep learning scenarios, there exists a unique mechanism behind DA's significant enhancement of visual RL training efficiency: its ability to effectively mitigate plasticity loss. Overcoming this RL-specific bottleneck should be a focal point for future research, emphasizing the development of more targeted augmentation strategies and other interventions (Section~\ref{Theoretic Frameworks for Data Augmentation}).
\item The difference between RL and SL should be given special attention when applying DA in visual RL, including fragile optimization process, the interactive data acquisition process and its absence of ground-truth labels (Section~\ref{Data Augmentation in Supervised Learning vs. in Reinforcement Learning}).
\item Visual RL, as a representative paradigm for learning control policies from high-dimensional features, remains a crucial component in the current era of large foundation models. DA, as a key element in achieving efficient and generalizable visual RL, warrants continued in-depth investigation (Section~\ref{The Role of Visual RL and DA in the Age of Foundation Models}).
\end{enumerate}

Overall, this survey strives to provide the first unified and principled framework for the large body of thriving research on DA in visual RL.
We expect it to serve as a valuable guide for researchers and practitioners, and stimulate more inspiration in this fascinating field.